\documentclass[12pt]{elsarticle}
\makeatletter
\def\ps@pprintTitle{%
    \let\@oddhead\@empty
    \let\@evenhead\@empty
    \let\@oddfoot\@empty
    \let\@evenfoot\@empty}
\makeatother
\usepackage{todonotes}
\usepackage{xpatch}
\usepackage{array}
\usepackage{siunitx}
\usepackage{amsmath}
\usepackage{amssymb}
\usepackage{xcolor}
\usepackage{bbm}
\usepackage{amsgen,amsmath,amstext,amsbsy,amsopn,tikz,amssymb}
\usepackage{listings}
\usepackage{xcolor}
\usepackage{algorithm}
\usepackage{algorithmicx}
\usepackage{hyperref}
\usepackage[noend]{algpseudocode}
\usepackage{tabularx}
\usepackage{rotating}
\usepackage{multirow}
\usepackage[export]{adjustbox}
\definecolor{codegreen}{rgb}{0,0.6,0}
\definecolor{codegray}{rgb}{0.5,0.5,0.5}
\definecolor{codepurple}{rgb}{0.58,0,0.82}
\definecolor{backcolour}{rgb}{0.95,0.95,0.92}
\usepackage{bm}
\usepackage{verbatim}
\usepackage{subcaption}
\usepackage[left=2.5cm, right=2.5cm, top=4cm, bottom=3cm, footskip=0.5cm]{geometry}

\begin{document}

\begin{frontmatter}

\title{Designing Graph Convolutional Neural Networks for Discrete Choice with Network Effects}
\date{ }

\author[1]{Daniel F. Villarraga}
\ead{dv275@cornell.edu}
\author[1]{Ricardo A. Daziano}
\ead{daziano@cornell.edu}

\address[1]{School of Civil and Environmental Engineering, Cornell University, 220 Hollister Hall, Ithaca, NY 14853, USA}

\begin{abstract}
We introduce a novel model architecture that incorporates network effects into discrete choice problems, achieving higher predictive performance than standard discrete choice models while offering greater interpretability than general-purpose flexible model classes. Econometric discrete choice models aid in studying individual decision-making, where agents select the option with the highest reward from a discrete set of alternatives. Intuitively, the utility an individual derives from a particular choice depends on their personal preferences and characteristics, the attributes of the alternative, and the value their peers assign to that alternative or their previous choices. However, most applications ignore peer influence, and models that do consider peer or network effects often lack the flexibility and predictive performance of recently developed approaches to discrete choice, such as deep learning. We propose a novel graph convolutional neural network architecture to model network effects in discrete choices, achieving higher predictive performance than standard discrete choice models while retaining the interpretability necessary for inference—a quality often lacking in general-purpose deep learning architectures. We evaluate our architecture using revealed commuting choice data, extended with travel times and trip costs for each travel mode for work-related trips in New York City, as well as 2016 U.S. election data aggregated by county, to test its performance on datasets with highly imbalanced classes. Given the interpretability of our models, we can estimate relevant economic metrics, such as the value of travel time savings in New York City. Finally, we compare the predictive performance and behavioral insights from our architecture to those derived from traditional discrete choice and general-purpose deep learning models.
\end{abstract}


\end{frontmatter}

\section{Introduction}
\label{sec:intro}

Individuals tend to face choice situations where they have to select among a discrete set of alternatives, such as a mode of transportation for commuting to work, a candidate in political elections, or a new laptop to purchase. Econometric discrete choice models (DCMs) study these types of choice situations, assuming that individuals seek to maximize utility by selecting the most rewarding option. Typically, for standard discrete choice models, this choice depends on both the characteristics and preferences of the individual, and the attributes of each available option. Standard discrete choice models are interpretable and are used to extract microeconomic information, such as marginal rates of substitution (including willingness to pay for quality improvements), expected market shares, and probability marginal effects and choice elasticities with respect to particular attributes. However, model specification often requires deep domain knowledge, as the functional form of the utilities needs to be specified a priori, often with simple and restrictive forms—such as linear in inputs and parameters to represent compensatory behavior. Furthermore, predictive performance of standard discrete choice tends to be lower than that of other approaches, such as ensemble models and deep neural networks \cite{wang2021comparing}.\\

Although not the primary focus of research in discrete choice, there has been increased interest in modeling social influence and peer effects on decision-making. Intuitively, if a large number of an individual's friends purchase—or consider purchasing—a particular brand of cellphone, that individual may be inclined to do the same. In fact, decisions depend not only on personal preferences and the attributes of available options but also on how peers or social networks value those alternatives, a phenomenon known as network or peer effects. Nevertheless, discrete choice models often overlook network effects, and those that do account for them (e.g., \cite{goetzke2008network}, \cite{li2020estimating}, \cite{paez2008discrete}, \cite{brock2003multinomial}, \cite{dugundji2005discrete}, \cite{bhat2015new}) arguably lack the flexibility and predictive capabilities of other model classes, such as deep neural networks.\\

For instance, in \cite{goetzke2008network}, the author models binary transit choice in New York City using a linear specification for the latent utilities conditional on previous choices. This model depends on alternative attributes, socio-demographics, and mode shares in a previous time. In \cite{li2020estimating}, the authors model career decisions after high school using a linear specification for the latent utilities that depend on alternative attributes, socio-demographics, and network latent utilities as an endogenous term. A more comprehensive model is proposed in \cite{paez2008discrete}, where the authors model the adoption of new technology using a linear specification for the latent utility associated with each alternative at every time step. This includes, apart from individual characteristics and alternative attributes, a term associated with previously acquired knowledge, a function of past individual decisions, and two types of social influence: (i) from individuals who previously chose the alternative and (ii) from those who did not.\\

In more theoretical approaches, such as the one described in \cite{brock2003multinomial}, the authors introduce network effects by including the strength of social utilities and a subjective expectation per individual about the share of people who make each choice. An important idea reflected in the model presented by the authors is that social interactions reflect what people think about the behavior of their peers (reference group), not necessarily their actual behavior. In their model, the authors impose that subjective beliefs are equal to conditional objective choice probabilities, which they assert are equivalent to rational expectations.\\

More recently, in \cite{bhat2015new}, the authors examined the choice of the commute mode using a model that includes the spatial lag effect and spatially correlated error terms, similar to the SAL and SAE models from the spatial econometrics literature \cite{anselin2001spatial}. In addition, their model captures random taste variations with a spatially correlated structure. The most significant contribution of their research is that by allowing for taste variations to account for spatial correlations, the authors can account for residential self-selection. To our knowledge, their multinomial probit model is the most flexible and comprehensive specification that accounts for network effects in the current literature.\\

In all these models, social influence is incorporated into the latent utility as a weighted average of past or concurrent decisions, latent utilities, or expectations over reference group probabilities. This average is then multiplied by a parameter that reflects the overall significance of network effects. Furthermore, the specification of the latent utility is linear in terms of model parameters, alternative attributes, and socio-demographics. Therefore, to accommodate more complex forms of latent utility in these specifications, data pre-processing and feature engineering, expert knowledge, and behavioral assumptions are required, with the former requiring input from the latter two, which are not always available.\\

Deep learning (DL) models have recently been applied to discrete choice problems, often exhibiting higher predictive performance (e.g., \cite{wang2021comparing}, \cite{han2022neural}, \cite{lu2021modeling}) and more flexibility, as universal approximators with automatic feature learning, than traditional DCMs. However, deep learning models are often regarded as black-box approaches with high on-sample predictive performance but limited potential for inference and, in some cases, poor generalization. Although this assertion may hold for general-purpose architectures, models that incorporate expert knowledge and inference frameworks accounting for parameter instability could potentially offer interpretability comparable to standard DCMs.\\

Recently, there has been growing interest in the deep learning literature regarding graph-based methods applied to tabular data. In \cite{Tomlinson_Benson_2024}, the authors present a brief survey of some of the available graph methods (e.g., Graph Neural Networks, Laplacian Regularization) applied to discrete choice problems. Graph Neural Networks have shown strong predictive performance, particularly in recent architectures like GCNII, as presented in \cite{chen2020simple}, which leverage residual connections (as in ResNET \cite{he2016deep}) and identity mappings. These architectures achieve the highest predictive performance on some benchmark datasets, second only to label graph-based post-processing methods like the correct and smooth approach described in \cite{huang2020combining}.\\

Recent research has demonstrated that deep learning models can be used to extract economic information, such as marginal substitution rates, resulting in insights that are as comprehensive as those obtained from standard models \cite{wang2020deep}. Furthermore, it has been shown that model ensembles, models with inductive biases, and training procedures incorporating gradient regularization \cite{feng2024deep} can produce insights consistent with behavioral intuition. However, despite these advancements, current interpretable deep learning models have yet to incorporate network effects.\\

In this study, we design an interpretable Graph Neural Network (GNN) model for discrete choice analysis that incorporates network effects and can induce independence from irrelevant alternatives (IIA) in the multinomial setting. To test our model, we use a dataset with revealed mode preferences, drawing from the 2010/2011 Regional Household Travel Survey conducted by the New York Metropolitan Transportation Council \cite{authority2014north}. This dataset was extended with travel time and cost estimates obtained via the Google API, as well as social connection graphs created using trip origins and destinations from the Census Bureau \cite{censusdata}, mapped to the census tracts recorded in the travel survey. We also use data from the 2016 U.S. election aggregated by county, along with a social connectedness graph, as in the case study presented in \cite{Tomlinson_Benson_2024}, to evaluate the performance of our model on problems with high class imbalance.\\

We compare our model's predictive performance with that of traditional discrete choice models (DCMs), such as the standard logit model, and general-purpose deep learning architectures. Our findings demonstrate that our proposed architecture not only delivers behavioral insights that better align with intuition than those from off-the-shelf deep learning models, but also surpasses traditional DCMs in predictive performance. Furthermore, we implement approximate Bayesian inference using stochastic gradient Langevin dynamics (SGLD) \cite{welling2011bayesian} and demonstrate its potential to enhance the interpretability of general-purpose architectures within the context of our empirical example.\\

In the next section, \ref{DC}, we provide a brief overview of standard DCMs, and then, in Section \ref{NET_DC}, we discuss the generalities of the current DCMs that incorporate spatial or network effects. In Section \ref{DL_DC}, we outline some guiding principles for designing interpretable deep learning models for discrete choice. Following this, in Section \ref{GNNs}, we describe the GNN architecture that our model builds upon and that has gained popularity in the machine learning community. Then, in Section \ref{SKIP_GNN}, we introduce our interpretable GNN that accounts for social influence in discrete choice problems. Subsequently, we discuss our case studies in Sections \ref{NYC} and \ref{sec:US_elect}, and present the results from our model in Section \ref{results}. Finally, in Section \ref{conclusions}, we offer concluding remarks and discuss new research avenues that we consider highly promising within the Deep Learning for Discrete Choice subfield.\\

\section{Standard Discrete Choice Models \label{DC}}

In the discrete choice framework employed in this work, we assume that when faced with the decision of choosing between two alternatives (e.g., public transit or a private car for commuting to work), an individual \(i\) will base their decision \(y_i\) on the maximization of their own utility \(u_i\). This utility comprises a deterministic part that considers the individual's characteristics and the attributes of the alternatives (differences in the binary setting), as well as a random part that captures unobservables.\\

In a standard binary discrete choice setting, the latent utility \(u_i\) that an individual \(i\) derives from choosing an alternative of interest over the second one (e.g., transit over car in a binary mode choice context) is represented by equation \ref{eq:Utility}:
\begin{equation}
u_i = v_i + \epsilon_i,
\label{eq:Utility}
\end{equation}
where \(\epsilon_i\) represents the random component used to account for unobserved effects, and \(v_i\) is the deterministic part of the utility, as given by equation \ref{eq:Vut}:
\begin{equation}
v_i = \bm{x_i}^T \bm{\beta} + \bm{q_i}^T \bm{\gamma}.
\label{eq:Vut}
\end{equation}
Here, \(\bm{\beta}\) and \(\bm{\gamma}\) are estimable vectors of parameters, \(\bm{x_i}\) is a vector that represents the differences between the mode attributes (e.g., the difference in travel time), and \(\bm{q_i}\) is a vector of individual characteristics (e.g., household income). In basic binary choice models, the random component of the utility \(\epsilon_i\) is typically assumed to be independent and identically distributed (i.i.d.) across decision makers, following either a standard normal distribution (binary probit model) or a standard logistic distribution (logistic regression or binary logit model\footnote{A binary logit model specifies a utility function for each alternative, with i.i.d. EV1 error terms. The likelihood of the binary logit model depends on the difference of the utilities, leading to \(u_i\) as specified above, and the difference of the two error terms, which, following the properties of the EV1 distribution, is logistically distributed, making the model in differences equivalent to a logistic regression.}).\\

In this binary setting, the probability of choosing the alternative of interest (i.e., \(y_i = 1\), where \(y_i\) is a binary choice indicator) is given by equation \ref{eq:pstandard}:
\begin{equation}
P(y_i = 1) = P(\epsilon_i \le v_i),
\label{eq:pstandard}
\end{equation}
and the likelihood function for this model specification with \(n\) individuals is given by equation \ref{eq:lstandard}:
\begin{equation}
\mathcal{L}(\bm{\beta}, \bm{\gamma}) = \prod_{i=1}^n P(\epsilon_i \le v_i)^{y_i} \left(1-P(\epsilon_i \le v_i)\right)^{1-y_i}.
\label{eq:lstandard}
\end{equation}
For a more detailed and general description of standard discrete choice models, refer to \cite{train_2009}.\\

\section{Network and Peer Effects in Discrete Choice \label{NET_DC}}
The presence of network or peer effects in a discrete choice problem invalidates the i.i.d. assumption for the error terms in a standard discrete choice model. This violation leads to several issues, such as biased standard errors and, in cases where some covariates share the same correlation structure as the latent utility, even biased parameter estimates.\\

Network and peer effects have been extensively studied in spatial econometrics and, to some extent, in discrete choice analysis. In this section, we provide a brief review of the models in the literature that we consider to be the most widely adopted and foundational. However, this is not intended to be an exhaustive review.\\

\subsection{Autoregressive Models from Spatial Econometrics}

There are two general models that account for correlated observations applicable to both continuous and limited dependent variables \cite{anselin2013advances}, namely the Spatially Autoregressive Error and the Spatially Autoregressive Lag models. In this paper, we describe these approaches in the context of limited dependent variables, as encountered in discrete choice problems. The primary distinction between the two models lies in the assumption about the source of correlation in the continuous underlying index function (i.e., individual utilities in a discrete choice context). The Spatially Autoregressive Error (SAE) model assumes that correlation originates solely from correlated errors, whereas the Spatially Autoregressive Lag (SAL) model attributes autocorrelation to interactions between individual utilities. Both models result in a non-spherical variance-covariance structure capable of representing heteroskedasticity.\\

In the binary setting, the SAE model, which involves \(n\) individuals, \(k\) alternative attributes, and \(r\) socio-demographic variables, can be represented using the following matrix form for the latent utility, as shown in Equation \ref{eq:u_sae}:
\begin{equation}
\bm{u} = \bm{X\beta} + \bm{Q\gamma} + (\bm{I}-\rho \bm{W})^{-1}\bm{\epsilon},
\label{eq:u_sae}
\end{equation}
where \(\bm{X}\) is an \(n\times k\) matrix that contains alternative attribute differences for each individual, \(\bm{Q}\) is an \(n\times r\) matrix that represents individual characteristics, \(\bm{I}\) is an \(n \times n\) identity matrix, \(\rho\) is a scalar parameter that identifies the strength of correlated unobserved effects, \(\bm{W}\) is an \(n\times n\) adjacency or connectivity matrix that represents the structure of interactions or connections between individuals or entities in a network, and \(\bm{\epsilon}\) is an error vector that contains \(n\) uncorrelated elements. The adjacency matrix in peer effects econometrics is a foundational tool that encodes the structure of social or peer interactions, enabling researchers to quantify and analyze how individuals’ behaviors or outcomes are shaped by their peers. In spatial econometrics, where the focus is on geographic or spatial relationships between observations (e.g., regions or households), the matrix is commonly called the spatial adjacency matrix or simply the spatial weight matrix.

The SAL model, on the other hand, has the latent utility matrix specification shown in Equation \ref{eq:u_sal}:
\begin{equation}
\bm{u} = (\bm{I}-\rho \bm{W})^{-1}\bm{X\beta} + (\bm{I}-\rho \bm{W})^{-1}\bm{Q\gamma} + (\bm{I}-\rho \bm{W})^{-1}\bm{\epsilon}.
\label{eq:u_sal}
\end{equation}

Social influence or peer effects are usually modeled using a framework akin to that of the Spatially Autoregressive Lag (SAL) model. To illustrate this, consider the latent utility function for the SAL model written in the following form:
\begin{equation}
\bm{u} = \rho\bm{Wu} + \bm{X\beta} + \bm{Q\gamma} + \bm{\epsilon}.
\label{eq:u_sal_u}
\end{equation}

In this formulation, \(\rho\bm{Wu}\) represents the part of the latent utility influenced by peers, and \(\rho\) models the importance of peer effects. The term \(\bm{X\beta} + \bm{Q\gamma}\) encapsulates the deterministic part of the private latent utility, which accounts for individual preferences and characteristics. Lastly, \(\bm{\epsilon}\) denotes the uncorrelated error term, as in the SAE model.\\

\subsection{State-of-the-Art DCMs with Network Effects}

As previously discussed, the model proposed by \cite{bhat2015new} could be considered the most comprehensive specification for discrete choice models accounting for correlations between observations, which is a general Spatial Autoregressive with Autoregressive Disturbances (SARAR) model. This model works with panel data, models peer effects/social/network influence, and can account for self-selection effects by allowing for correlated random unobserved preference heterogeneity. A simplified binary specification of SARAR in matrix form, reminiscent of a more general form of the SAE and SAL models, follows the equation presented below:
\begin{equation}
    \bm{u} = \rho\bm{Wu} + \bm{X}\Bigl(b + (\bm{I}-\rho_{\beta}\bm{W})^{-1}\bm{\tau_{\beta}}\Bigr) + \bm{Q}\gamma + (\bm{I}-\rho_{\epsilon}\bm{W})^{-1}\bm{\epsilon}
\end{equation}

In this specification, the social part of the utility is modeled as in the SAL model, and the correlated unobserved effects are modeled as in the SAE model. However, there are a couple of terms not present in either the SAE or SAL models. These terms are those used to account for self-selection effects, namely:
\begin{equation}
    \beta = b + (\bm{I}-\rho_{\beta}\bm{W})^{-1}\bm{\tau_{\beta}}
\end{equation}
where \(\bm{\tau_{\beta}}\) is an uncorrelated random vector of size \(n\), \(b\) models the expected marginal utilities with respect to the alternative attributes, and \(\rho_{\beta}\) models the strength of self-selection based on attributes. With \(\bm{\tau_{\beta}}\) and \(\bm{\epsilon}\) as normally distributed random variables, this latent utility specification follows a normal distribution with a non-spherical variance-covariance matrix, as in the SAE and SAL models.\\

It is evident that restricting this SARAR specification allows us to recover either the SAE or SAL models, as well as the standard discrete choice model presented earlier. To illustrate this fact, setting \(\rho=0\) and \(\rho_{\beta}=0\) would yield the SAE model, and further setting \(\rho_{\epsilon}=0\) would yield the standard choice model discussed at the beginning of this paper.\\

This SARAR specification is fully transparent and interpretable; however, similar to standard discrete choice models, its specification limits the latent utilities to a linear function relative to alternative attributes and socio-demographics. Incorporating any additional complexity into the model would require expert knowledge as well as feature selection and engineering. Alternatively, we propose designing an interpretable graph neural network model. Such a strategy will enable us to retain the economic structure and information provided by the extant approaches while offering automatic feature learning and leveraging the predictive performance capabilities of deep learning.\\

\subsection{Brief Note on Convergent Matrices and Affine Systems}

In the specifications presented earlier, the term \((\bm{I} - \rho \bm{W})^{-1}\bm{z}\) (where \(\bm{z}\) is a vector) appears multiple times, warranting a more in-depth discussion. Suppose that the individual latent utilities—or other individual-specific variables—are continuously updated according to the neighbors' values and a constant term before the individuals decide which alternative to select. In discrete time, following the previously defined notation, this assumption can be represented as:
\begin{equation}
    \bm{u}(t + 1) = \rho \bm{W u}(t) + \bm{z}
\end{equation}

Assuming that \(\rho \bm{W}\) is a convergent matrix (which can be controlled by design), the sole equilibrium point for this system is given by:
\begin{equation}
    \bm{\bar{u}} = (\bm{I} - \rho \bm{W})^{-1}\bm{z}
\end{equation}

Additionally, under the same assumption regarding \(\rho \bm{W}\), the system reaches equilibrium in the limit:
\begin{equation}
    \lim_{t \to \infty} \bm{u}(t) = \bm{\bar{u}} = (\bm{I} - \rho \bm{W})^{-1}\bm{z}
\end{equation}

Therefore, \((\bm{I} - \rho \bm{W})^{-1}\bm{z}\) represents the fixed point reached in the limit of the system defined by \(\bm{u}(t + 1) = \rho \bm{W u}(t) + \bm{z}\). For instance, with \(\bm{z} = \bm{X\beta} + \bm{Q\gamma} + \bm{\epsilon}\), the affine system
\[
\bm{u}(t + 1) = \rho \bm{W u}(t) + \bm{z}
\]
reaches equilibrium at:
\begin{equation}
\bm{u} = (\bm{I}-\rho \bm{W})^{-1}\bm{X\beta} + (\bm{I}-\rho \bm{W})^{-1}\bm{Q\gamma} + (\bm{I}-\rho \bm{W})^{-1}\bm{\epsilon},
\end{equation}
which corresponds to the latent utility specification for the SAL model.\\

Although these facts might not seem immediately relevant, the reader will later find that they significantly clarify the application of Graph Neural Networks (GNNs) for discrete choice with social or network/peer effects.\\

\section{Deep Learning for Discrete Choice \label{DL_DC}}

The primary limitation of deep learning models applied to discrete choice is their inherent lack of interpretability right out of the box and their tendency to overfit the training set, leading to overall poor generalization. In discrete choice research, inference is often considered more important than prediction. Therefore, applying general-purpose architectures from deep learning, such as fully connected neural networks, to discrete choice problems without meaningful architectural adaptations can result in economic insights that are not useful—a fact that has discouraged the research community.\\

The lack of interpretability in deep learning models applied to discrete choice stems from three main challenges, namely: (i) model architectures that lack behavioral interpretation, (ii) unstable parameter estimates, and (iii) a lack of methods for representing epistemic uncertainty. The first challenge was highlighted by \cite{wang2020deep}, who demonstrated how to extract economic information as comprehensive as that provided by standard discrete choice models and illustrated methods for incorporating behavioral assumptions into model specifications through architectural design. The second challenge relates to the shape of the cost function for training when the model is underspecified by the data, as discussed by \cite{wilson2020bayesian}. Specifically, when the model architecture involves a large number of parameters but is constrained by limited data, the cost function tends to exhibit extensive valleys and multiple local minima. This cost function shape can result in various parameter estimates with equivalent predictive performance on the training data, but with multiple economic implications. The third challenge-related closely to the second one-is associated with the fact that, in most inference problems, researchers work with limited data, leading to uncertainty in the hypotheses derived from the model-fitting process \cite{wilson2020case}. To our knowledge, the applications of deep learning models to discrete choice have not addressed the representation of epistemic uncertainty, so deep learning applications to discrete choice have focused solely on point estimation.\\

In this paper, we primarily focus on addressing the first challenge by designing an interpretable deep learning architecture to model discrete choice problems with network or social effects, exploiting convolutional graph neural networks. Additionally, we explore methods to address unstable parameter estimates and represent epistemic uncertainty by effectively implementing approximate Bayesian inference. However, the second and third challenges will be addressed in full in future work.\\

\subsection{Stochastic Langevin Dynamics and Weight Averaging for Approximate Bayesian Inference}

Recent applications of deep learning (DL) to discrete choice modeling have demonstrated that the interpretability of these models can be enhanced through the use of model ensembles \cite{wang2020deep} for inference, rather than relying solely on a single set of model weights \(\bm{w}\). The effectiveness of this relatively straightforward technique stems from the fact that, for many problems, deep learning architectures tend to be underspecified by the training data. This underspecification leads to irregular loss landscapes characterized by large connected valleys and multiple modes \cite{izmailov2018averaging, li2018visualizing}. Therefore, for better interpretability, researchers are leaning towards approximate Bayesian approaches to deep learning, such as Stochastic Weight Averaging (SWA), Stochastic Weight Averaging Gaussian (SWAG) \cite{izmailov2018averaging}, and Stochastic Gradient Langevin Dynamics (SGLD) \cite{welling2011bayesian}.\\

Stochastic Weight Averaging (SWA) is a computationally efficient way to deal with unstable parameter estimates. SWA offers better generalization and more stable solutions than those obtained from standard training methods while potentially providing better behavioral insights. The SWA approach averages the weight iterates from stochastic gradient descent during the learning process. The algorithm introduces minimal computational and memory overhead since the weight average is computed as a running average every \(c\) gradient updates. A simplified version of the learning procedure, presented in \cite{izmailov2018averaging}, is shown below:

\begin{algorithm}
\caption{Stochastic Weight Averaging \cite{izmailov2018averaging}}
\begin{algorithmic}[1]
\Require initial weights \(\tilde{\bm{w}}\), cycle length \(c\), number of epochs \(e\), learning rate \(\alpha\), loss function \(L\)
\State \(\bm{w} \gets \tilde{\bm{w}}\) \Comment{Initialize weights with \(\tilde{\bm{w}}\)}
\State \(\bm{w}_{SWA} \gets \tilde{\bm{w}}\) \Comment{Initialize SWA weights}
\State \(\eta_{models} \gets 0\) \Comment{Initialize number of models in average}
\For{\(i \gets 1, 2, \dots, e\)}
    \State \(\bm{w} \gets \bm{w} - \alpha \nabla L(\bm{w})\) \Comment{Stochastic gradient update}
    \If{\(\mod(i, c) = 0\)}
        \State \(\bm{w}_{SWA} \gets \frac{\bm{w}_{SWA} \cdot \eta_{models} + \bm{w}}{\eta_{models} + 1}\) \Comment{Compute model average}
        \State \(\eta_{models} \gets \eta_{models} + 1\) \Comment{Increment the number of models in the average}
    \EndIf
\EndFor
\end{algorithmic}
\end{algorithm}

SWAG has a very similar implementation to SWA but differs in that it also computes the running weights' variance and uses these statistics to model weights as random variables drawn from a high-dimensional multivariate normal distribution. In that sense, SWA only provides a set of weights in a flatter region of the parameter space that has the potential for better generalization \cite{izmailov2018averaging} but remains a single point estimate, while SWAG gives a true approximation to the weights' posterior that can be used for Bayesian inference.\\

Similar to SWA and SWAG, SGLD uses iterates from stochastic gradient descent. It approximates Langevin dynamics to obtain an approximation of the weights' posterior distribution that can be treated as MCMC iterates and used for Bayesian inference. The algorithm works by injecting noise \(\bm{\eta}_t\) into the (batched) stochastic gradient updates, as described by equations \ref{eq:sgld_1} and \ref{eq:sgld_2}, and saving the weight iterates \(\bm{w}_t\). In those equations, \(q_{ti}\), \(x_{ti}\), and \(y_{ti}\) represent the socio-demographics, alternative attributes, and selected alternative for individual \(i\) in batch \(t\). \(p(\bm{w}_t)\) is the prior distribution for the weights (e.g., \(\ell_1\) or \(\ell_2\) regularization in deep learning frameworks), and \(p(q_{ti}, x_{ti}, y_{ti} \mid \bm{w}_t)\) is the likelihood of the observed individual under the set of weights \(\bm{w}_t\). In Equation \ref{eq:sgld_1}, \(N\) denotes the total number of training observations, \(n\) denotes the number of observations in the batch, and $\alpha_t$ denotes the gradient step size at $t$.

\begin{equation}
\Delta \bm{w}_t = \frac{\alpha_t}{2} \left( \nabla \log p(\bm{w}_t) + \frac{N}{n} \sum_{i=1}^n \nabla \log p(q_{ti}, x_{ti}, y_{ti} \mid \bm{w}_t) \right) + \bm{\eta}_t
\label{eq:sgld_1}
\end{equation}

\begin{equation}
\bm{\eta}_t \sim \mathcal{N}(0, \alpha_t)
\label{eq:sgld_2}
\end{equation}

With this gradient update structure, \(\bm{w}_t\) approaches samples from the posterior \(p(\bm{w} \mid q_{ti}, x_{ti}, y_{ti})\), as these updates approximate Langevin dynamics, which converge to the posterior distribution \cite{welling2011bayesian}. In contrast with SWAG, SGLD does not approximate the posterior mode using a Gaussian distribution and does not require additional post-estimation sampling.\\

In this paper, we implement SGLD to perform approximate Bayesian inference using our models. However, as pointed out earlier, an exhaustive analysis of Bayesian deep learning for discrete choice, including interval estimation and hypothesis testing, is beyond the scope of this paper. We will address this in future work focusing on epistemic uncertainty representation.\\

\section{One-size-fits-all Graph Convolutional Neural Networks \label{GNNs}}

In this study, we use a type of Graph Neural Network (GNN) known as Graph Convolutional Neural Networks (GCNs), as detailed in Kipf and Welling (2016) \cite{kipf2016semi}. GNNs are neural networks specifically designed to learn from graph-structured data, which includes tabular datasets with an underlying network structure. In the context of discrete choice, these network structures can be associated with social or geographic ties, as previously discussed. GCNs have been applied in various domains, including the analysis of physical systems, the prediction of protein interfaces, and the classification of diseases \cite{zhou2020graph}. In discrete choice problems, GCNs have been used by \cite{tomlinson2022graph} with standard architectures on benchmark datasets, with a focus on prediction capabilities.\\

To understand how GCNs are used within our discrete choice framework, consider the forward computation in a GCN layer, which is defined as:
\begin{equation}
    \bm{A}^{(l + 1)} = g^{(l+1)}\Bigl(\bm{W} \, \bm{A}^{(l)} \, \bm{\Theta}^{(l+1)}\Bigr)
\end{equation}
In this equation, \(\bm{A}^{(l)} \in \mathbb{R}^{n \times o}\) denotes the hidden representation at layer \(l\) for all observations in the choice dataset, where \(n\) is the number of observations and \(o\) is the size chosen for the hidden representation at layer \(l\). The activation function \(g^{(l)}\) is a non-linearity (such as ReLU, as proposed in \cite{kipf2016semi}), and \(\bm{\Theta}^{(l)} \in \mathbb{R}^{o \times p}\) is a matrix of learnable parameters that maps the hidden representations from \(\mathbb{R}^{o}\) to \(\mathbb{R}^{p}\). As before, the matrix \(\bm{W} \in \mathbb{R}^{n \times n}\) encodes the social connections or network structure.\\

If this operation is applied directly to the inputs of the neural network, the expression becomes:
\begin{equation}
    \bm{A}^{(1)} = g^{(1)}\Bigl(\bm{W} \, \text{CONCAT}(\bm{X}, \bm{Q}) \, \bm{\Theta}^{(1)}\Bigr).
\end{equation}
Here, we substitute \(\bm{A}^{(0)}\) with the operator \(\text{CONCAT}(\bm{X}, \bm{Q})\), which concatenates socio-demographics and alternative attributes for each observation. When used in the output layer \(L\), the computation then becomes:
\begin{equation}
    \bm{\hat{y}} = \sigma\Bigl(\bm{W} \, \bm{A}^{(L-1)} \, \bm{\Theta}^{(L)}\Bigr),
\end{equation}
where \(\bm{\hat{y}}\) is the vector of predictions and \(\sigma(\cdot)\) is the logistic function (for the binary case). Since GCNs can be applied across any layer of the network, one could design a model consisting entirely of graph convolutional operations applied from the input to the output layers. For instance, a two-layer GCN model using the ReLU activation function would be given by:
\begin{equation}
    \bm{\hat{y}} = \sigma\Bigl(\bm{W} \, \text{ReLU}\Bigl(\bm{W} \, \text{CONCAT}(\bm{X}, \bm{Q}) \, \bm{\Theta}^{(1)}\Bigr) \, \bm{\Theta}^{(2)}\Bigr).
\end{equation}
This entire function is almost everywhere differentiable, allowing the model to provide economic insights comparable to those derived from standard discrete choice models. However, it is important to note that these insights may not always be behaviorally reasonable (e.g., not respecting monotonicity). Note that the network structure captured by \(\bm{W}\) is considered for each individual's choice prediction, and the model is no longer linear in inputs or parameters.\\

This GCN architecture is generic and could be applied to any binary classification problem with an underlying network structure. In this setup, network effects are intertwined with the private component of the deterministic utility (i.e., \(\bm{X\beta} + \bm{Q\gamma}\) in a linear utility model), and the architecture does not directly enforce specific behavioral assumptions. In the following section, we build upon this foundation by designing a model specifically tailored for discrete choice, incorporating architectural design choices that are supported by the Machine Learning literature and ensure better behavioral interpretability.\\

\section{GCNs Tailored for Discrete Choice: Skip-GNN \label{SKIP_GNN}}

The model presented in the previous section represented the latent utility as:

\begin{equation}
    \bm{u} = \bm{W} \bm{A^{(L-1)}} \bm{\Theta^{(L)}},
\end{equation}
where $\bm{A^{(L-1)}} = f(\bm{X}, \bm{Q})$ is a non-linear function of socio-demographics and alternative attributes. We will depart from this general formulation of the latent utilities for our proposed architecture, integrating meaningful design choices based on two inductive biases that have proven empirically useful for estimation in the machine learning literature, namely: skip connections and batch normalization. \\

On the one hand, skip connections\footnote{In this paper, the term 'skip connections' is used to encompass both residual and skip connections. In both cases, information from previous layers is propagated forward in the network and is either concatenated or summed with the inputs of subsequent layers. These terms are often used interchangeably in the literature, their advantages during training and in terms of generalization are equivalent, and the distinction between them is not particularly relevant for the purposes of this paper.} \cite{he2016deep} have been shown to be necessary for training deep learning models that are under-specified by the data. It was demonstrated in Li et al. (2018) \cite{li2018visualizing} that architectures with skip connections have a dramatic effect on the loss landscape. The authors provide visualizations on random directions of the parameter space for different architectures and illustrate how skip connections prevent the loss landscape from exhibiting problematic levels of non-convexity. Their study elucidates the positive impact of skip connections on training speed and generalization, as observed empirically.  \\

On the other hand, batch normalization, developed by Ioffe and Szegedy \cite{ioffe2015batch}, was introduced to account for covariate shift (when the input distribution of a learning system changes) by normalizing the inputs to certain layers of deep networks trained with random batches of data. The inclusion of batch normalization in deep learning architectures has been shown to speed up the training process while having state-of-the-art performance --or better if combined with model ensembles \cite{ioffe2015batch}.\\

In our architecture, we will use skip connections to distinguish between the linear and non-linear parts of the latent utilities, as well as between private and social utilities (the latter influenced by peers). Batch normalization will be implemented in one of the last layers of the model as a means to control the general scale and location of the latent utilities. In addition to incorporating behavioral insights, these inductive biases have the potential to produce model architectures that are easier to train. After training, these models are also likely to converge to generalize better. We name our architecture Skip-GNN because it relies heavily on skip and residual connections.


\subsection{Linear and Non-Linear Private Utility Components}

The first assumption we make for our model, without any loss of generality, is the existence of an underlying private component in the latent utility that includes both linear and non-linear parts, as in a semi-parametric specification:
\begin{equation}
    \bm{u_{pr}} = \bm{X\beta} + \bm{Q\gamma} + f(\bm{X}, \bm{Q}),
    \label{eq:private}
\end{equation}
where $f(\cdot)$ denotes a non-linear function, such as a fully connected neural network. This specific structure of the latent utilities can be efficiently implemented by employing skip connections from the input layer to the final layer of a deep learning architecture.

\subsection{Private and Social Utilities}

Next, to incorporate network effects in the model, consider the following system:
\begin{equation}
\begin{split}
    \bm{a^{(l+1)}} &= \bm{W}\bm{a^{(l)}} \theta^{(l)} + \bm{u_{pr}} \\ & = \bm{W}\bm{a^{(l)}} \theta^{(l)} + \bm{X\beta} + \bm{Q\gamma} + f(\bm{X}, \bm{Q}).
\end{split}
\end{equation}

This system could be modeled using a Graph Convolutional Neural Network with one-dimensional latent representations $\bm{a^{(l)}} \in \mathbb{R}^{n \times 1}$, learnable parameters $\theta^{(l)} \in \mathbb{R}$, linear activation functions, and skip connections from the private utilities to each graph convolutional layer. \\

With a large number of layers $L$ and setting $\theta^{(l)} = \rho, \forall l$ (such that $\bm{W}\bm{a^{(l)}} \theta^{(l)} = \rho \bm{W}\bm{a^{(l)}}$), and following the discussion on convergent matrices and affine systems from section \ref{NET_DC}, we can recover the following version of the SAL model:

\begin{equation}
    \bm{u} = \bm{a^{L}} = (\bm{I}-\rho \bm{W})^{-1} \bm{X\beta} + (\bm{I}-\rho \bm{W})^{-1} \bm{Q\gamma} + (\bm{I} - \rho \bm{W})^{-1} f(\bm{X}, \bm{Q}).
\end{equation}

This specification would then incorporate a non-linear term that depends on alternative attributes and socio-demographics in the latent utility representation instead of the random term $\bm{\epsilon}$ from the original SAL model. Again, this structure for the latent utility can be conveniently implemented using skip connections. \\

Allowing the GCN to have non-linear mappings $g^{(l)}(\cdot)$, and for high-dimensional hidden representations $\bm{A^{(l-1)}}$, we would have:

\begin{equation}
   \bm{A^{(l)}} = g^{(l)}(\bm{W} \bm{A^{(l-1)}} \bm{\theta^{(l)}} + \bm{X\beta} + \bm{Q\gamma} + f(\bm{X}, \bm{Q})),
\end{equation}
leading to the following latent utility model:
\begin{equation}
    \bm{u} = \bm{W A^{(L-1)}} \bm{\theta^{(L-1)}}  + \bm{X\beta} + \bm{Q\gamma} + f(\bm{X}, \bm{Q}).
\end{equation}

The operations in Graph Convolutional layers can be interpreted as a social process where individuals update their utilities (represented as a latent high-dimensional embedding) in discrete time. These representations, which may be high-dimensional, are based on their neighbors' representations and the parameters $\bm{\theta^{(l)}}$. In contrast to the SAL model and the SARAR model proposed by Bhat (2015) \cite{bhat2015new}, the significance of peer effects is not encapsulated in a single parameter $\rho$, but rather in the set of parameters $\bm{\theta^{(l)}}$ associated with the GCN. \\

Additionally, this model can capture a form of exogenous interaction effects in the latent utilities, akin to those in the spatial Durbin model (see LeSage and Pace, 2009 \cite{lesage2009introduction}), by setting:
\begin{equation}
    \bm{A^{(0)}} = \text{CONCAT}(\bm{X}, \bm{Q}),
\end{equation}
and,
\begin{equation}
   \bm{A^{(l)}} = \text{CONCAT}(g^{(l)}(\bm{W} \bm{A^{(l-1)}} \bm{\theta^{(l)}} + \bm{u_{pr}}), \bm{X}, \bm{Q}),
\end{equation}
so that the operation $\bm{W} \bm{A^{(l-1)}} \bm{\theta^{(l)}}$ has terms of the form $\bm{W} \bm{X} \bm{\theta_X^{(l)}}$ and $\bm{W} \bm{Q} \bm{\theta_Q^{(l)}}$ that capture a form of exogenous interaction effects, as conceptualized in the spatial Durbin model. The learnable parameters are $\bm{\theta}^{(l)} \in \mathbb{R}^{1 + k + r}$ for layers $l>1$, and  $\bm{\theta}^{(l)} \in \mathbb{R}^{k + r}$ for $l = 1$ (with $k$ equal to the number of alternative attributes and $r$ equal to the number of socio-demographics).

\subsection{Setting the General Scale of Utilities Through Batch Normalization}

Finally, batch normalization is employed in the non-linear part of the private utility $f(\bm{X}, \bm{Q})$. This means that, during training, batch statistics are computed for $f(\bm{X}, \bm{Q})$ to normalize their value according to the algorithm \ref{algo:batch} presented below\footnote{In practice, the output $\hat{z}_i$ is re-scaled by two learned parameters. In our context, this re-scaling operation would not determine the location or scale of the utilities. For our purposes, we ignore that re-scaling operation so that the batch normalization layer does not have any learnable parameters.}. At prediction time, the whole sample statistics are used instead. 
\begin{algorithm}
\caption{BatchNorm: Batch Normalizing Transform, applied to activation \(z\) over a mini-batch. \cite{ioffe2015batch}}
\begin{algorithmic}
\State \textbf{Input:} Values of \(z\) over a mini-batch: \(B = \{z_{1}, \ldots, z_{m}\}\)
\State \textbf{Output:} \(\{\hat{z}_i = \text{BN}_{\gamma,\beta}(z_i)\}\)
\State
\State \(\mu_B \gets \frac{1}{m} \sum_{i=1}^{m} z_i \) \Comment{mini-batch mean}
\State \(\sigma_B^2 \gets \frac{1}{m} \sum_{i=1}^{m} (z_i - \mu_B)^2 \) \Comment{mini-batch variance}
\State \(\hat{z}_i \gets \frac{z_i - \mu_B}{\sqrt{\sigma_B^2 + \epsilon}} \) \Comment{normalize}
\end{algorithmic}
\label{algo:batch}
\end{algorithm}

This operation is analogous to setting the location and general scale for the utilities in the standard logit and probit models \cite{train_2009}. With these design decisions, the general GCN model for discrete choice implemented in this paper comprises three main blocks: i) the linear part of the private utilities, ii) a fully-connected neural network $f(\bm{X},\bm{Q})$ with normalized outputs representing the non-linear part associated with the private utilities, and iii) a series of GCN layers used to model network effects and exogenous interaction effects. The entire model structure is summarized as follows: 

\begin{equation}
    \bm{u_{pr}} = \bm{X\beta} + \bm{Q\gamma} + \text{BatchNorm}(f(\bm{X}, \bm{Q}))
    \label{eq:private_2}
\end{equation}

\begin{equation}
    \bm{A^{(0)}} = \text{CONCAT}(\bm{X}, \bm{Q})
\end{equation}

\begin{equation}
   \bm{A^{(l)}} = \text{CONCAT}(g^{(l)}(\bm{W} \bm{A^{(l-1)}} \bm{\theta^{(l)}} + \bm{u_{pr}}), \bm{X}, \bm{Q})
\end{equation}

\begin{equation}
  \bm{u} = \bm{a^{L}} = \bm{W A^{(L-1)}} \bm{\theta^{(L)}}  + \bm{u_{pr}}
\end{equation}

\begin{equation}
  \bm{\hat{y}} = \bm{p} = \sigma(\bm{u}),
\end{equation}
where $\bm{\theta^{l}} \in \mathbb{R}^{1 + k + r}$ for all GCN layers $l>1$ (with $k$ equal to the number of alternative attributes and $r$ equal to the number of socio-demographics). \\

\begin{figure}[h!]
    \centering
    \includegraphics[width=\linewidth]{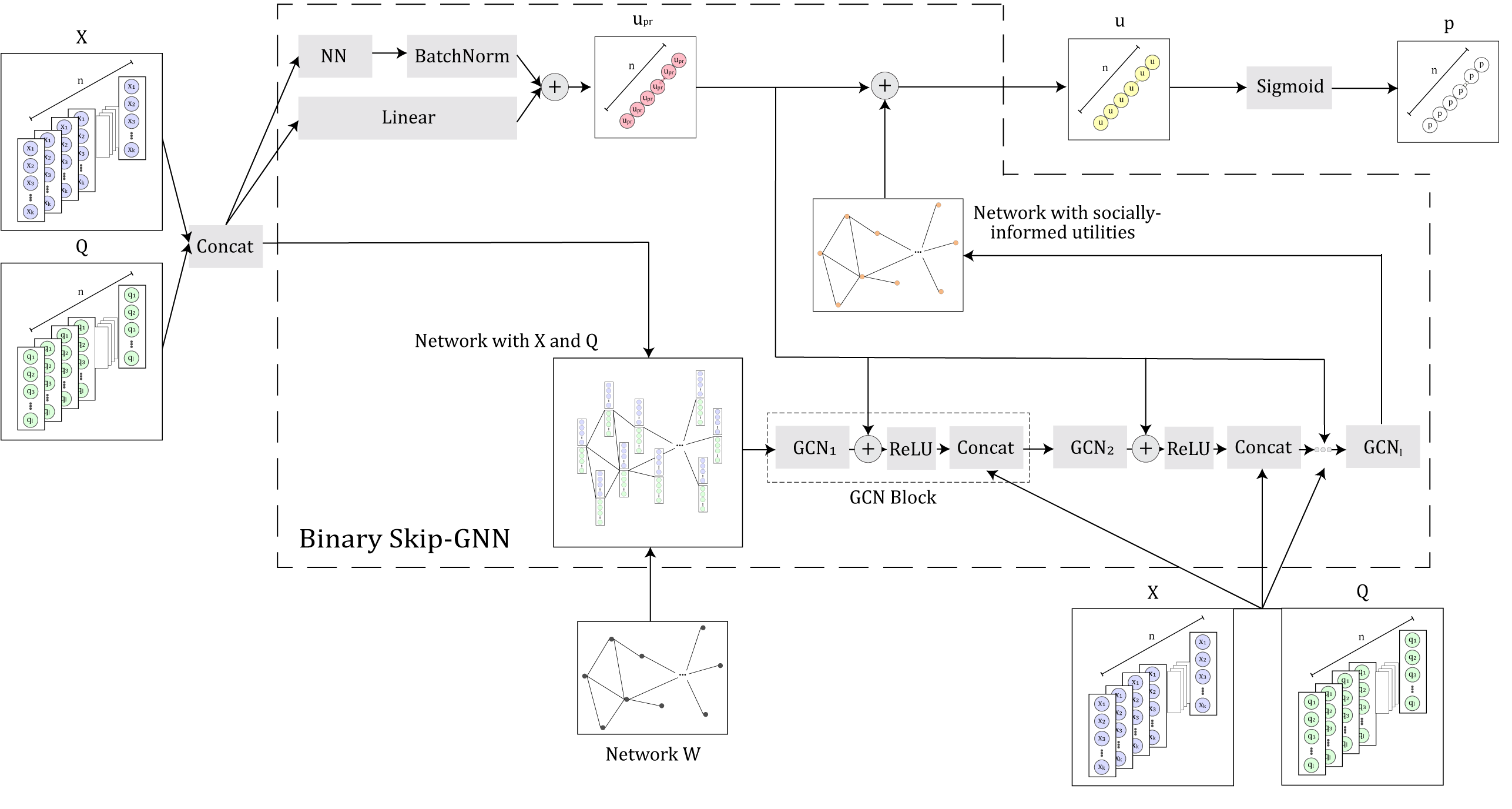}
    \caption{Binary Skip-GNN model architecture.}
    \label{fig:skip_gnn_binary}
\end{figure}

The binary model architecture is illustrated in Figure \ref{fig:skip_gnn_binary}. In this figure, the private utilities are computed by passing the alternative attributes and socio-demographics ($\text{CONCAT}(X, Q)$) through a fully connected neural network (NN) with normalized outputs (BatchNorm) and, in parallel, a linear layer (Linear), and summing their outputs. The GCN blocks, up to $L-1$, contain four sequential operations: i) $\text{GCN}_l$ represents the $\bm{W} \bm{A^{(l-1)}} \bm{\Theta^{(l)}}$ operation, ii) the output from $\text{GCN}_l$ is summed with the private utilities $\bm{u_{pr}}$, iii) the result is passed through a ReLU non-linearity operator, and iv) the output from the activation is concatenated (CONCAT) with $X$ and $Q$. The last GCN operation ($\text{GCN}_L$) outputs the socially-informed part of the utility. To obtain the latent utilities $\bm{u}$, the socially informed utilities are summed with the private utilities. Finally, to generate probability predictions $\bm{p}$ in the binary problem setting, the latent utilities are passed through a sigmoid activation function. The part of the model within the dashed frame will be referred to here as a Binary Skip-GNN block, which is convenient for illustrating our model in the multinomial setting. \\

\subsection{Multinomial Extension, IIA, and Shared Alternative Parameters}

Our architecture can be extended for scenarios involving multiple alternatives. However, in such multinomial cases, a decision must be made regarding whether to ensure independence from irrelevant alternatives (IIA), as in \cite{wang2020deep}, or allowing the model to actually learn substitution patters from the data. In this subsection, we discuss both options. \\

The more straightforward case involves the model that ignores the Independence of Irrelevant Alternatives (IIA). In such cases, similar to standard discrete choice models, it is necessary to ensure that socio-demographic variables are incorporated into the utility functions for at most $J-1$ alternatives (where $J$ represents the total number of alternatives considered). If this is not ensured, the shape of the loss-function will include additional modes, affecting the stability of the parameters. Econometrically, the model would not be identified. Another important consideration pertains to non-linearity in the output layer. For the multinomial case, it is necessary to replace the Sigmoid function with a Softmax function to have an output size equal to the number of alternatives $J$, such that a full set of choice probabilities $P_{ij}$ is produced. Other architectural adjustments are merely dimensional changes to accommodate $J$ alternatives. \\

The more complex case would be ensuring the Independence of Irrelevant Alternatives (IIA). Whereas IIA is a reasonable axiom of preferences in certain theoretical contexts, it becomes problematic in logit-type models because IIA imposes restrictive and often unrealistic substitution patterns. However, IIA provides a baseline for checking regularity and ensures that predictions do not deviate too far from established economic theory, making it a useful tool for model validation and interpretability.   One way to impose IIA employs 1D convolutions that effectively create independent alternative model blocks with shared parameters, as demonstrated in \cite{wang2020deep} and \cite{arkoudi2023combining}. With these types of architectures, the original socio-demographics (common to all alternatives) cannot directly influence the alternative-specific utilities from the first layer, since the associated parameters are actually shared and thus rendered irrelevant. To address this issue, the socio-demographics should go through another model (e.g., a fully connected neural network) before entering the alternative-specific utilities after a predetermined tunable number of layers. This model is used to create a socio-demographic embedding of size $J\times K$, so that the utilities for each alternative depend on $K$ socio-demographic related features. This approach mirrors the one proposed in \cite{arkoudi2023combining}, with the key distinction being that the socio-demographic embeddings are not restricted to entering the utility function at the final layer. This additional flexibility allows for models that incorporate interactions between socio-demographics and alternative attributes.\\

\begin{figure}[h!]
    \centering
    \includegraphics[width=0.9\linewidth]{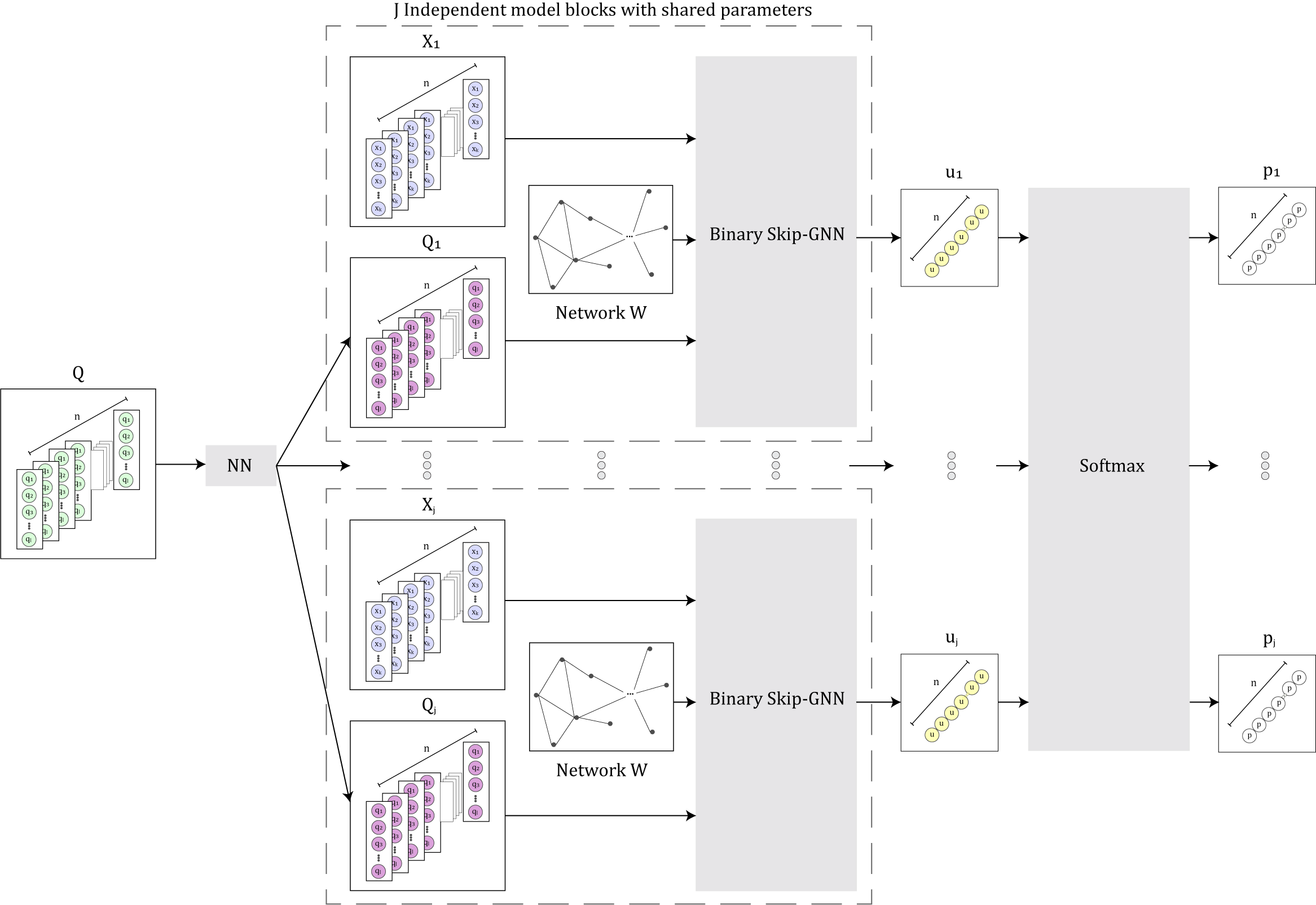}
    \caption{Multinomial Skip-GNN with IIA model architecture.}
    \label{fig:skip_gnn_multi}
\end{figure}

The multinomial model architecture with IIA is depicted in Figure \ref{fig:skip_gnn_multi}. As shown in that figure, the socio-demographics $Q$ pass through a fully connected neural network (NN) that outputs $J$ new socio-demographic vector representations $Q_1, \dots, Q_J$, one per alternative. The model has $J$ binary skip-GNN blocks with shared parameters. The utilities $\bm{u}$ output by the $J$ binary skip-GNNs go through a Softmax function to compute probability predictions $p$ for each alternative.

\section{Mode Choice in New York City \label{NYC}}

We use the 2010/2011 Regional Household Travel Survey by the New York Metropolitan Transportation Council \cite{authority2014north} to build mode choice data in New York City. The survey contains travel data from 18,965 households from areas in New York City, Long Island, the Hudson Valley, New Jersey, and Connecticut. The data is stored in a relational database with information on travel and mobility patterns, including socio-demographics, trip origins and destinations, and mode choices. 

\subsection{Binary Mode Choice: Public Transit vs Private Car}

Following the reasoning presented in \cite{goetzke2008network}, we limit our analysis to the New York City areas. Additionally, we selected the home-based work (HBW) trips completed in a private car or transit longer than 1.5 miles, and we discarded observations with missing data. After the data selection and cleaning process, 2,444 trips are left for our analysis.  \\

We use the US census block data \cite{censusdata} to obtain the longitude and latitude of origins and destinations for every trip. We used the Google API to retrieve estimates for each trip's cost and travel time for various modes of transportation. This effort results in a novel binary mode choice dataset for NYC, containing trip origin and destination coordinates, trip costs, travel times, and socio-demographic variables (as presented in \cite{VILLARRAGA2025103132}). For our models, we consider the following variables: i) trip cost difference (transit vs. car), ii) travel time difference (transit vs. car), iii) an indicator for access to a private car in the household, iv) an indicator for the destination being in Manhattan, v) an indicator for high-income level, and vi) an indicator for declared gender. The description of these variables along with their mean values are presented in Table \ref{tab:summary_binary}.  \\

\begin{table}[h!]
	\centering 
	\caption{Summary of the variables considered for the binary mode choice problem.}
	\begin{tabular}{p{5cm}p{8cm}S[table-format=3.2]}
		\hline 
		\centering Variable & \centering  Description & \multicolumn{1}{c}{Mean}  \\
		\hline
		\centering Trip cost difference &  Transit cost minus private car cost (USD). & -3.36  \\
		\centering Trip time difference &  Transit travel time minus private car travel time (Minutes). & 35.89\\
		\centering Vehicle availability &  Indicator variable for car availability in the household. It takes the value of 1 if no cars are available in the household. & 0.34 \\
		\centering High income &  Indicator variable for high income level ($>100$k USD per year). & 0.33 \\
		\centering Manhattan &  Indicator variable for destinations in Manhattan. &  0.47 \\
		\centering Gender &  Indicator variable for the male gender. & 0.48  \\
		\centering \textbf{Mode choice ($y$)} &  Whether transit was selected as the travel mode. A choice indicator that takes the value of 1 if transit was selected. & 0.61 \\
		\hline
	\end{tabular}
	\label{tab:summary_binary}
\end{table}

\begin{figure}[h!]
\centering
\begin{minipage}[c]{0.47\linewidth}
	\includegraphics[scale=0.5, right]{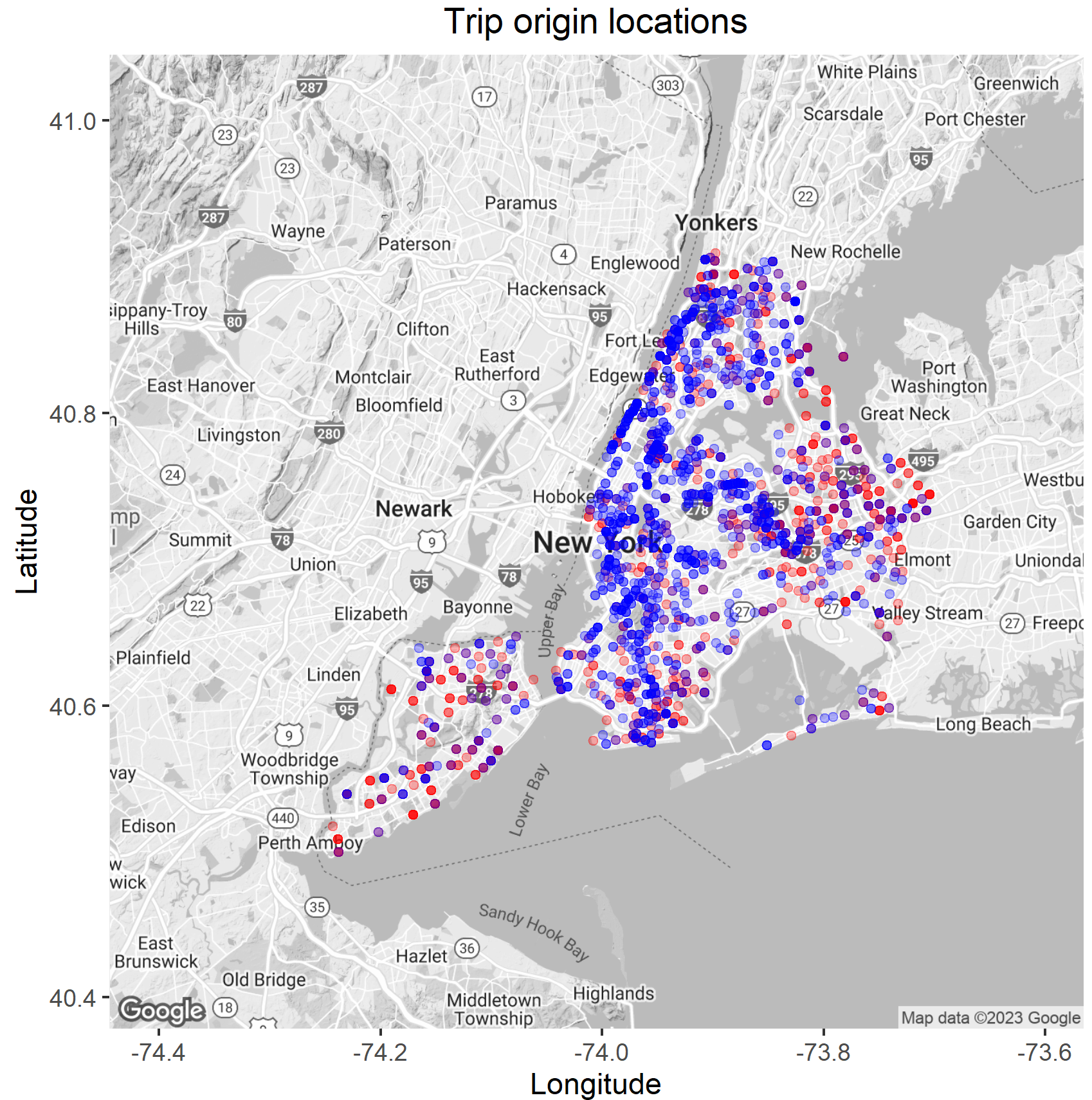}
	\caption{Trip origins and mode choice in New York City.}
	\label{fig:o_mc}
\end{minipage}\hfill
\begin{minipage}[c]{0.47\linewidth}
	\includegraphics[scale=0.5, left]{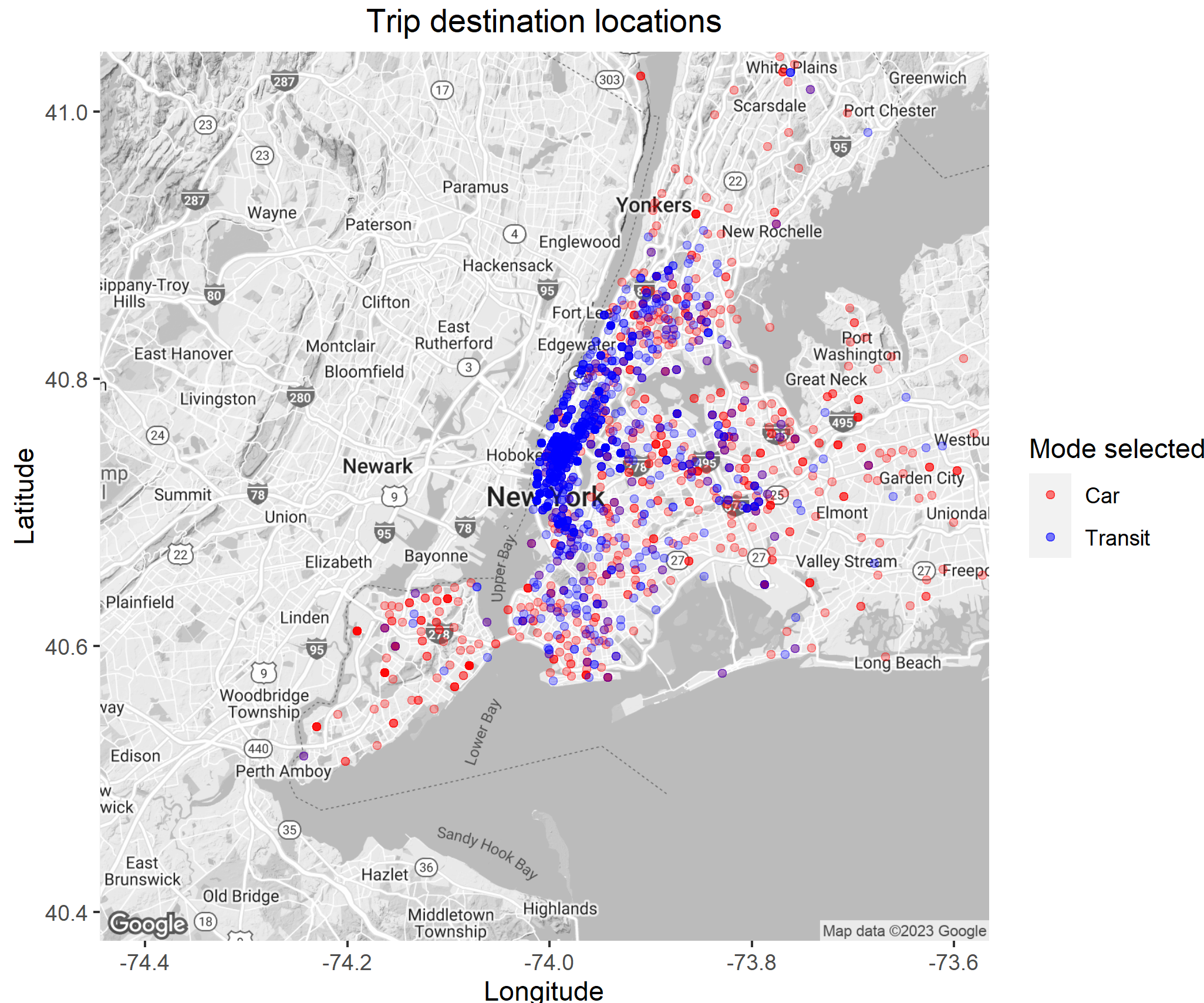}
	\caption{Trip destinations and mode choice in New York City.}
	\label{fig:d_mc}
\end{minipage}
\end{figure}

Figures \ref{fig:o_mc} and \ref{fig:d_mc} show the origin and destination locations (i.e. households and work locations), respectively. The red points represent the trips for which the individual selected a private car, and the blue points represent the ones for which the individual chose public transit. As shown in these plots, it is clear that there might be unobserved spatial effects that influence mode choice in New York City. For instance, for trips ending in Manhattan, individuals seem more likely to choose public transit over private cars. \\

\begin{figure}[h!]
	\begin{minipage}[c]{0.47\linewidth}
		\centering 
		\includegraphics[scale=0.55, right]{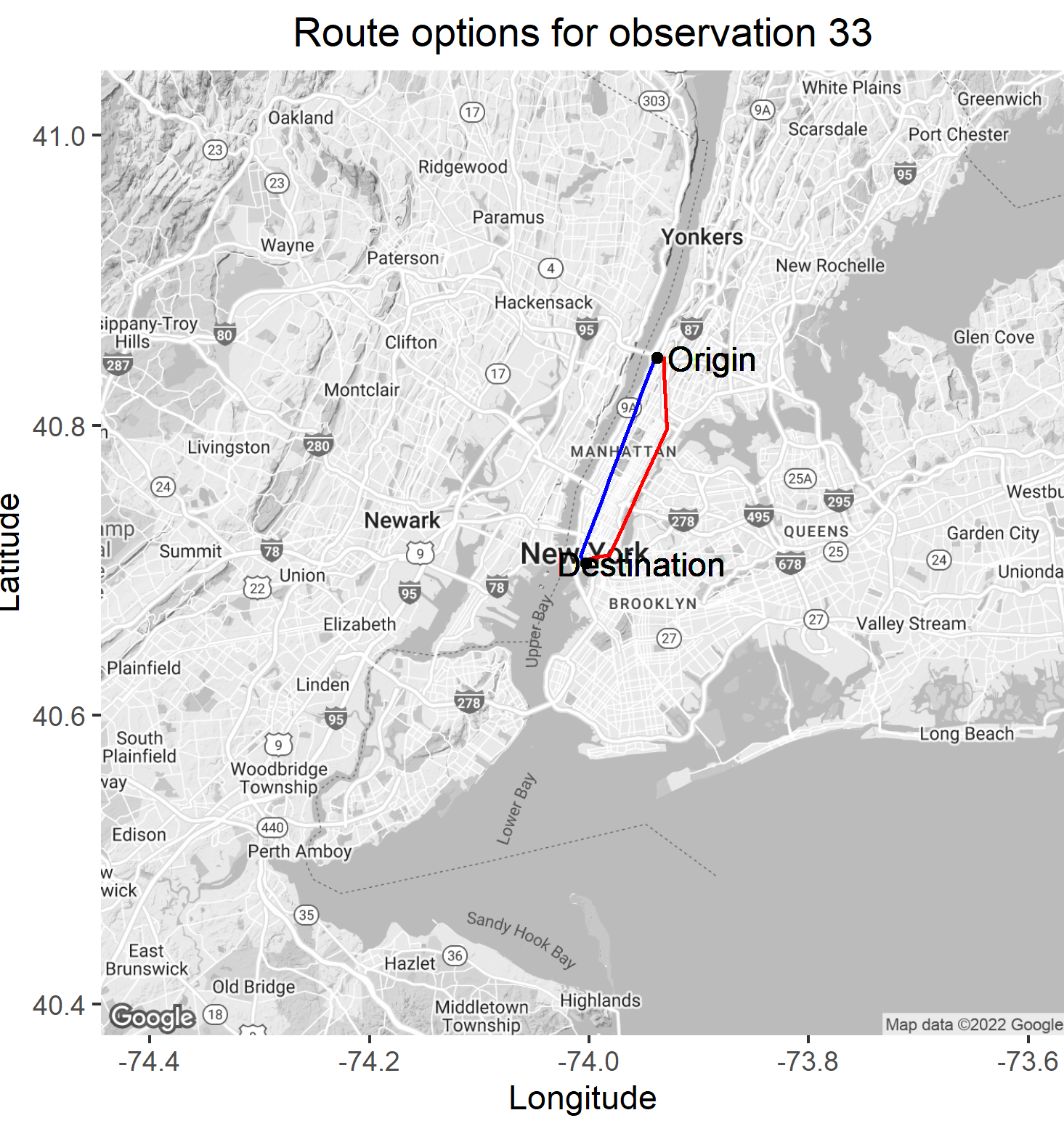}
		\caption{Trip 33 alternative routes for public transit and private car. For this trip, the cost of using a private car is \$4.75 higher than the cost of using public transportation.}
		\label{fig:r_33}
	\end{minipage}\hfill
	\begin{minipage}[c]{0.47\linewidth}
		\centering 
		\includegraphics[scale=0.55, left]{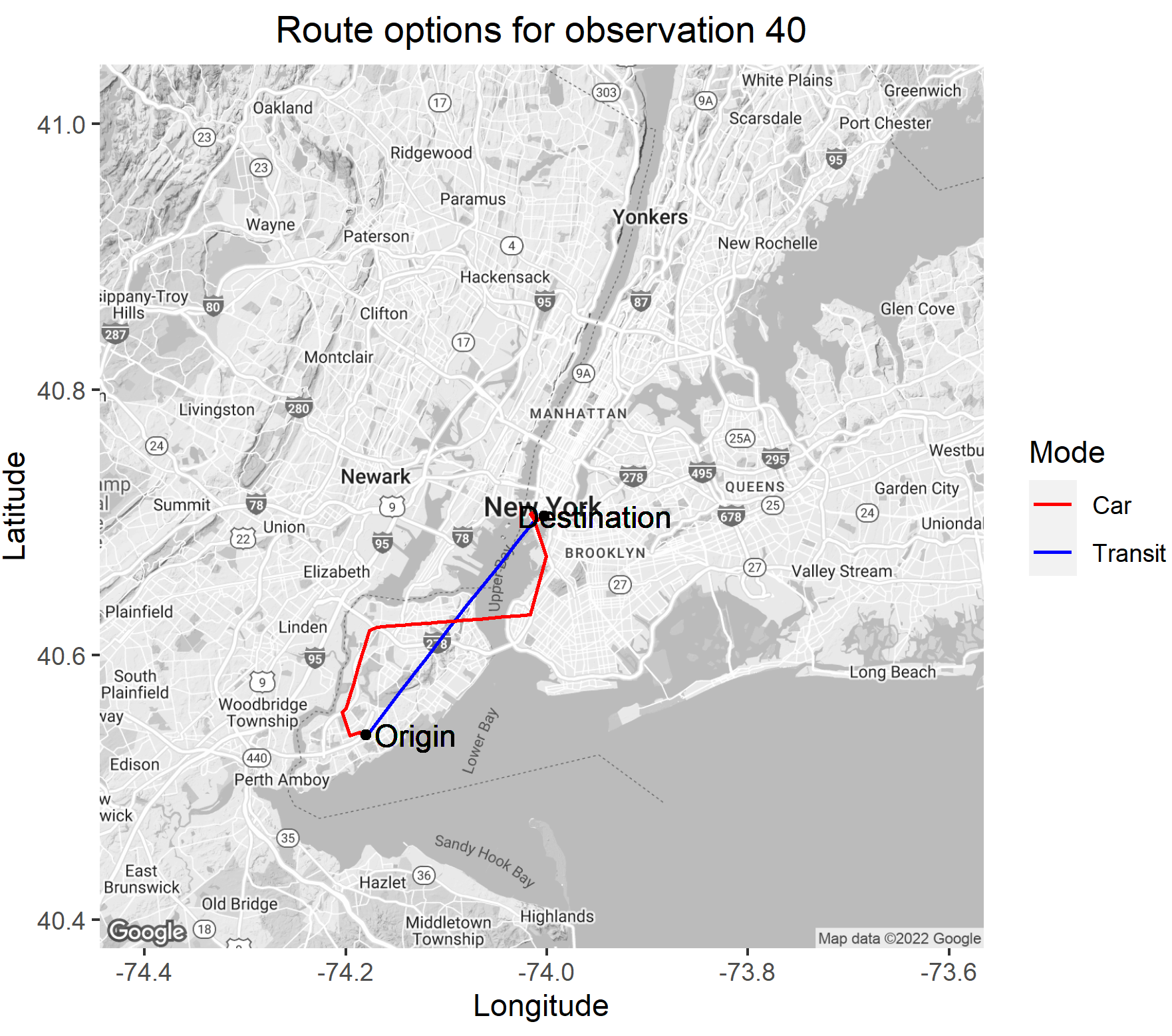}
		\caption{Trip 40 alternative routes for public transit and private car. For this trip, the cost of using a private car is \$11.77 higher than the cost of using public transportation.}
		\label{fig:r_40}
	\end{minipage}
\end{figure}

\begin{figure}[h!]
	\begin{minipage}[c]{0.47\linewidth}
		\centering 
		\includegraphics[scale=0.55, right]{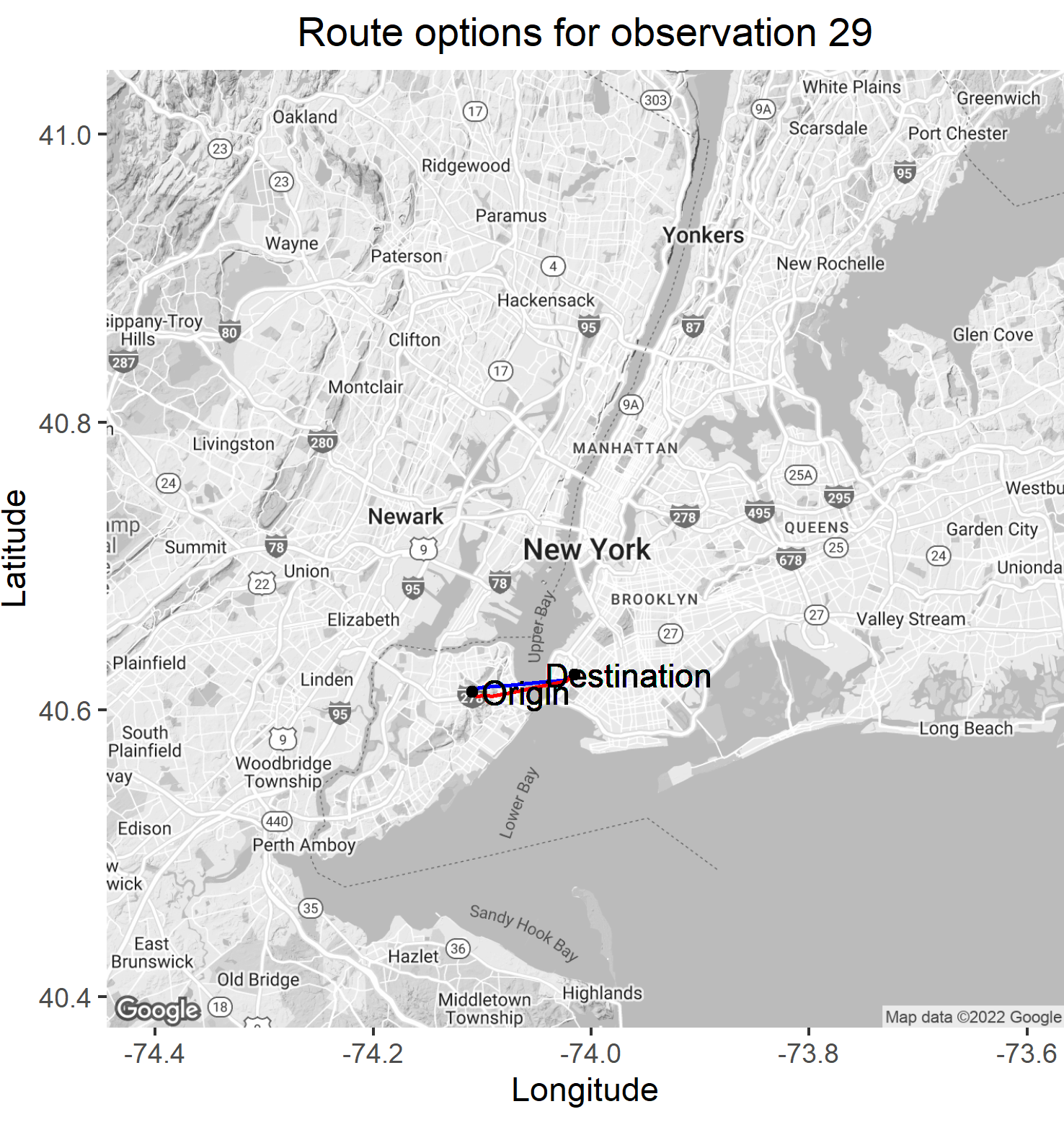}
		\caption{Trip 29 alternative routes for public transit and private car. For this trip, the cost of using a private car is \$1.06 higher than the cost of using public transportation.}
		\label{fig:r_29}
	\end{minipage}\hfill
	\begin{minipage}[c]{0.47\linewidth}
		\centering 
		\includegraphics[scale=0.55, left]{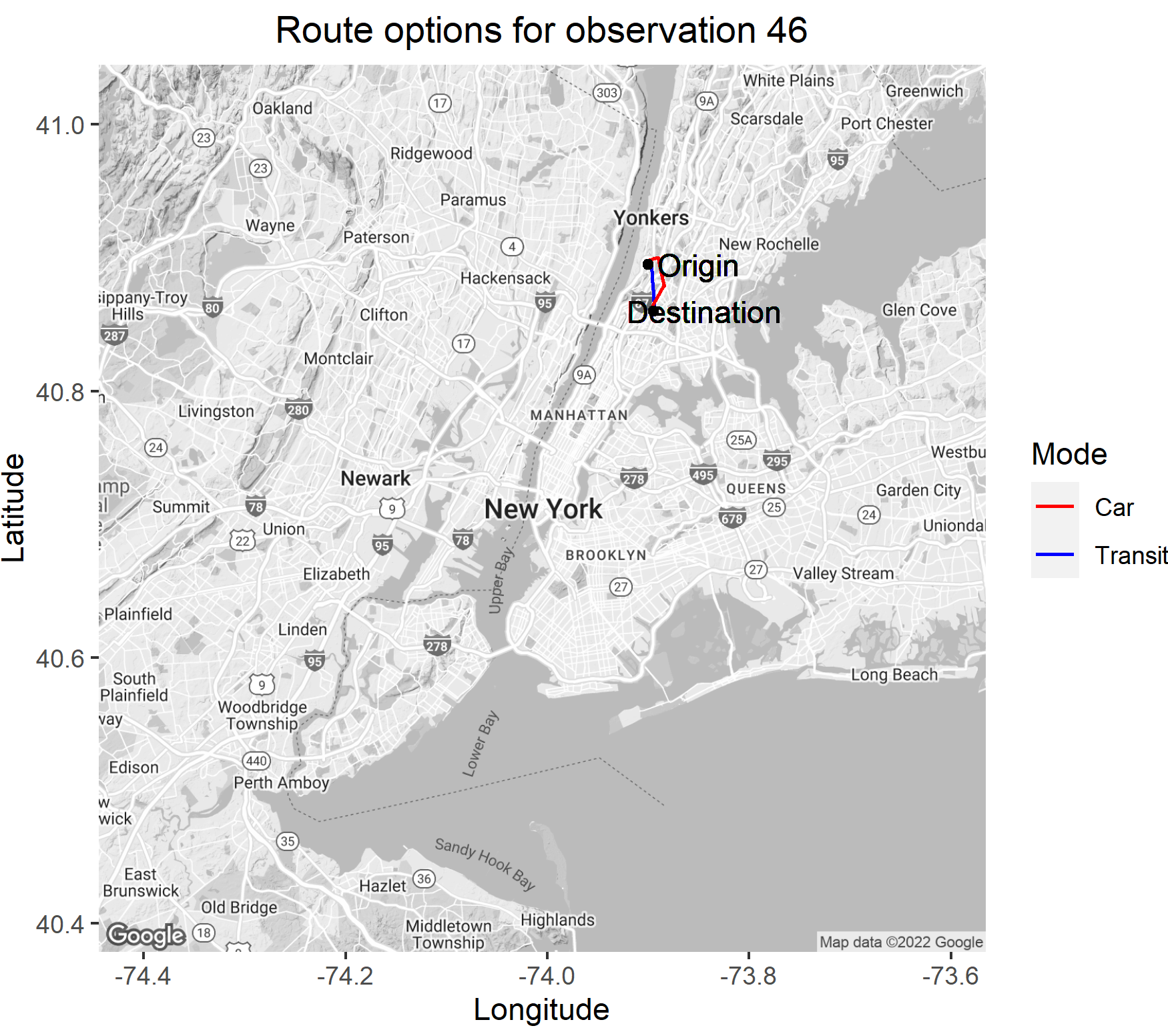}
		\caption{Trip 46 alternative routes for public transit and private car. For this trip, the cost of using a private car is \$0.10 higher than the cost of using public transportation.}
		\label{fig:r_46}
	\end{minipage}
\end{figure}

In figures \ref{fig:r_33} and \ref{fig:r_40}, we show trip examples for which public transit was selected, and in figures \ref{fig:r_29} and \ref{fig:r_46}, we show trip examples for which private car was selected. The paths in blue correspond to the public transit routes, and the paths in red are for car. We show these figures to illustrate that mode choice is influenced by the alternative route characteristics, particularly the cost and time difference between modes. \\


\subsection{Multinomial Mode Choice: Public Transit, Private Car and Non-motorized}

For the multinomial case, we also limit our analysis to the New York City areas. We select home-based work (HBW) trips completed in transit, private car, or non-motorized modes (i.e. walking and bicycling) without limiting the trip length. Trip cost for the non-motorized alternative is set to zero and the travel time is computed using the Google API for the cases where it is not revealed. We discarded observations with missing data, so after the data selection and cleaning process, 3,277 trips are left for our analysis. The description for the variables considered in the multinomial problem is presented in Table \ref{tab:summary_multi}. As for the binary problem, this dataset was also used in \cite{VILLARRAGA2025103132}.

\begin{table}[h!]
	\centering 
	\caption{Summary of the variables considered for the multinomial mode choice problem.}
	\begin{tabular}{p{5cm}p{8cm}S[table-format=3.2]}
		\hline 
		\centering Variable & \centering  Description & \multicolumn{1}{c}{Mean}  \\
		\hline
		\centering Trip cost transit &  Transit cost (USD). &  2.35 \\
          \centering Trip cost car &  Car cost (USD). &  4.64 \\
          \centering Trip cost non-motorized &  Non-motorized cost (USD). & 0.00  \\
		\centering Trip time transit &  Transit travel time (Minutes). & 50.38 \\
  \centering Trip time car &  Car travel time (Minutes). & 18.32 \\
  \centering Trip time non-motorized &  Non-motorized travel time (Minutes). & 93.85 \\
		\centering Vehicle availability &  Indicator variable for car availability in the household. It takes the value of 1 if no cars are available in the household. & 0.36 \\
		\centering High income &  Indicator variable for high income level ($>100$k USD per year). & 0.32 \\
		\centering Manhattan &  Indicator variable for destinations in Manhattan. &  0.46 \\
		\centering Gender &  Indicator variable for the male gender. & 0.47  \\
		\centering \textbf{Car mode share} &  Proportion of trips completed by car & 0.37 \\
      \centering \textbf{Transit mode share} &  Proportion of trips completed by transit & 0.51 \\
      \centering \textbf{Non-motorized mode share} &  Proportion of trips completed by non-motorized means of transportation & 0.12 \\
		\hline
	\end{tabular}
	\label{tab:summary_multi}
\end{table}

\section{US county election 2016}\label{sec:US_elect}

As second case study, we use the 2016 county election dataset as presented in \cite{Tomlinson_Benson_2024}. The data includes socio-demographic characteristics (e.g., death and birth rates, net migration, median income) aggregated by county, election results, and a network constructed using the Social Connectedness Index, which measures the relative frequency of Facebook friendships between each pair of counties. We use this dataset in a binary discrete choice setting with two candidate options, namely: Hillary Clinton and Donald Trump. All other candidates are excluded for simplicity, as no county selected a candidate outside of these two. In total, there are 3112 counties included in the dataset.  \\

\begin{table}[h!]
	\centering 
	\caption{Summary of the variables considered for the binary election choice problem.}
	\begin{tabular}{p{5cm}p{8cm}c}
		\hline 
		\centering Variable & \centering  Description & \multicolumn{1}{c}{Mean}  \\
		\hline
		\centering Death Rate & Percentage of deaths in the population annually. & 10.81\\
        \centering Birth Rate & Percentage of births in the population annually. & 11.62\\
        \centering Net Migration Rate & Percentage change in population due to migration (inflow minus outflow). & -0.04\\
        \centering Bachelor Rate & Percentage of adults aged 25 and older with at least a bachelor's degree. & 21.56\\
        \centering Median Income & Median household income in 2016 US dollars. & 49403\\
        \centering Unemployment Rate & Percentage of the civilian labor force that is unemployed and seeking employment. & 5.20\\
        \centering Rural-urban Continuum Code 2013 & Classification of counties based on population density and proximity to metropolitan areas (1 = most urban, 9 = most rural). Dummy encoded and first category dropped. & -\\
        \centering Economic Typology 2015 & Classification of counties based on predominant economic activity (e.g., farming, manufacturing, mining). Dummy encoded and first category dropped. & -\\
        \centering \textbf{Election Choice ($y$)} & Binary variable indicating whether Donald Trump (1) or Hillary Clinton (0) received the majority vote in the county. & 0.8425 \\
		\hline
	\end{tabular}
	\label{tab:summary_election}
\end{table}

\begin{figure}[h!]
    \centering
    \includegraphics[width=1\linewidth]{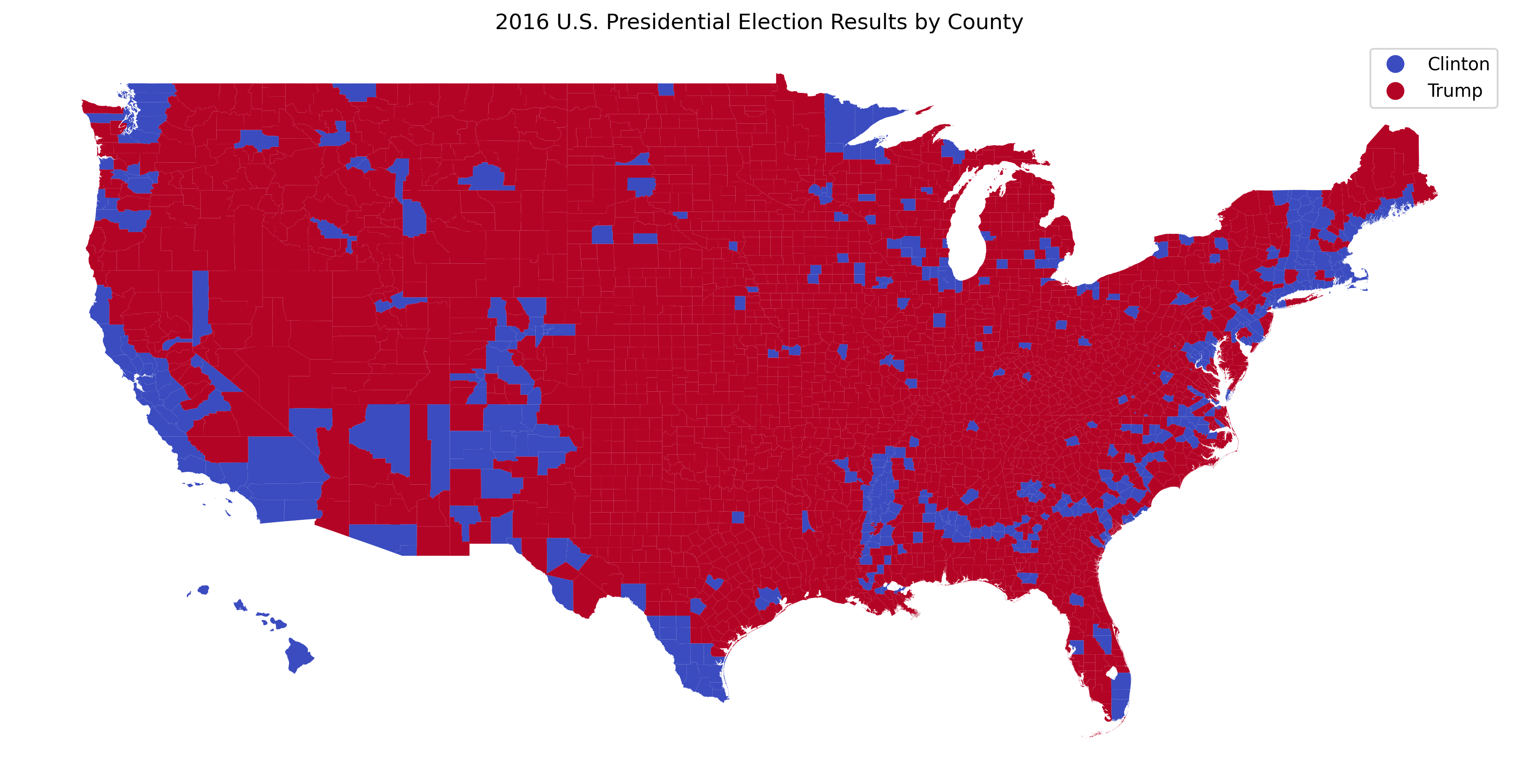}
    \caption{2016 US Election Results by County.}
    \label{fig:map_election}
\end{figure}

The map of the election results by county is presented in Figure \ref{fig:map_election}. As shown, nearly 85\% of counties had a majority of the population voting for Trump. This figure makes evident that neighboring counties often voted similarly. This is particularly noticeable in the blue clusters on the map, where several adjacent counties supported the Democrat candidate. \\

\section{Results and Discussion \label{results}}

In this section, we evaluate and compare the predictive performance and behavioral insights obtained from our proposed architecture against those derived from an out-of-the-shelf GNN and a traditional logit model. The comparison includes results from both the standard models and their Stochastic Gradient Langevin Dynamics (SGLD) counterparts.\\

For the deep learning models, we perform a random grid search for the layer width, the number of fully connected neural network layers, the number of GCN layers, the weight decay rate, and the learning rate. We use ReLU activations for all non-linearities except for the output layer, for which we use the Sigmoid activation (binary) or Softmax (multinomial). For SGLD, we run the algorithm for 100,000 epochs with thinning every 1,000 epochs to reduce MCMC iterate autocorrelation.\\

\subsection{Case study: Binary Mode Choice in NYC}

In this subsection, we present the marginal utilities and value of travel time savings VOTTs estimated for the binary NYC mode choice dataset using a standard logit model, our Skip-GNN architecture, and a general-purpose GNN. We demonstrate that our Skip-GNN model provides insights that align with behavioral intuition, while SGLD (Stochastic Weight Averaging) improves the behavioral alignment of the off-the-shelf GNN. \\ 

\subsubsection{Marginal Utilities}

\begin{figure}[ht]
    \centering
    \begin{minipage}{.45\textwidth}
        \centering
        \includegraphics[width=\linewidth]{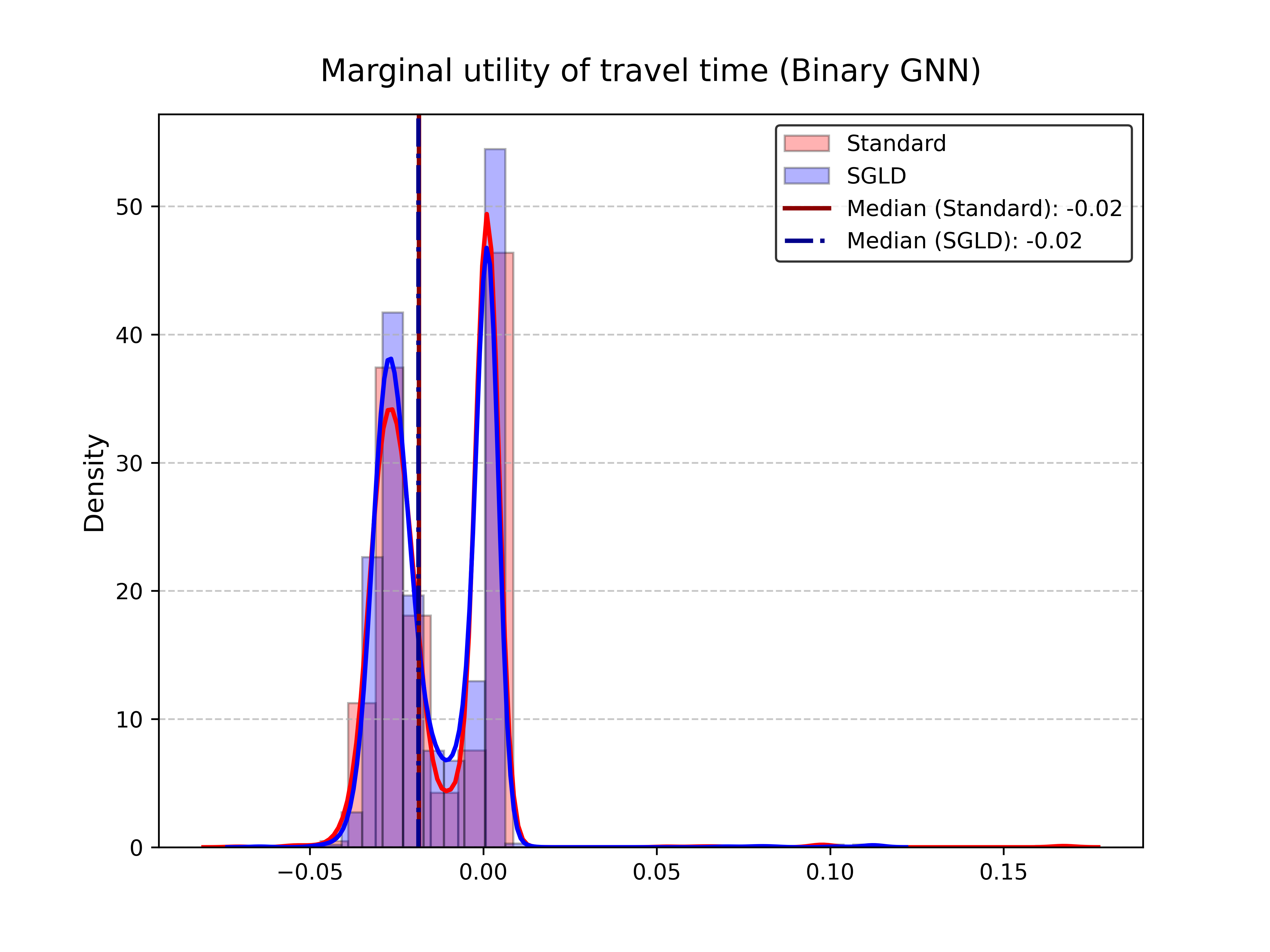}
        \caption{Marginal utility of travel time from general purpose GNN.}
        \label{fig:m_t_GNN}
    \end{minipage}%
    \hfill
    \begin{minipage}{.45\textwidth}
        \centering
        \includegraphics[width=\linewidth]{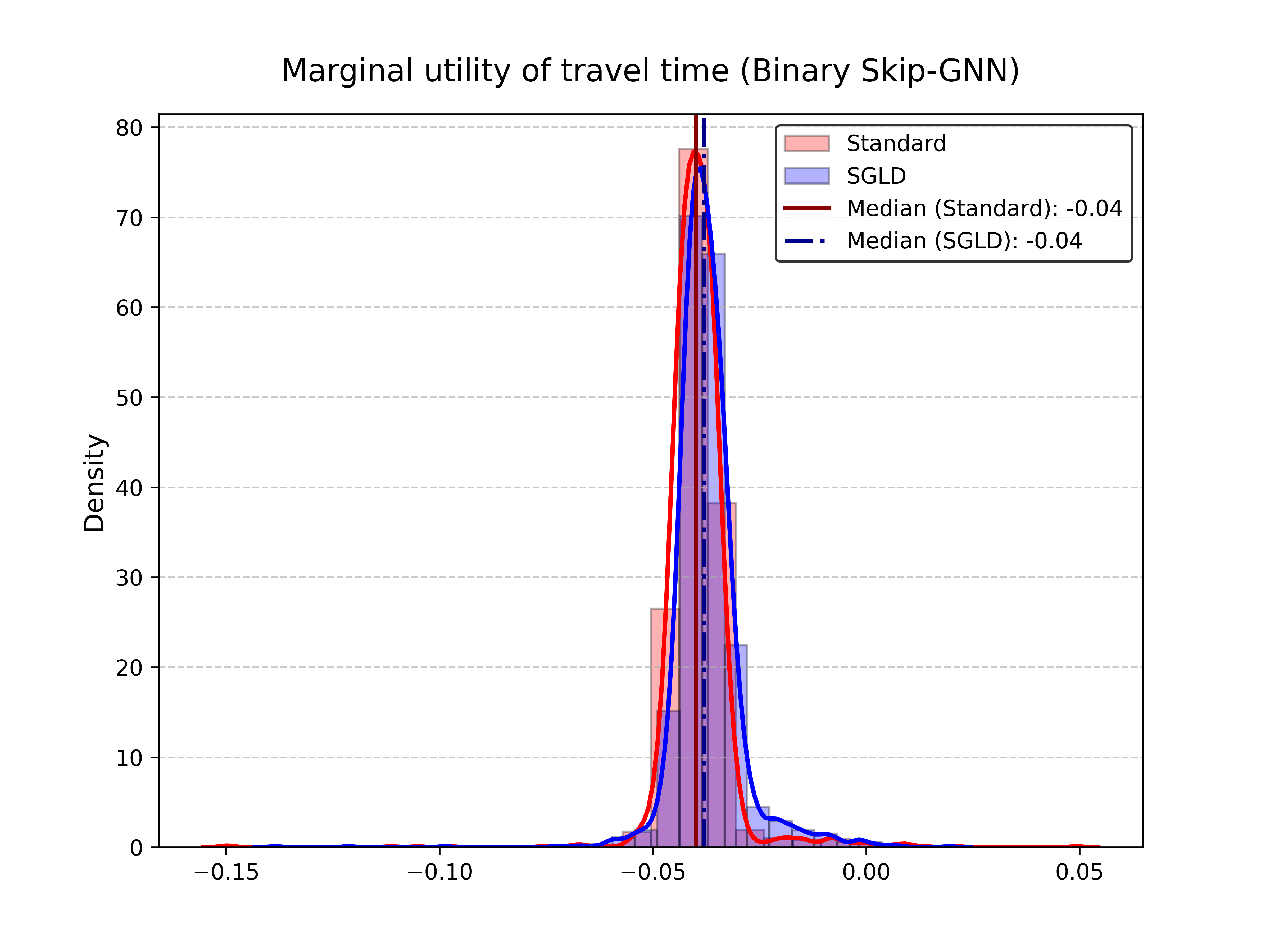}
        \caption{Marginal utility of travel time from the Skip-GNN.}
        \label{fig:m_t_Skip_GNN}
    \end{minipage}
\end{figure}

\begin{figure}[ht]
    \centering
    \begin{minipage}{.45\textwidth}
        \centering
        \includegraphics[width=\linewidth]{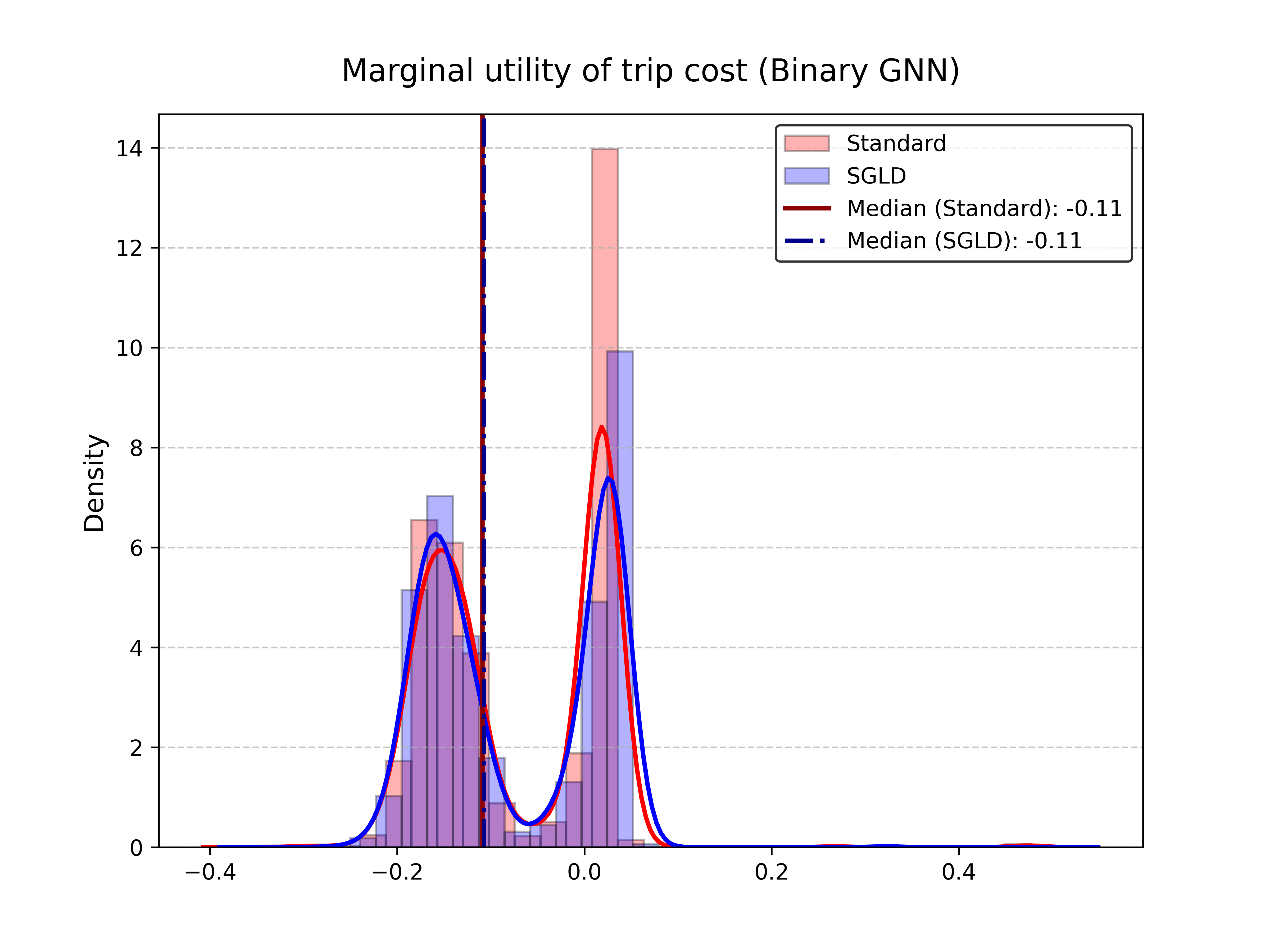}
        \caption{Marginal utility of trip cost for the general purpose GNN.}
        \label{fig:m_c_GNN}
    \end{minipage}%
    \hfill
    \begin{minipage}{.45\textwidth}
        \centering
        \includegraphics[width=\linewidth]{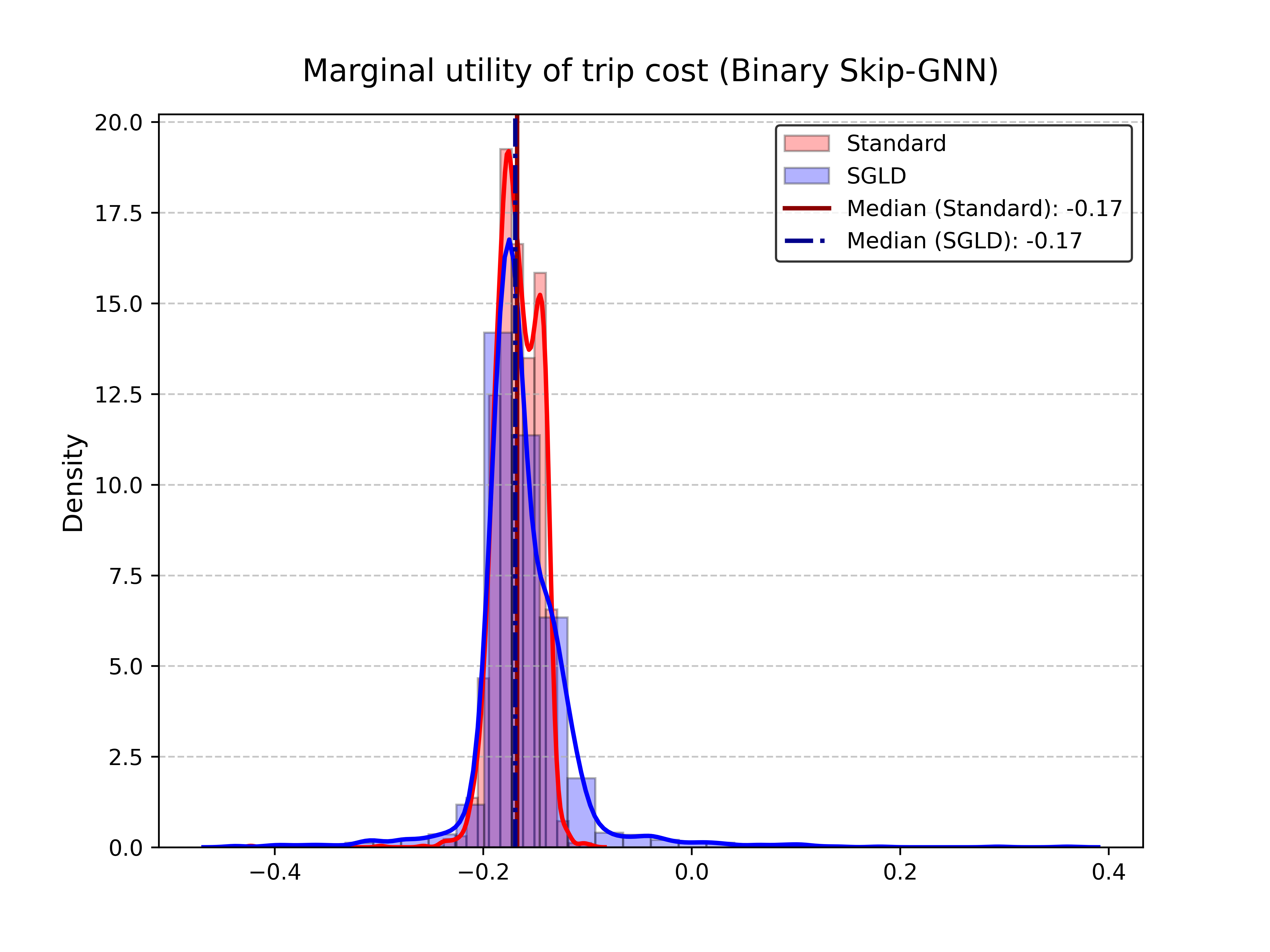}
        \caption{Marginal utility of trip cost for the Skip-GNN.}
        \label{fig:m_c_Skip_GNN}
    \end{minipage}
\end{figure}

For the logit model, we found the marginal utility of travel time, denoted as $\beta_t$, to be -0.022, and the marginal utility of trip cost, denoted as $\beta_c$, to be -0.101. These negative values align with behavioral expectations, as anticipated. In the case of the general-purpose graph neural network, the individual marginal utility of travel time and trip cost are illustrated in Figures \ref{fig:m_t_GNN} and \ref{fig:m_c_GNN}, respectively. For our proposed Skip-GNN model, the corresponding marginal utilities are detailed in Figures \ref{fig:m_t_Skip_GNN} and \ref{fig:m_c_Skip_GNN}. \\

Regarding the marginal utilities of travel time, it is observed that the general-purpose graph neural network (GNN) yields positive marginal utilities for a significant proportion of individuals (as shown in Figure \ref{fig:m_t_GNN}), which contradicts micro-economic expectation. Nonetheless, the median marginal utility of travel time across the sample aligns with the expected sign, being equal to -0.02. The discrepancies between the Stochastic Gradient Langevin Dynamics (SGLD) approach and the regular estimation procedure in this context are not pronounced. Conversely, for our proposed skip-GNN model, the marginal utilities of travel time are consistently negative across the entire sample (as shown in Figure \ref{fig:m_t_Skip_GNN}), thereby enhancing regularity of the outcome.\\

For the marginal utility of trip cost, we observe similar results as those just discussed for the marginal utility of travel time. The general-purpose architecture yields positive values for a substantial proportion of individuals (as indicated in Figure \ref{fig:m_c_GNN}), in contrast to the consistently negative values generated by our proposed skip-GNN model across the dataset (Figure \ref{fig:m_c_Skip_GNN}). Nevertheless, similar to the findings for travel time, the median marginal utility of trip cost derived from the general-purpose GNN aligns with the expected negative marginal effect, being -0.11.

\subsubsection{Value of Travel Time Savings}

\begin{figure}[ht]
    \centering
    \begin{minipage}[t]{.45\textwidth}
        \centering
        \includegraphics[width=\linewidth]{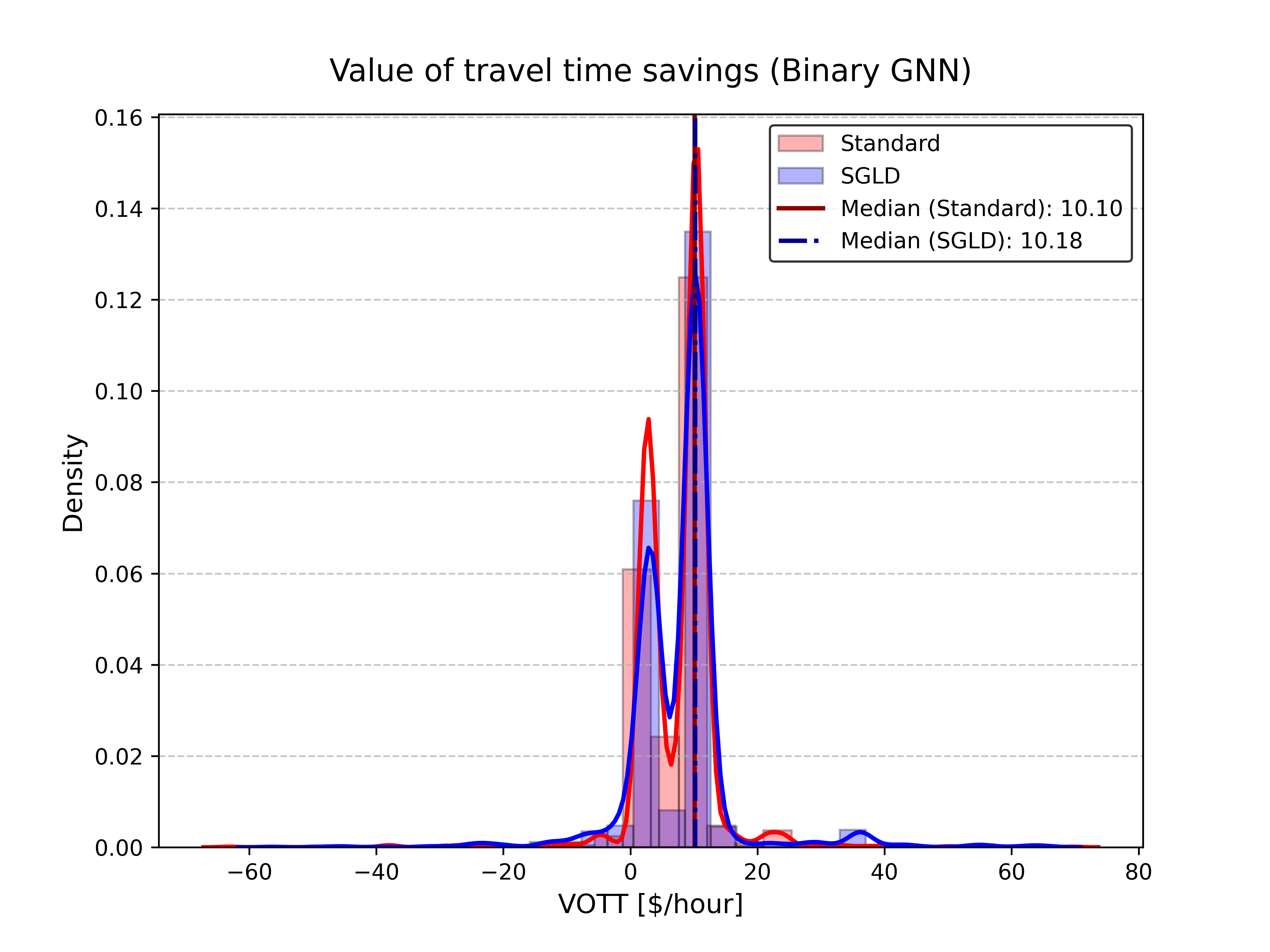}
        \caption{value of travel time savings from the general purpose GNN.}
        \label{fig:VOT_GNN}
    \end{minipage}%
    \hfill
    \begin{minipage}[t]{.45\textwidth}
        \centering
        \includegraphics[width=\linewidth]{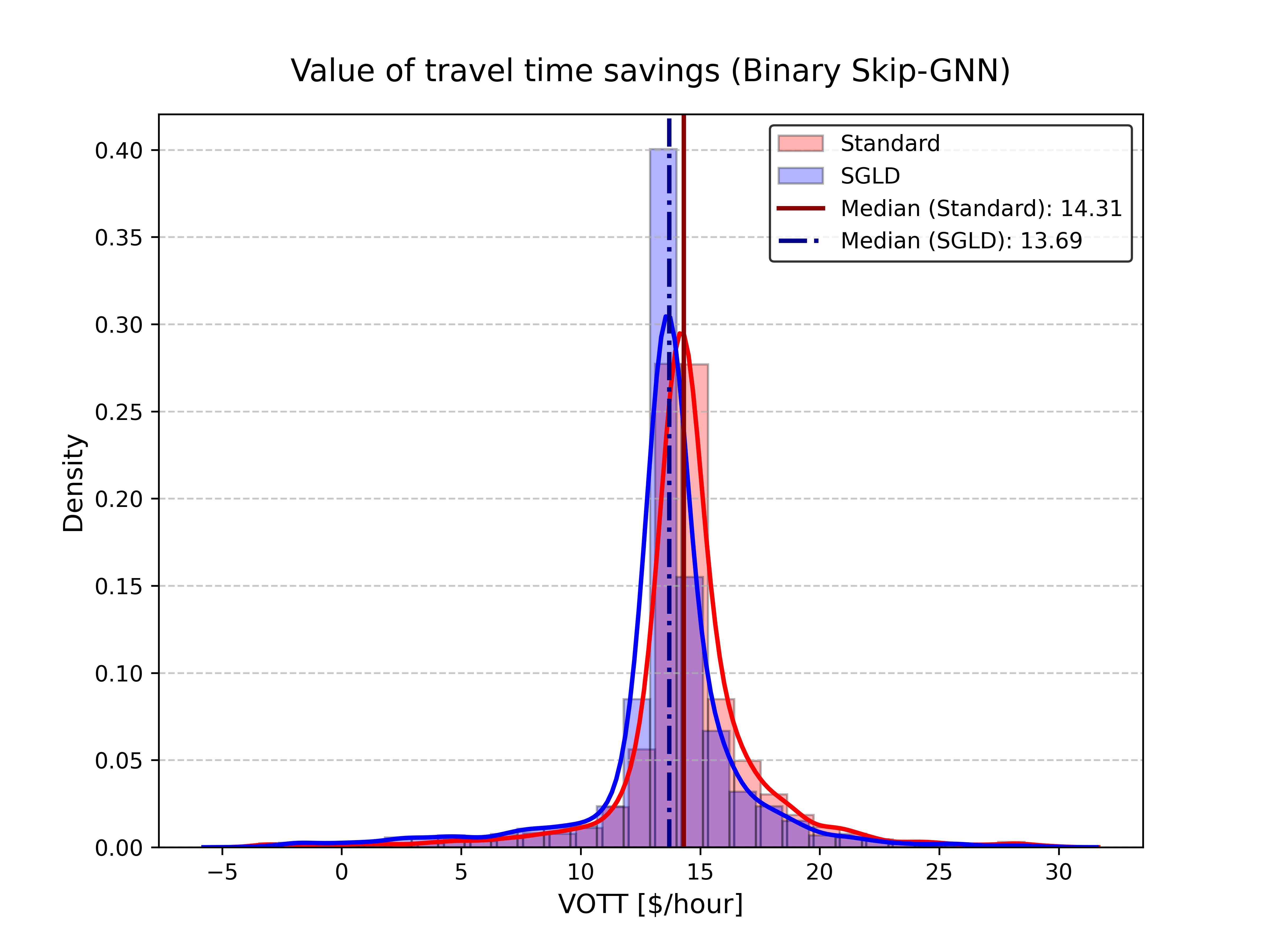}
        \caption{value of travel time savings from the Skip-GNN.}
        \label{fig:VOT_Skip_GNN}
    \end{minipage}
\end{figure}

Regarding the value of travel time savings (VOTT), which is the marginal rate of substitution between travel time and travel cost and thus expected to be positive (when representing the willingness to pay to reduce travel time by a marginal unit), the point estimate obtained using the logit model is 15.75 USD per hour reduction in travel time. The VOTT values for the entire sample as derived from the general-purpose graph neural network (GNN) are illustrated in Figure \ref{fig:VOT_GNN}, while those obtained from our skip-GNN model are presented in Figure \ref{fig:VOT_Skip_GNN}. These figures provide insights into the distribution or average values of VOTT, as determined by the respective models.\\

For the standard GNN, the median VOTT across the sample is calculated to be 10.10 USD per hour. In contrast, for the SGLD variant of this model, the VOTT is slightly higher, at 10.18 USD per hour. As depicted in Figure \ref{fig:VOT_GNN}, the distribution of VOTT estimates for the standard GNN model includes a significant proportion of individuals with negative VOTT values, with values ranging from -70 USD to 80 USD. However, the SGLD version shows a more concentrated distribution around the mean with a reduced frequency of negative VOTT values.\\

The median VOTT estimated using our skip-GNN model is 14.31 USD per hour for the conventional estimates. In the case of the SGLD variant, the median VOTT is slightly lower, at 13.69 USD per hour. As illustrated in Figure \ref{fig:VOT_GNN}, the VOTT estimates for the entire sample using both the regular and SGLD versions of the skip-GNN model exhibit very sensible values.\footnote{In \cite{ren2023block}, the authors found an average value of travel time savings (VOTT) for the low-income population to be 21.67 USD/hour, for the not low-income population to be 28.05 USD/hour, for the student population to be 10.96 USD/hour, and the average VOTT for the senior population to be 10.93 USD/hour. Their estimates are based on a group-level agent-based mixed (GLAM) logit applied to synthetic nation-wide level data. To the best of our knowledge, there are no other recent estimates derived from discrete choice models for the value of travel time savings in NYC. }

\subsection{Case Study: Multinomial Mode Choice in NYC}

In this subsection, we present the marginal utilities and values of time estimated for the multinomial mode choice dataset using a standard conditional logit model, our Skip-GNN architecture (both with and without representing the Independence of Irrelevant Alternatives (IIA)), and a general-purpose GNN. We demonstrate that our proposed model offers insights that align more closely with behavioral intuition than those provided by the general-purpose GNN. Additionally, we observe that SGLD have a notable effect on the marginal utilities and VOTT for our Skip-GNN model.\\

\subsubsection{Marginal Utilities}

\begin{figure}[ht]
    \centering
    \begin{minipage}{.45\textwidth}
        \centering
        \includegraphics[width=\linewidth]{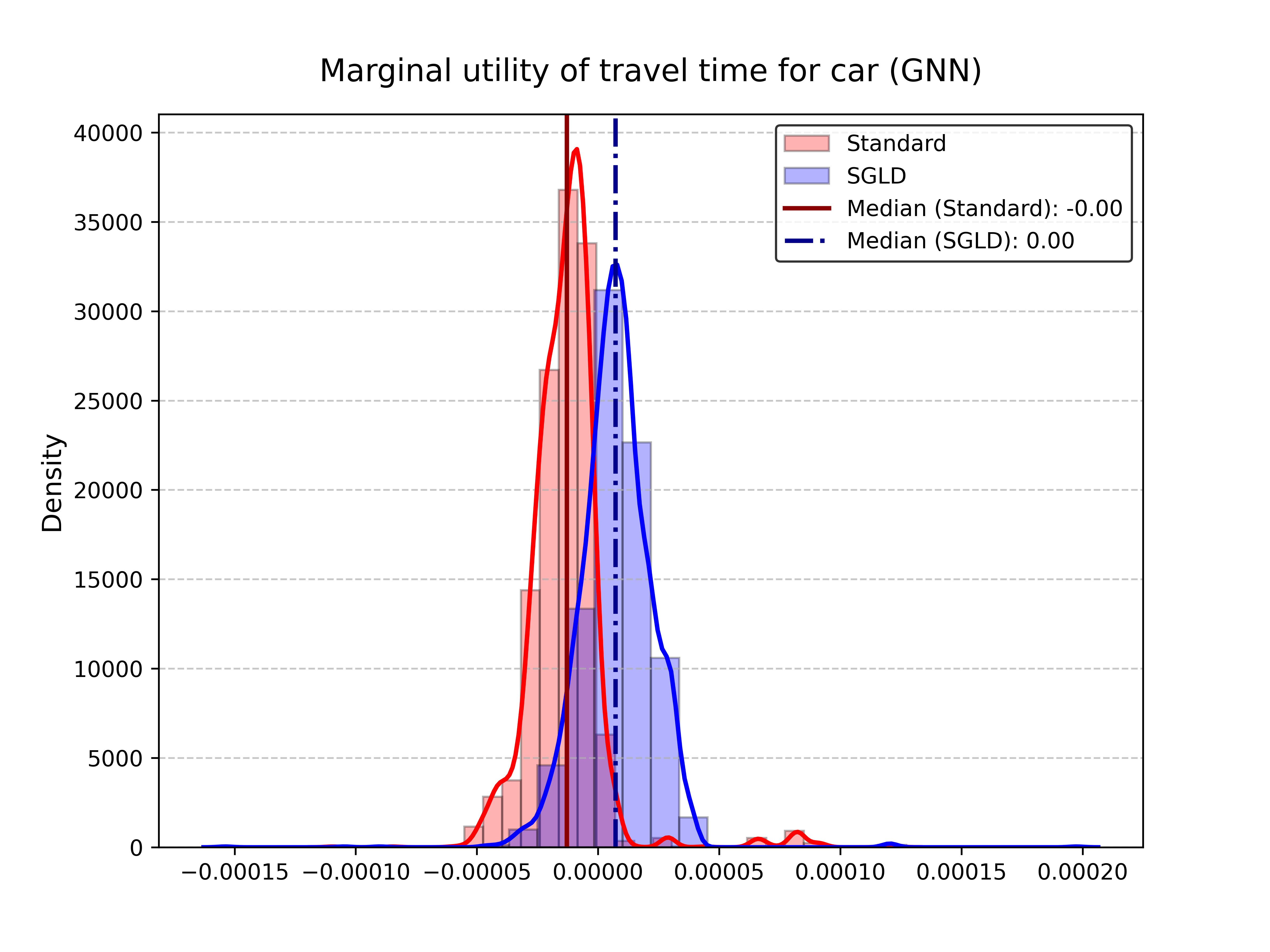}
        \caption{Marginal utility of travel time for car from general purpose GNN.}
        \label{fig:m_t_car_GNN Multinomial}
    \end{minipage}%
    \hfill
    \begin{minipage}{.45\textwidth}
        \centering
        \includegraphics[width=\linewidth]{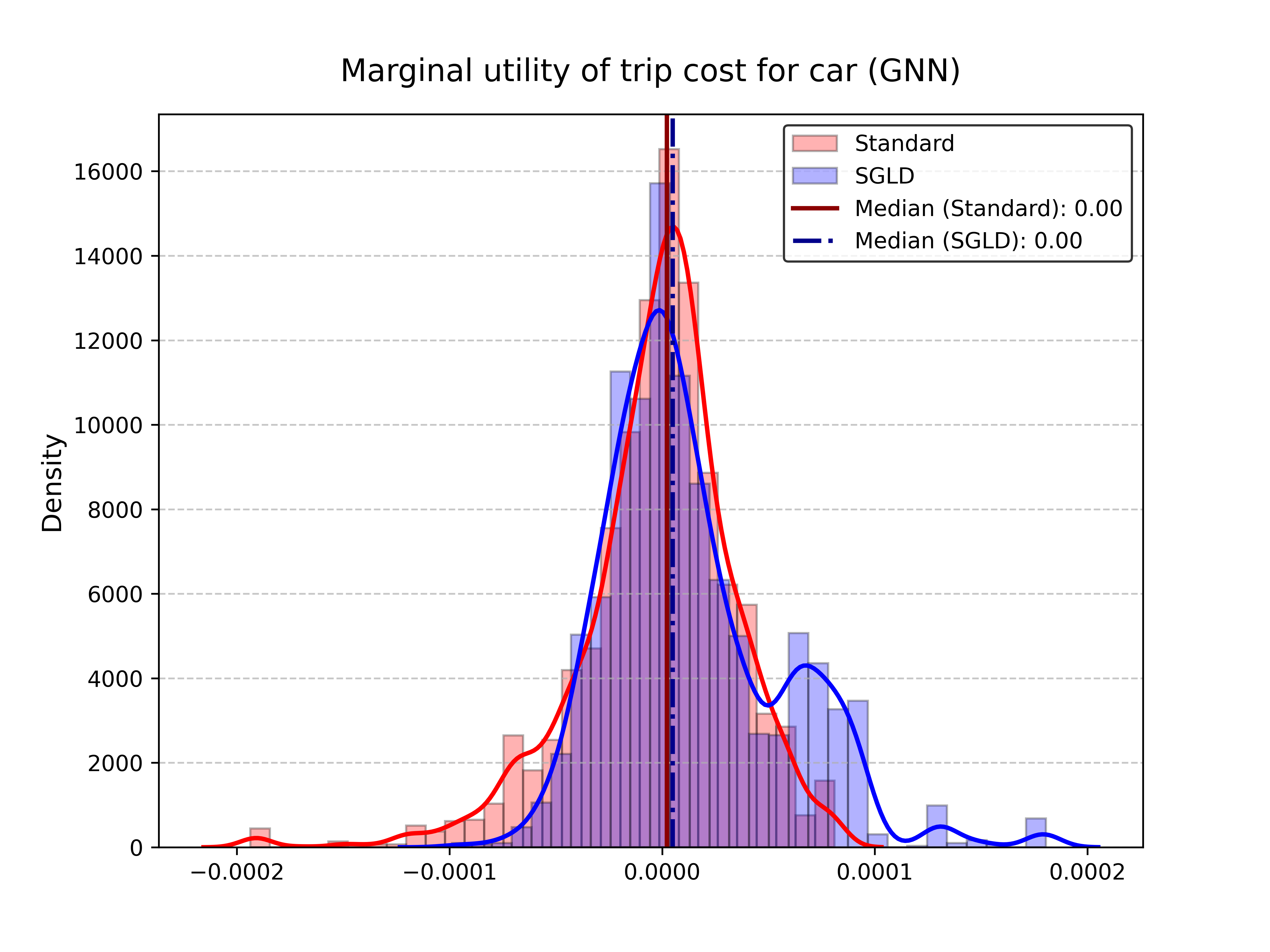}
        \caption{Marginal utility of trip cost for car from general purpose GNN.}
        \label{fig:m_c_car_GNN Multinomial}
    \end{minipage}
\end{figure}

\begin{figure}[ht]
    \centering
    \begin{minipage}{.45\textwidth}
        \centering
        \includegraphics[width=\linewidth]{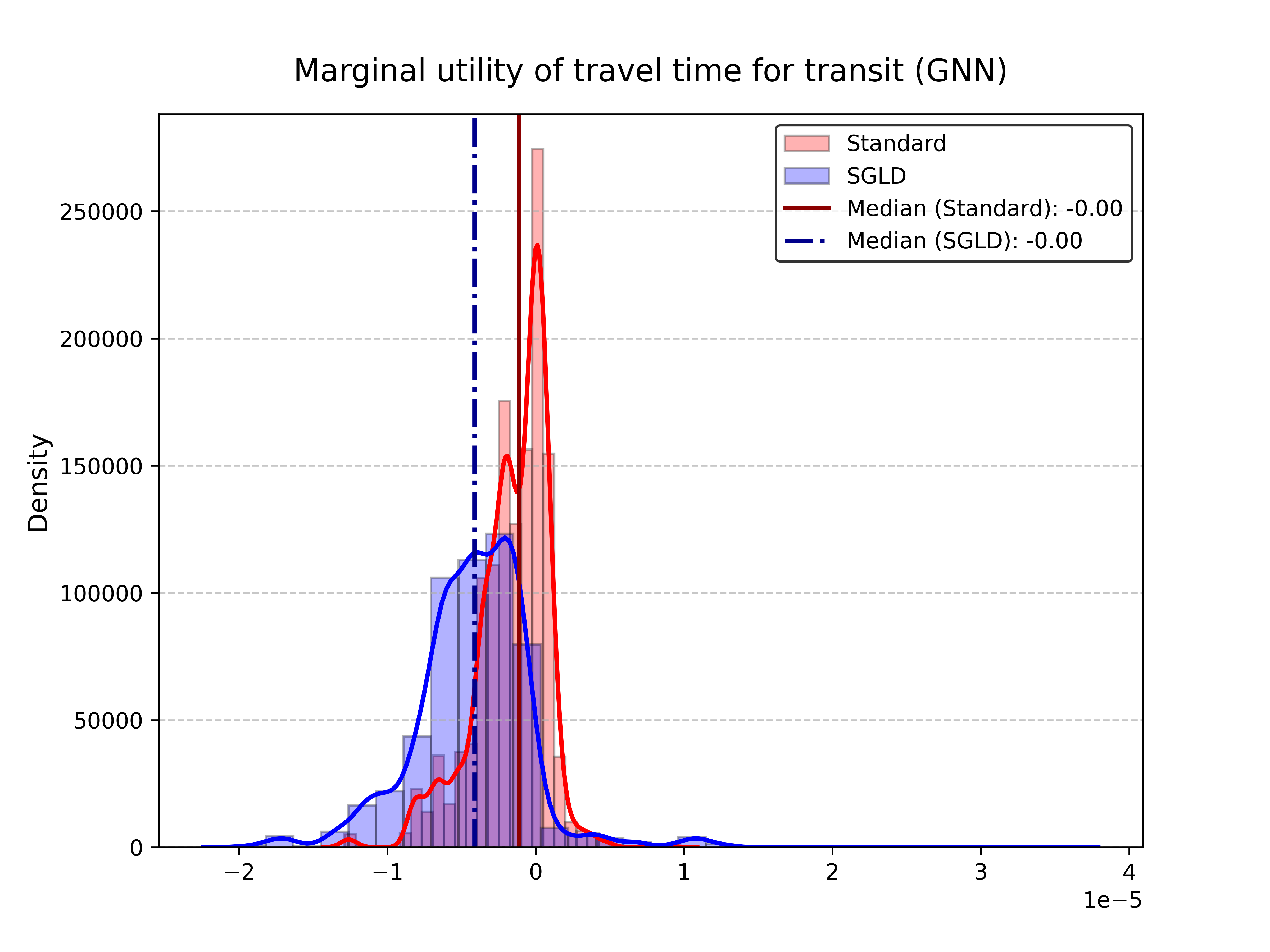}
        \caption{Marginal utility of travel time for transit from general purpose GNN.}
        \label{fig:m_t_transit_GNN Multinomial}
    \end{minipage}%
    \hfill
    \begin{minipage}{.45\textwidth}
        \centering
        \includegraphics[width=\linewidth]{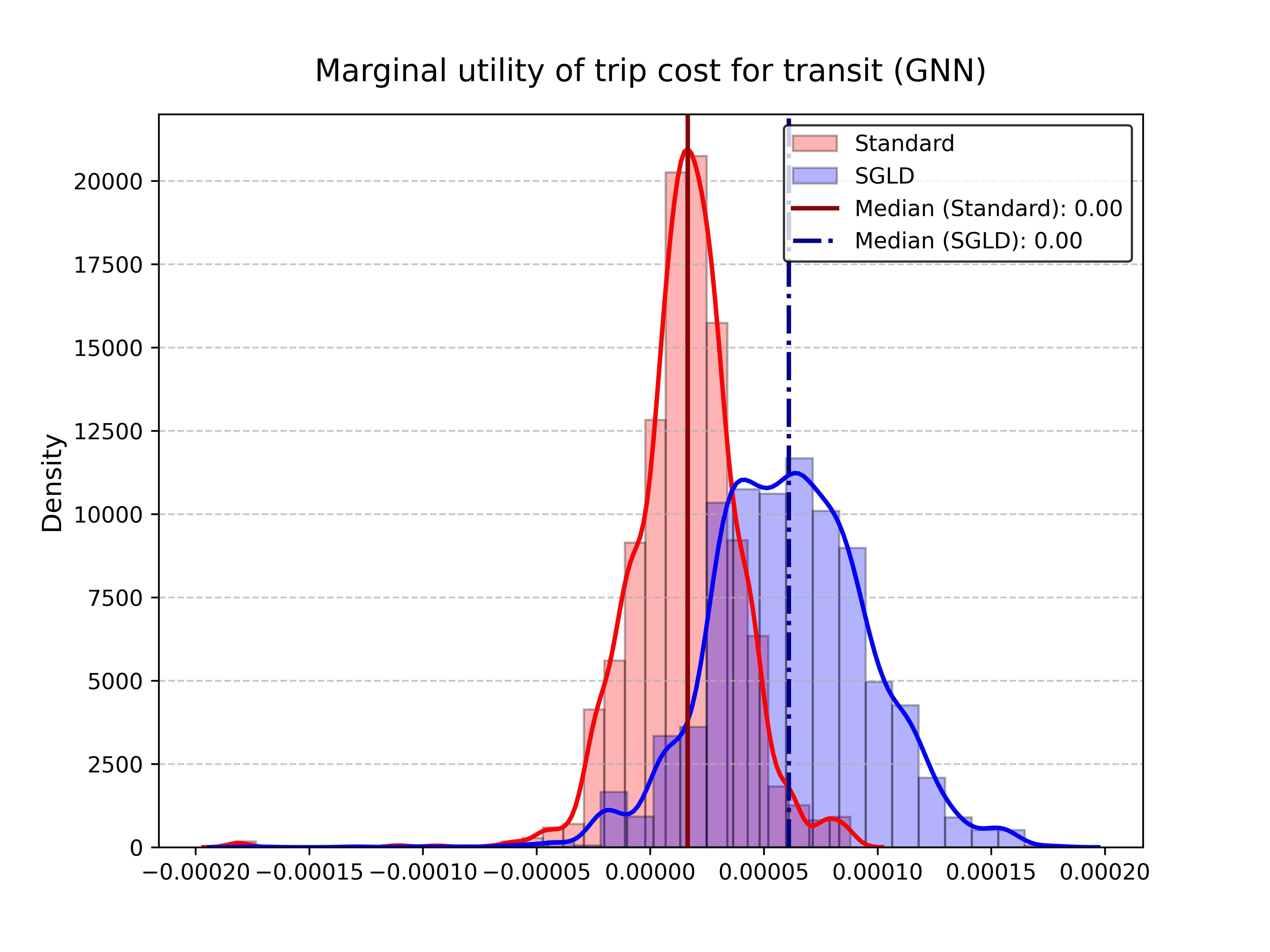}
        \caption{Marginal utility of trip cost for transit from general purpose GNN.}
        \label{fig:m_c_transit_GNN Multinomial}
    \end{minipage}
\end{figure}

We discuss first the histograms for the marginal utilities of travel time and trip cost for transit and private car for the whole sample of individuals. The marginal utilities for car travel time and trip cost, obtained from the general-purpose GNN model, are illustrated in Figures \ref{fig:m_t_car_GNN Multinomial} and \ref{fig:m_c_car_GNN Multinomial}, respectively. The marginal utilities for transit travel time and trip cost are illustrated in Figures \ref{fig:m_t_transit_GNN Multinomial} and \ref{fig:m_c_transit_GNN Multinomial}. As shown in these figures, the marginal utilities exhibit both positive and negative values across the sample at seemingly equal proportions. This observation contradicts micro-economic sign expectation, which would suggest that travel mode utilities should be lower for alternatives with higher costs or longer trip times. \\

\begin{figure}[ht]
    \centering
    \begin{minipage}{.45\textwidth}
        \centering
        \includegraphics[width=\linewidth]{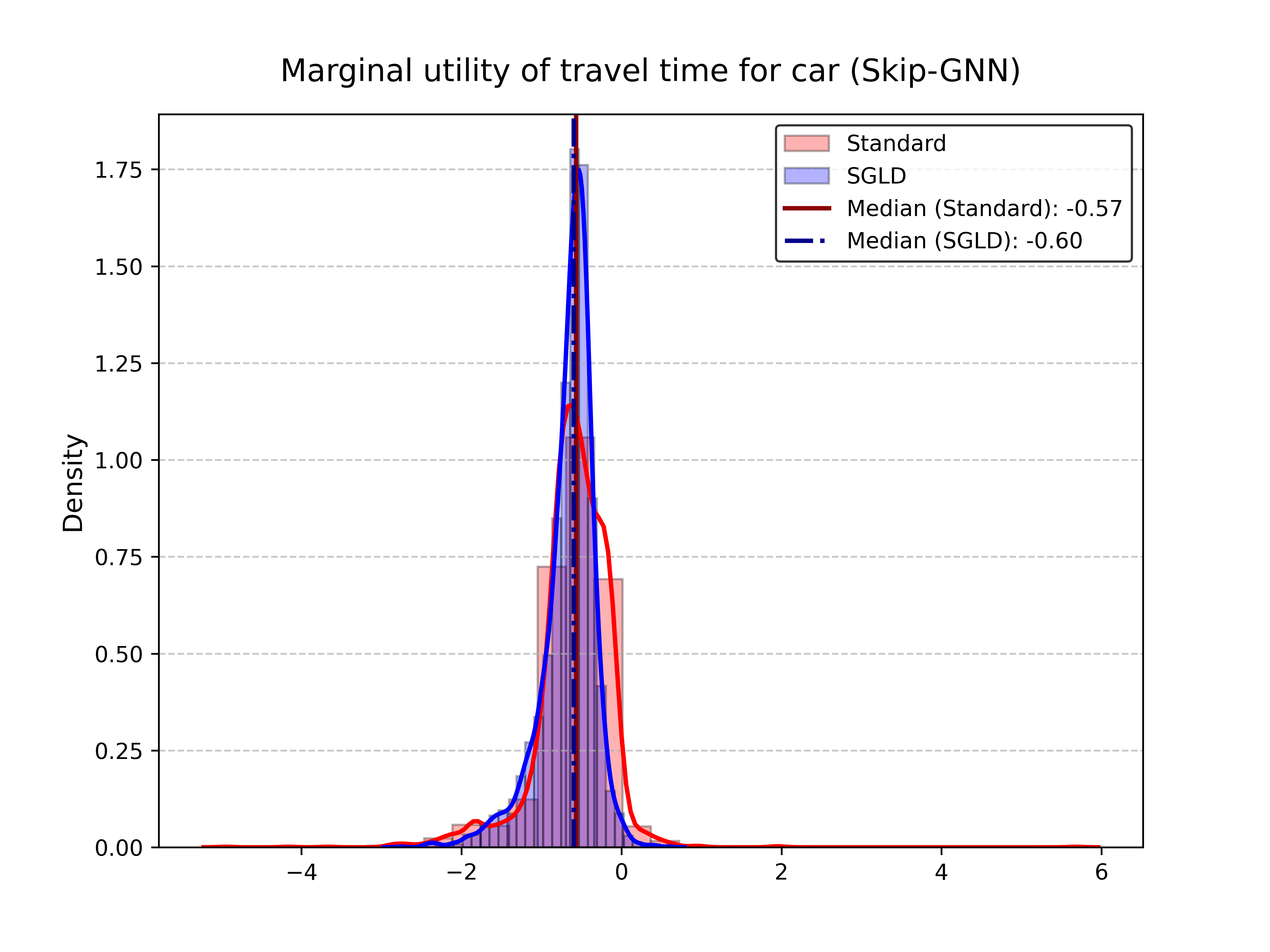}
        \caption{Marginal utility of travel time for car from Skip-GNN model.}
        \label{fig:m_t_car_Skip-GNN Multinomial}
    \end{minipage}%
    \hfill
    \begin{minipage}{.45\textwidth}
        \centering
        \includegraphics[width=\linewidth]{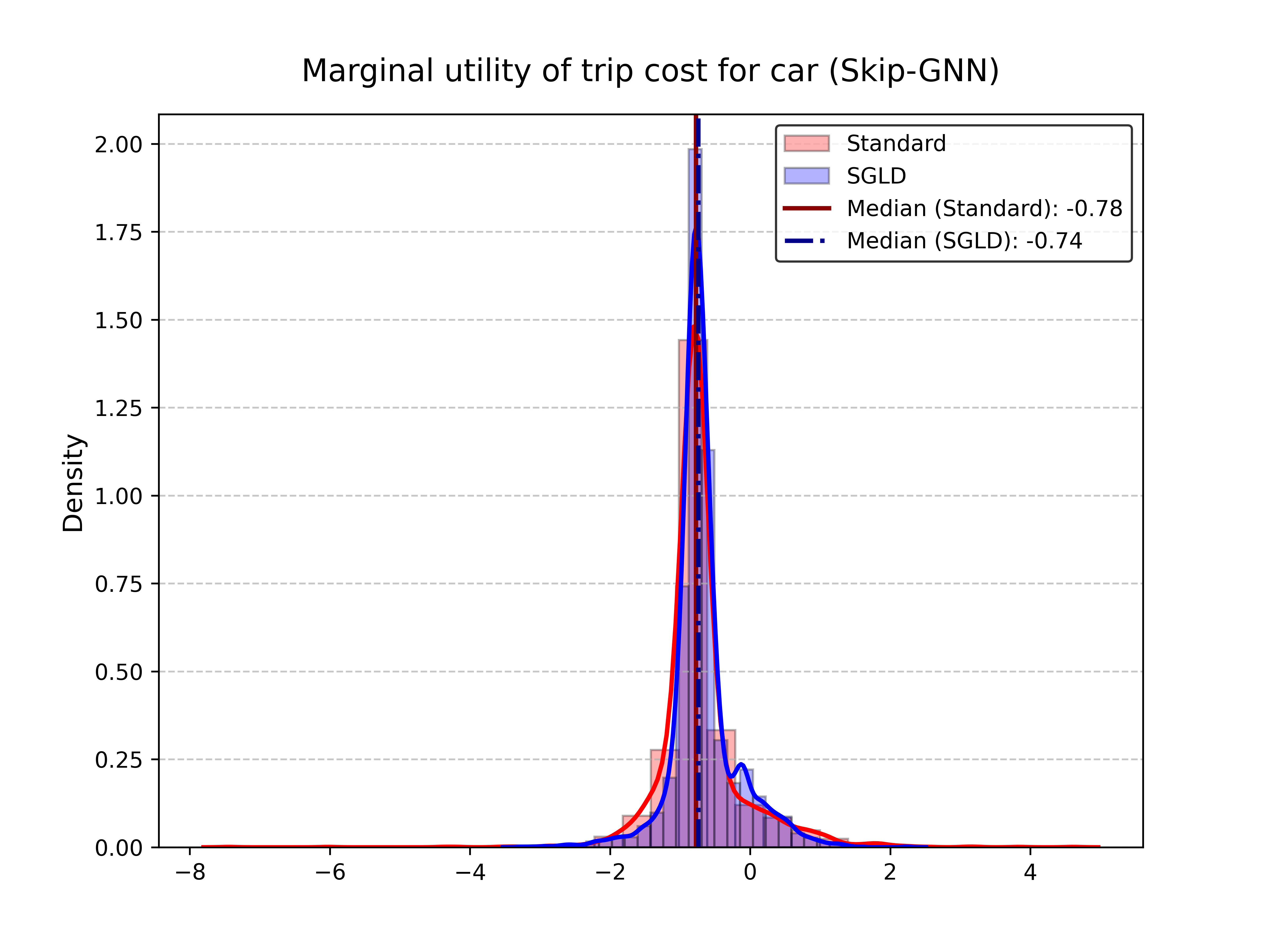}
        \caption{Marginal utility of trip cost for car from Skip-GNN model.}
        \label{fig:m_c_car_Skip-GNN Multinomial}
    \end{minipage}
\end{figure}

\begin{figure}[ht]
    \centering
    \begin{minipage}{.45\textwidth}
        \centering
        \includegraphics[width=\linewidth]{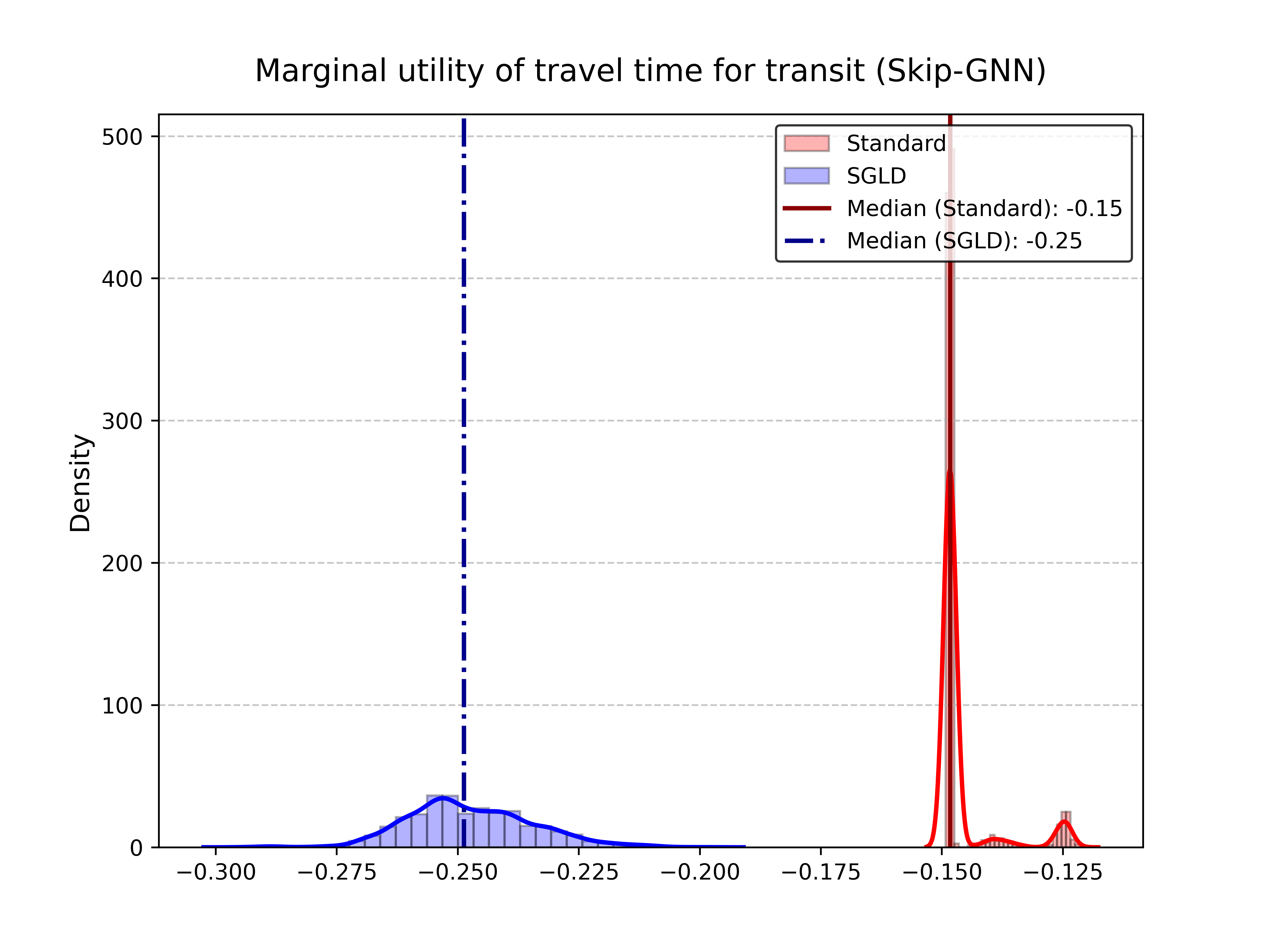}
        \caption{Marginal utility of travel time for transit from Skip-GNN model.}
        \label{fig:m_t_transit_Skip-GNN Multinomial}
    \end{minipage}%
    \hfill
    \begin{minipage}{.45\textwidth}
        \centering
        \includegraphics[width=\linewidth]{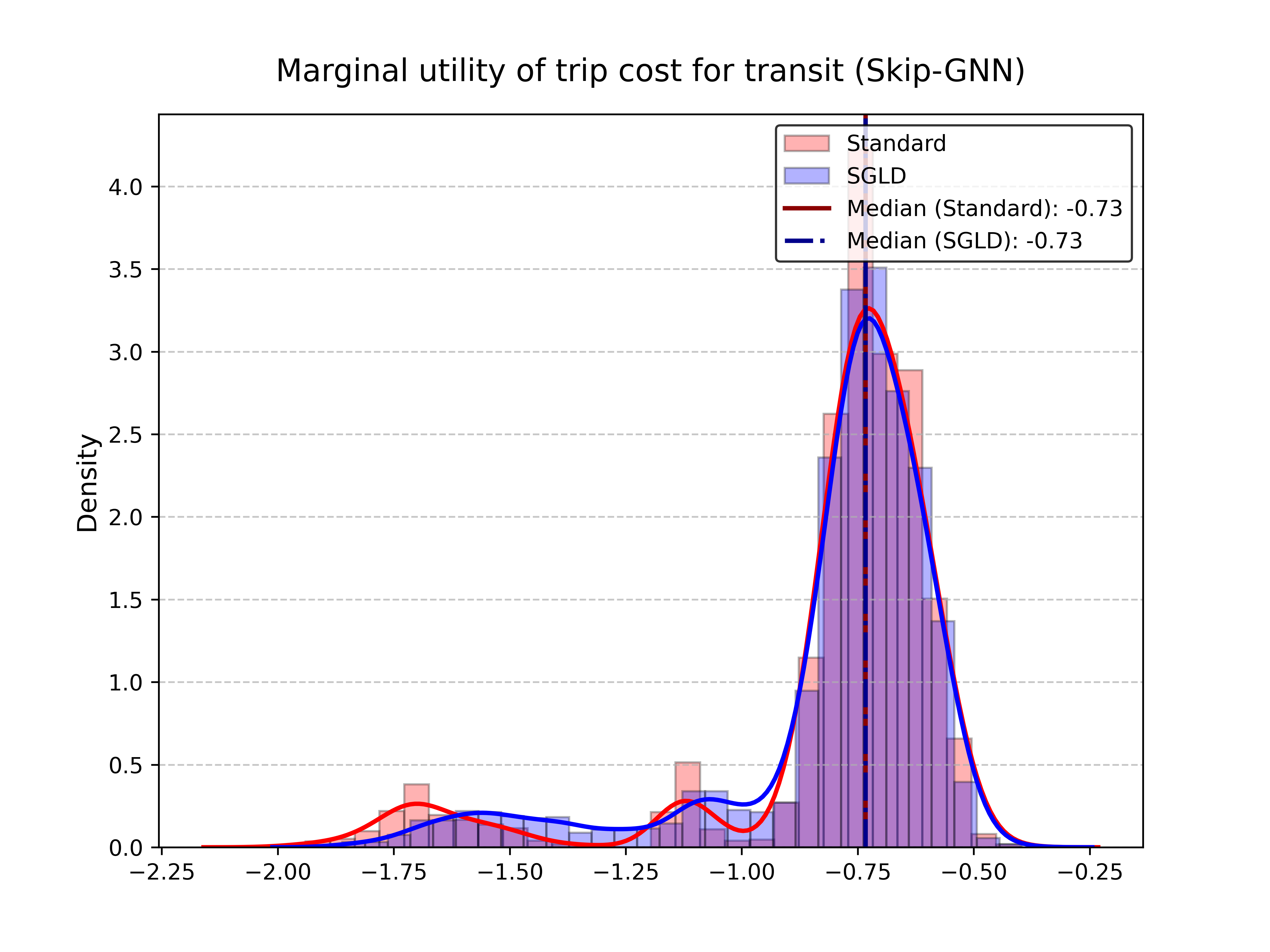}
        \caption{Marginal utility of trip cost for transit from Skip-GNN model.}
        \label{fig:m_c_transit_Skip-GNN Multinomial}
    \end{minipage}
\end{figure}

The marginal utilities for car travel time and trip cost, derived from our Skip-GNN model, are depicted in Figures \ref{fig:m_t_car_Skip-GNN Multinomial} and \ref{fig:m_c_car_Skip-GNN Multinomial}, respectively. Similarly, the marginal utilities for the transit mode are shown in Figures \ref{fig:m_t_transit_Skip-GNN Multinomial} and \ref{fig:m_c_transit_Skip-GNN Multinomial}. Contrasting these results to those from the general-purpose architecture, our findings reveal marginal utilities that are more consistent with behavioral expectations, as evidenced by histograms that show very few observations with positive marginal utilities. \\


\begin{figure}[ht]
    \centering
    \begin{minipage}{.45\textwidth}
        \centering
        \includegraphics[width=\linewidth]{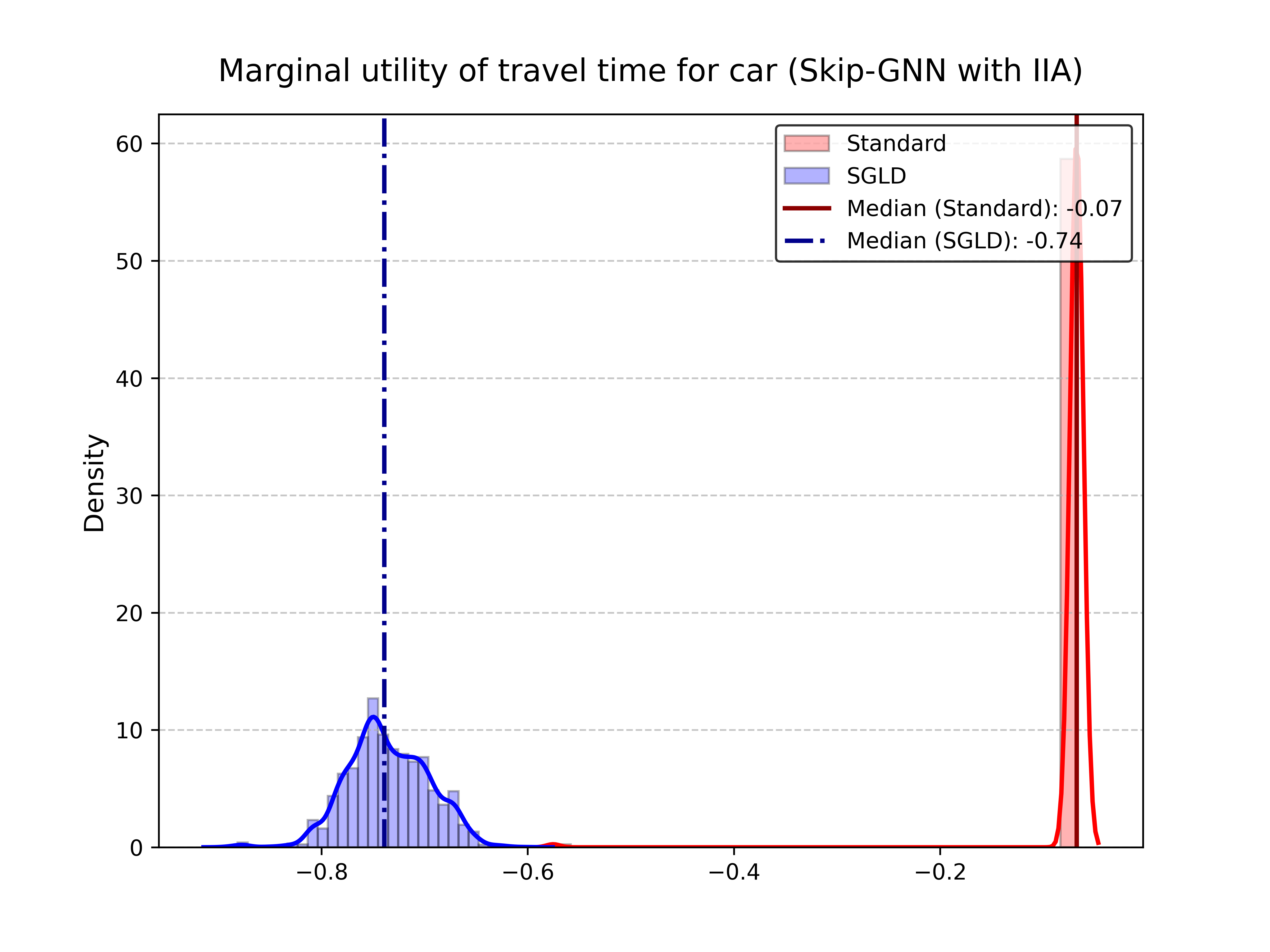}
        \caption{Marginal utility of travel time for car from Skip-GNN-IIA model.}
        \label{fig:m_t_car_Skip-GNN-IIA Multinomial}
    \end{minipage}%
    \hfill
    \begin{minipage}{.45\textwidth}
        \centering
        \includegraphics[width=\linewidth]{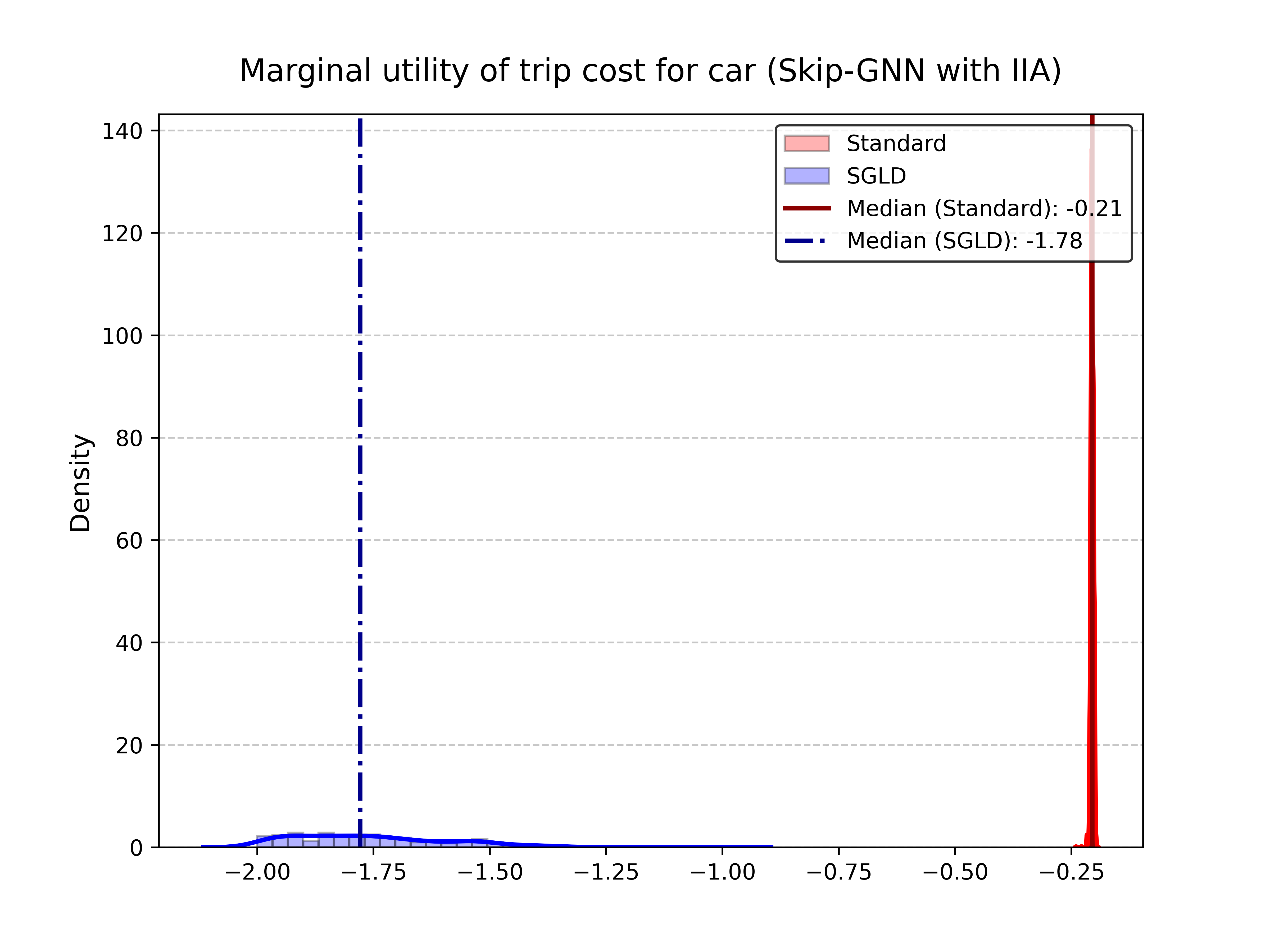}
        \caption{Marginal utility of trip cost for car from Skip-GNN-IIA model.}
        \label{fig:m_c_car_Skip-GNN-IIA Multinomial}
    \end{minipage}
\end{figure}

\begin{figure}[ht]
    \centering
    \begin{minipage}{.45\textwidth}
        \centering
        \includegraphics[width=\linewidth]{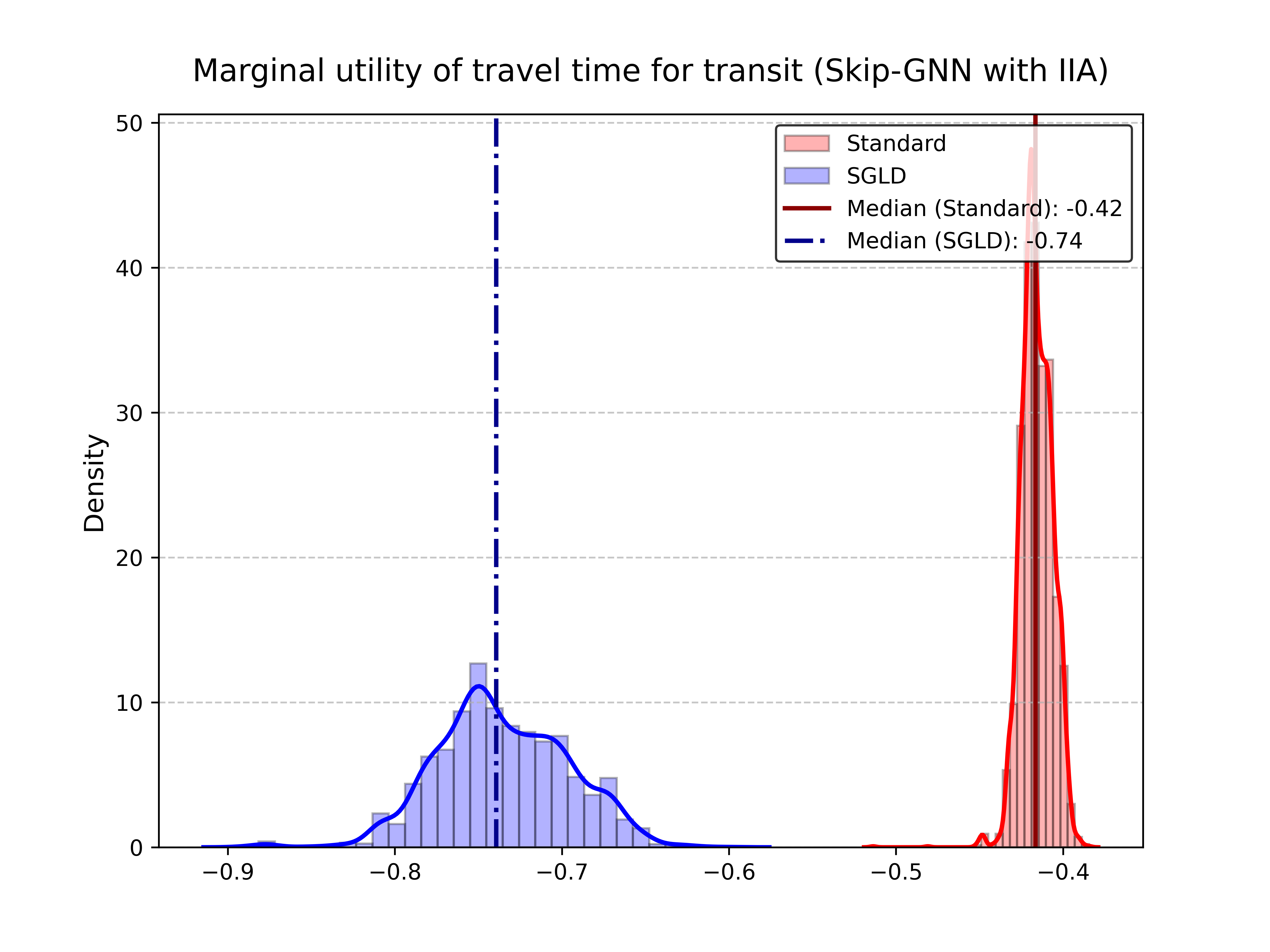}
        \caption{Marginal utility of travel time for transit from Skip-GNN-IIA model.}
        \label{fig:m_t_transit_Skip-GNN-IIA Multinomial}
    \end{minipage}%
    \hfill
    \begin{minipage}{.45\textwidth}
        \centering
        \includegraphics[width=\linewidth]{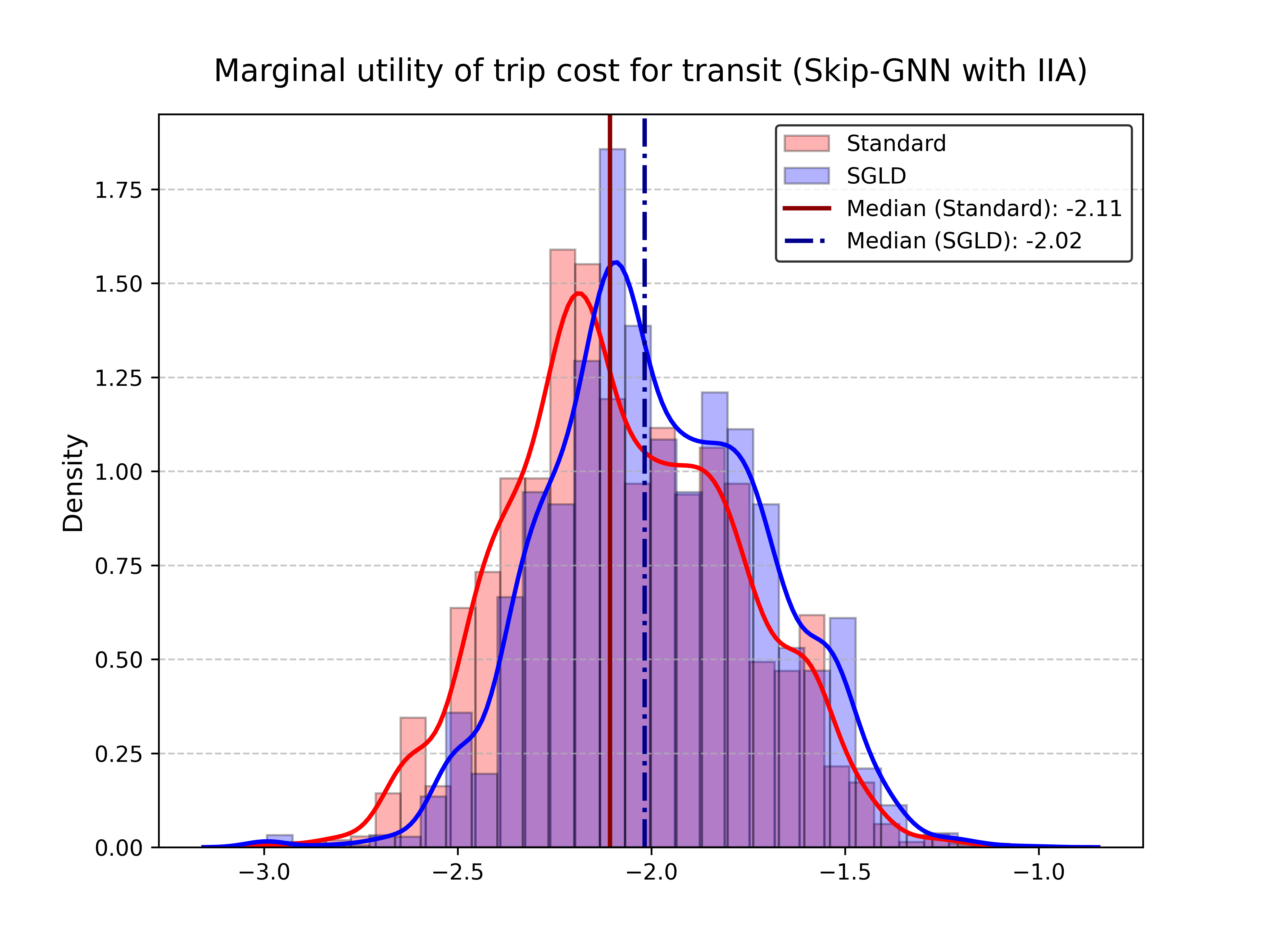}
        \caption{Marginal utility of trip cost for transit from Skip-GNN-IIA model.}
        \label{fig:m_c_transit_Skip-GNN-IIA Multinomial}
    \end{minipage}
\end{figure}

The marginal utilities for car travel time and trip cost, obtained from our Skip-GNN-IIA model, are illustrated in Figures \ref{fig:m_t_car_Skip-GNN-IIA Multinomial} and \ref{fig:m_c_car_Skip-GNN-IIA Multinomial}, respectively. Similarly, the marginal utilities for the transit mode are displayed in Figures \ref{fig:m_t_transit_Skip-GNN-IIA Multinomial} and \ref{fig:m_c_transit_Skip-GNN-IIA Multinomial}. The results from this variant, which applies the Independence of Irrelevant Alternatives (IIA) restriction, are on par with those from the Skip-GNN model without IIA. This consistency indicates that our model can operate effectively with or without the IIA constraint, without compromising its ability to align with expected behavioral regularity\footnote{For model selection, when behavioral regularity is met, out-of-sample prediction metrics should be used.}. It is also important to note that the marginal utilities obtained with SGLD tend to be further away from zero than those found using the regular estimation procedure, which translates to a lower—or even null—proportion of individuals with marginal utilities that defy behavioral intuition.\\


\subsubsection{Value of Travel Time Savings}

The estimated value of travel time savings (VOTT), using a standard conditional logit model for the multinomial mode choice problem, was determined to be 12.64 USD. This estimate, along with the mean VOTT estimates from our models and the general-purpose GNN, are summarized in Table \ref{tab:VOT_multinomial}. Our proposed models yield reasonable median VOTT estimates. In contrast, the general-purpose deep learning architecture produced negative median VOTT estimates for both modes. These findings, along with the ones for the marginal utilities presented earlier, highlight the potential of our models to provide more plausible estimates than the ones that could be generated from off-the-shelf deep learning models. \\

\begin{table}
    \centering
    \caption{Median Value of travel time savings [USD/hour].}
    \begin{tabular}{ccccc}
    \hline
         &  Conditional Logit & GNN & Skip-GNN & Skip-GNN-IIA\\
    \hline
       VOTT Car  & 12.64 & -9.03 & 35.54 & 19.72\\
       VOTT Transit  & 12.64  & -2.61 & 12.12 & 11.95 \\
    \hline
    \end{tabular}
    \label{tab:VOT_multinomial}
\end{table}

\begin{figure}[ht]
    \centering
    \begin{minipage}[t]{.45\textwidth}
        \centering
        \includegraphics[width=\linewidth]{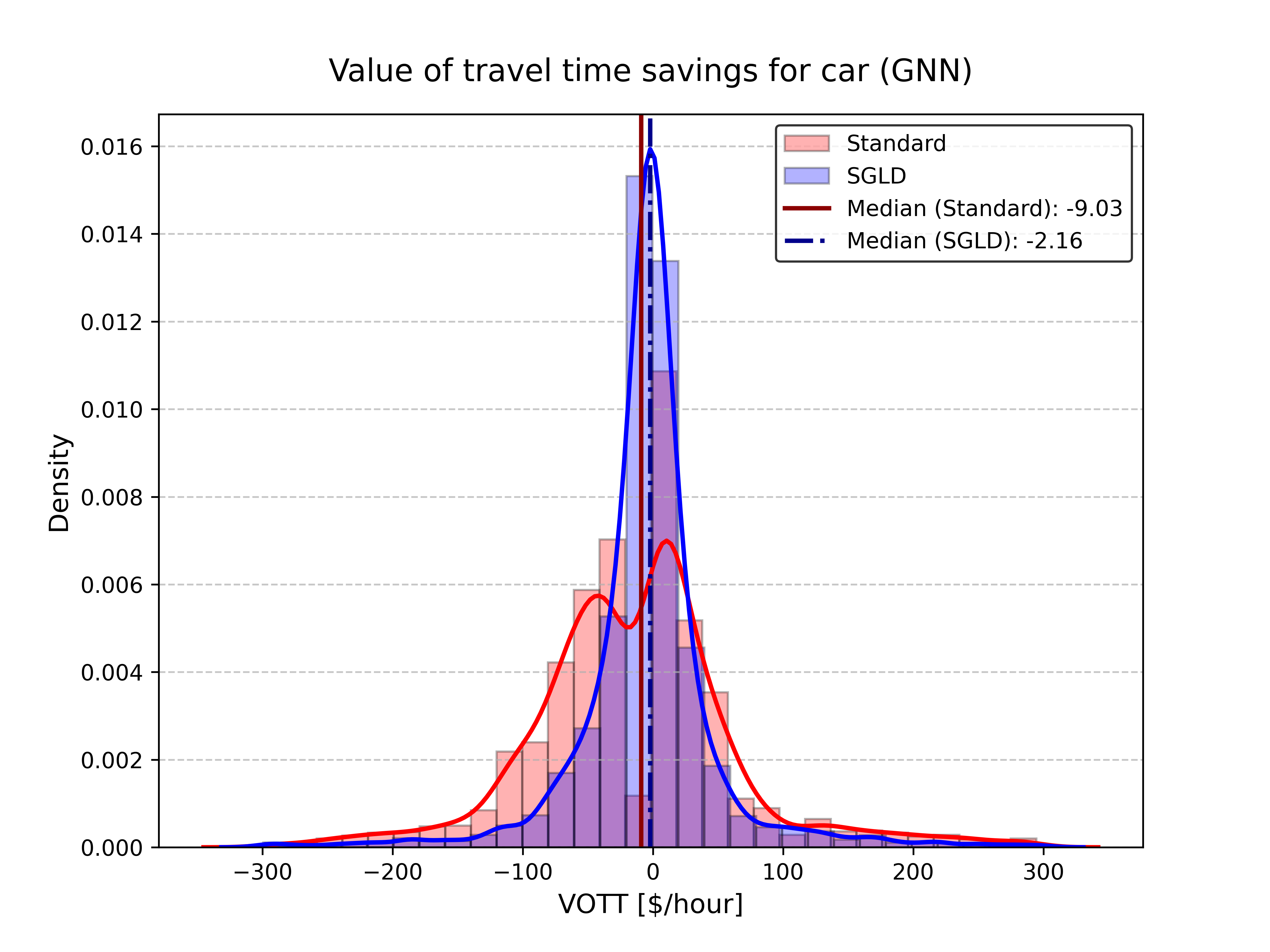}
        \caption{Value of travel time savings for car from the general purpose GNN.}
        \label{fig:VOT_car_GNN_Multinomial}
    \end{minipage}%
    \hfill
    \begin{minipage}[t]{.45\textwidth}
        \centering
        \includegraphics[width=\linewidth]{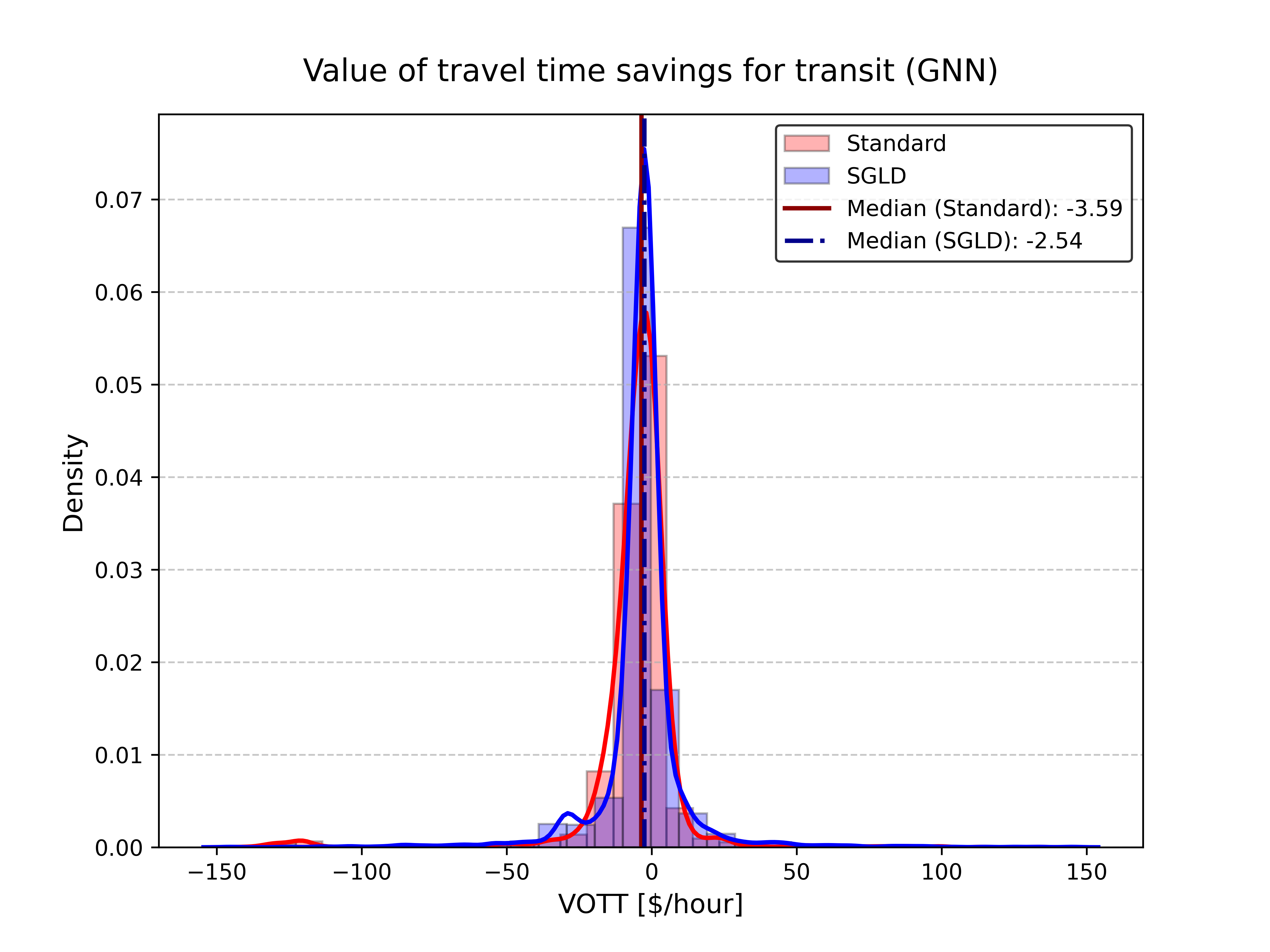}
        \caption{Value of travel time savings for transit from the general purpose GNN.}
        \label{fig:VOT_transit_GNN_Multinomial}
    \end{minipage}
\end{figure}

\begin{figure}[ht]
    \centering
    \begin{minipage}[t]{.45\textwidth}
        \centering
        \includegraphics[width=\linewidth]{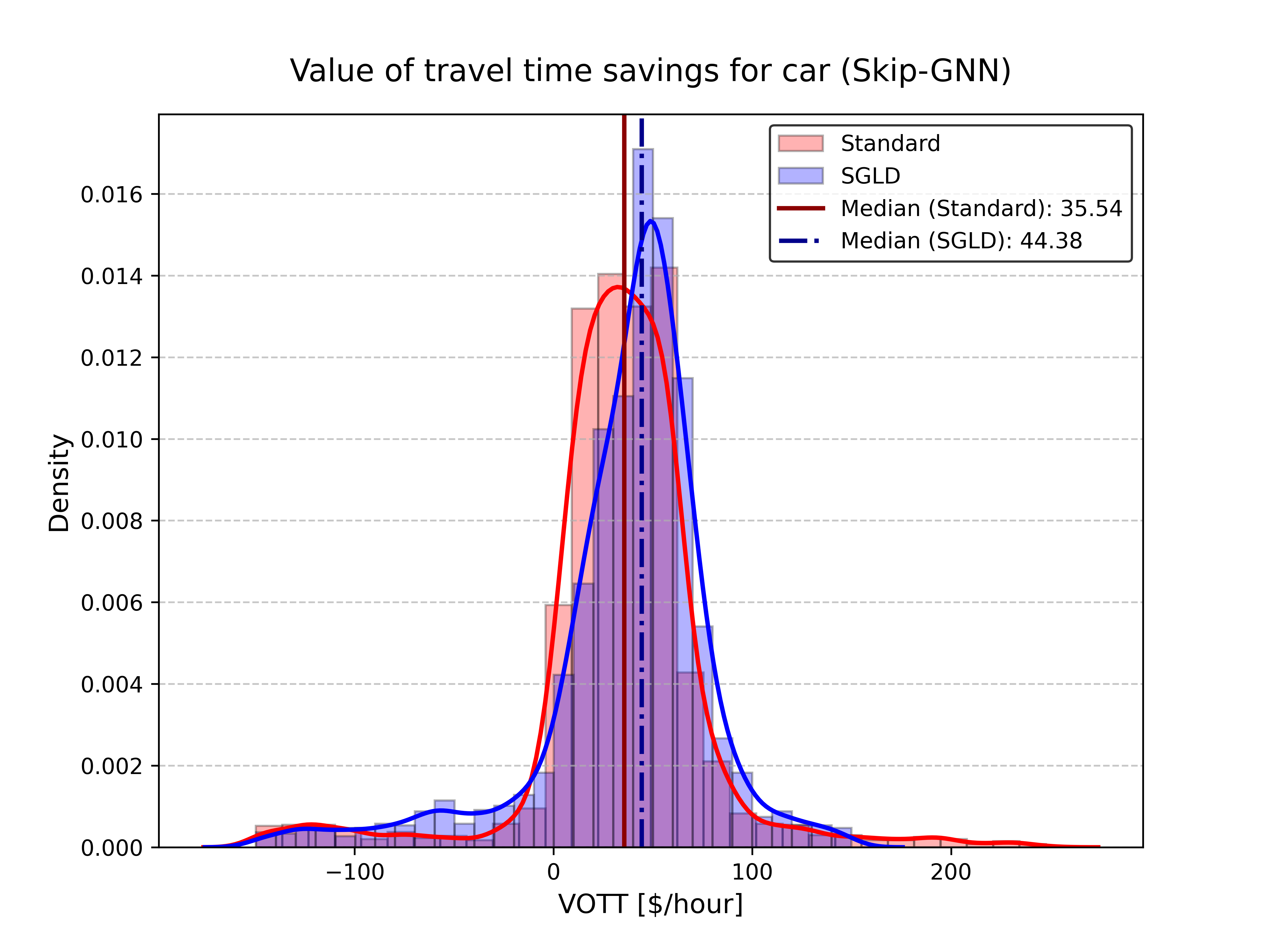}
        \caption{value of travel time savings for car from the Skip-GNN model.}
        \label{fig:VOT_car_Skip-GNN_Multinomial}
    \end{minipage}%
    \hfill
    \begin{minipage}[t]{.45\textwidth}
        \centering
        \includegraphics[width=\linewidth]{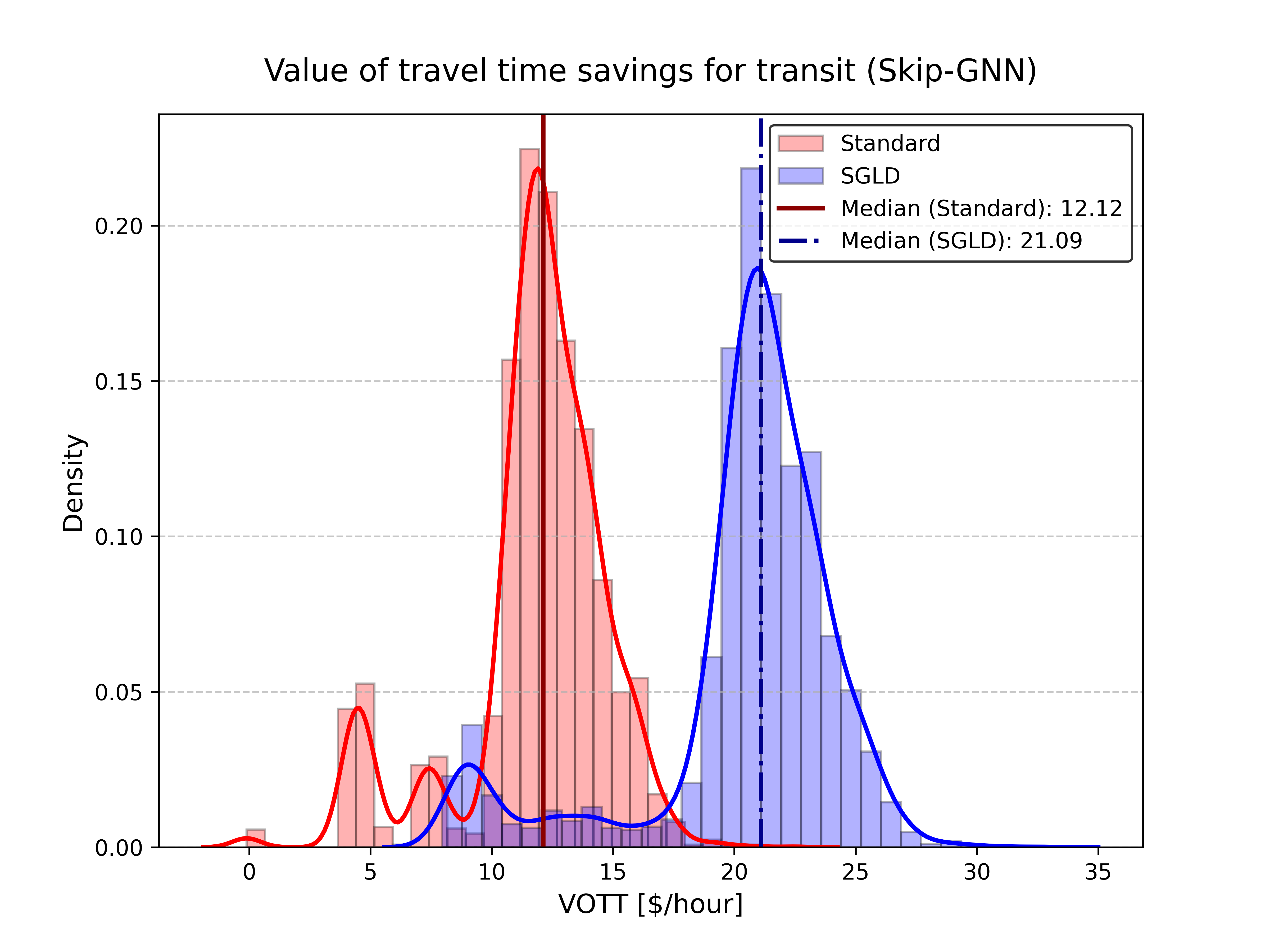}
        \caption{value of travel time savings for transit from the Skip-GNN model.}
        \label{fig:VOT_transit_Skip-GNN_Multinomial}
    \end{minipage}
\end{figure}

\begin{figure}[ht]
    \centering
    \begin{minipage}[t]{.45\textwidth}
        \centering
        \includegraphics[width=\linewidth]{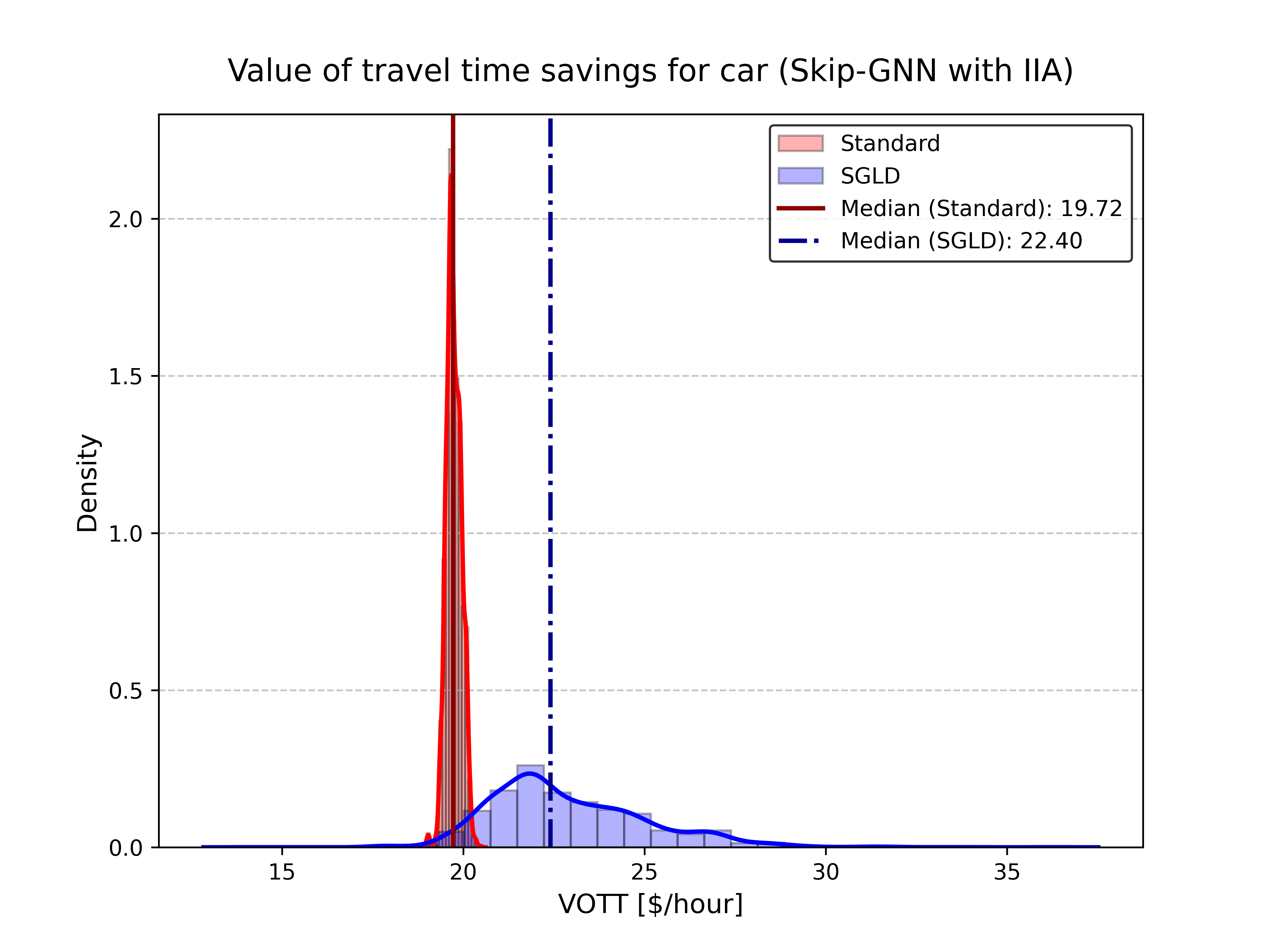}
        \caption{value of travel time savings for car from the Skip-GNN-IIA model.}
        \label{fig:VOT_car_Skip-GNN-IIA_Multinomial}
    \end{minipage}%
    \hfill
    \begin{minipage}[t]{.45\textwidth}
        \centering
        \includegraphics[width=\linewidth]{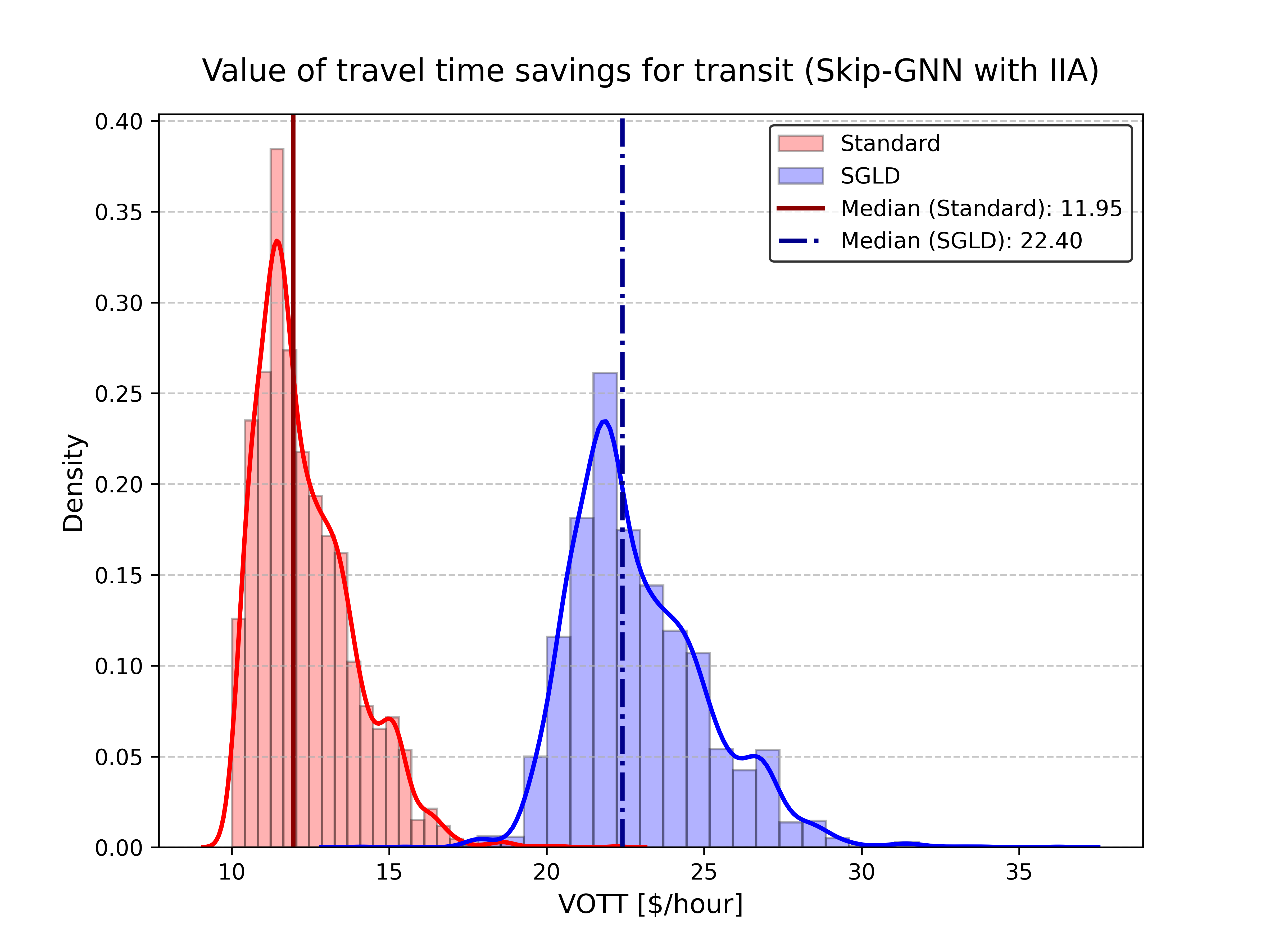}
        \caption{value of travel time savings for transit from the Skip-GNN-IIA model.}
        \label{fig:VOT_transit_Skip-GNN-IIA_Multinomial}
    \end{minipage}
\end{figure}

In Figures \ref{fig:VOT_car_GNN_Multinomial} through \ref{fig:VOT_transit_Skip-GNN-IIA_Multinomial}, we present the histograms of the value of travel time savings (VOTT) for car and transit modes as estimated by the general-purpose GNN and our two model variants. For the general-purpose GNN, a significant number of individuals exhibit negative VOTTs, which contrasts with the consistently positive VOTT estimates for the majority of the sample produced by our models. This result aligns with our expectations, as we have previously discussed that our models' marginal utilities are consistent with behavioral intuition. \\


\subsection{Case Study: Binary US Election}

Inference for the binary U.S. election is conducted with respect to median income, unemployment rate, and bachelor's degree rate. We compute the odds ratios for these variables across all models and compare their values. For logit models, the odds ratios are constant and given by $\exp(\hat{\beta_k})$, where $\hat{\beta_k}$ is the estimated parameter associated with the variable of interest. For the GNN and Skip-GNN models,  odds ratios are computed from marginal effects using the partial derivative of the representative utilities with respect to the socio-demographic $x_k$. Since the odds ratios for the GNN and Skip-GNN models depend on the values of the socio-demographic vector $\bm{x}$, we compute the odds ratios for each individual, calculate their means and medians across the sample, and present their histograms.\\

\subsubsection{Median Income Odds Ratio}

Using the logit model, we found an odds ratio of 1.10 for a \$1000 USD increase in county median income. This implies that with a \$1000 USD increase, the odds of voting for Trump increase by 10\% according to the logit model. In Table \ref{tab:odds_income}, we present the mean and median values of the odds ratio for median income found using the GNN and Skip-GNN models.\\

\begin{table}[h!]
    \centering
    \caption{Median Income Odds Ratio.}
    \begin{tabular}{ccccc}
    \hline
         &  GNN & GNN (SGLD) & Skip-GNN & Skip-GNN (SGLD)\\
    \hline
       Mean & 1.14 & 1.28 & 1.13 & 1.15 \\
       Median  & 1.15 & 1.28 & 1.10 & 1.17 \\
    \hline
    \end{tabular}
    \label{tab:odds_income}
\end{table}

\begin{figure}[ht]
    \centering
    \begin{minipage}[t]{.45\textwidth}
        \centering
        \includegraphics[width=\linewidth]{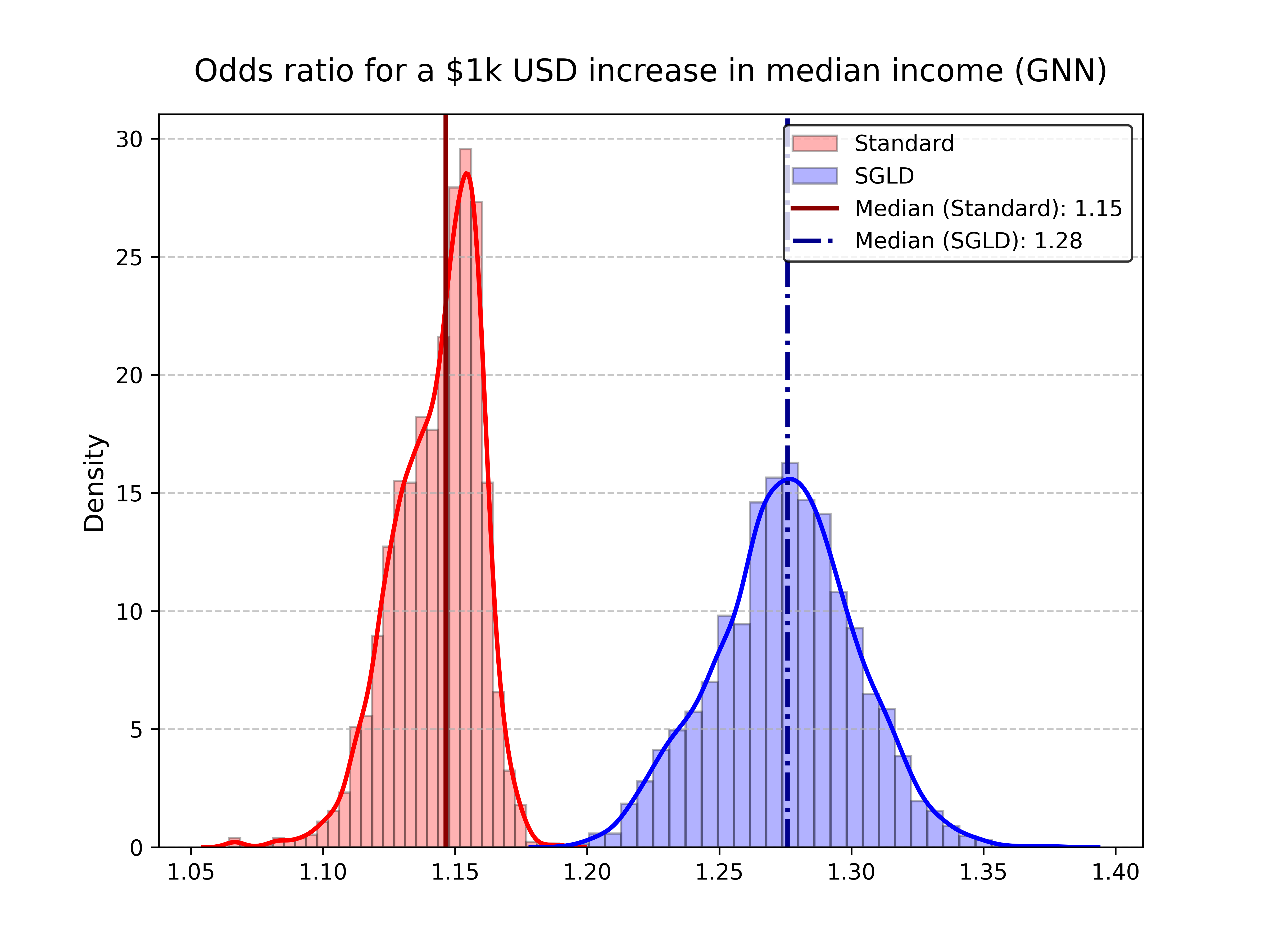}
        \caption{Odds ratio for an income increase of \$1000 USD under the GNN model.}
        \label{fig:odds_income_GNN}
    \end{minipage}%
    \hfill
    \begin{minipage}[t]{.45\textwidth}
        \centering
        \includegraphics[width=\linewidth]{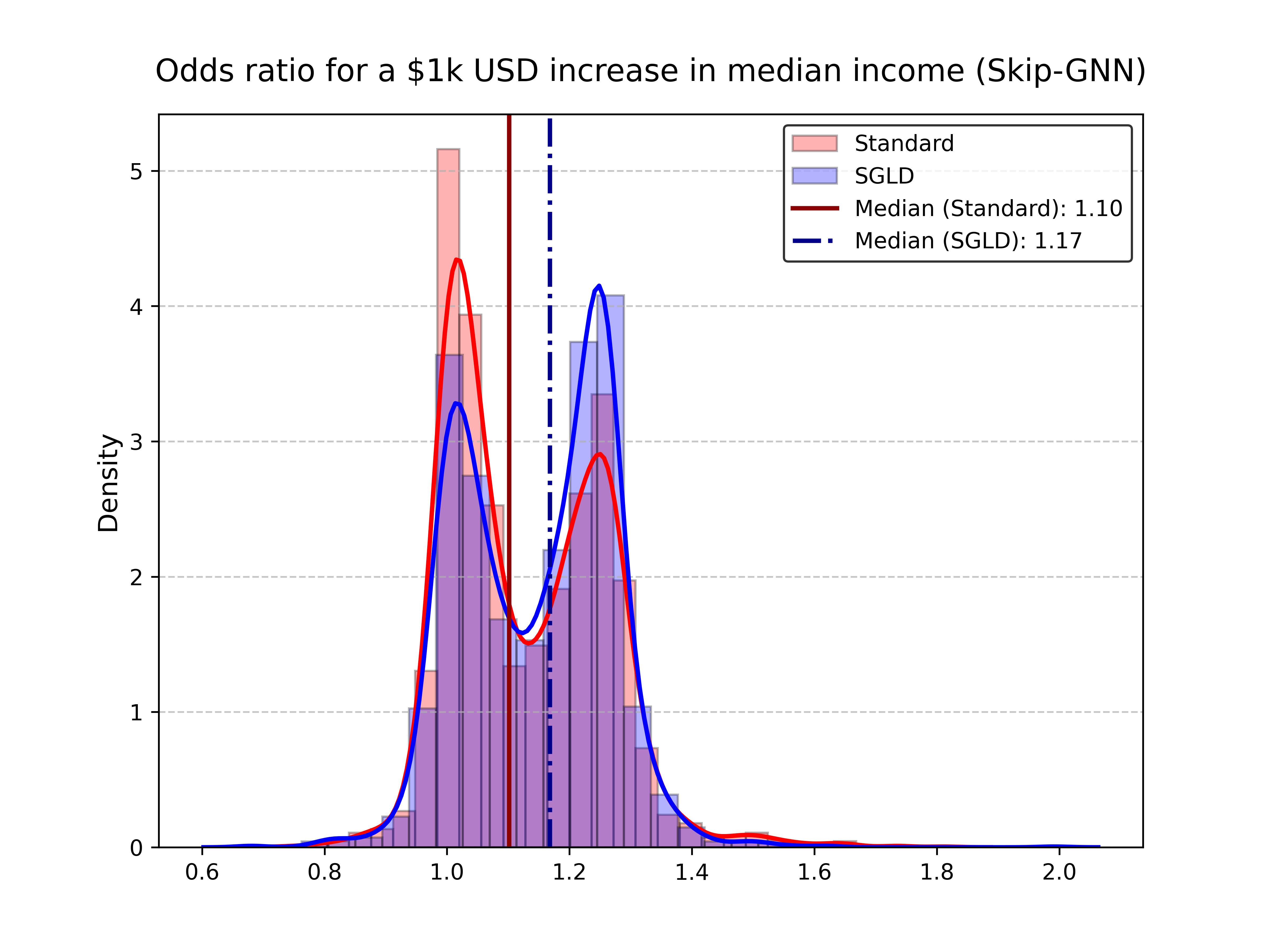}
        \caption{Odds ratio for an income increase of \$1000 USD under the Skip-GNN model.}
        \label{fig:odds_income_skip_GNN}
    \end{minipage}
\end{figure}

The odds ratio histogram for the GNN model is presented in Figure \ref{fig:odds_income_GNN}, while the histogram for the Skip-GNN model is presented in Figure \ref{fig:odds_income_skip_GNN}. \\

 For both models, the odds ratios are generally above 1.00, which indicates that an increase in the average county income increases the odds of voting for Trump in most counties.

\subsubsection{Bachelor Degree Rate Odds Ratio}

Under the logit model, the odds ratio for a 1\% increase in the bachelor's degree rate is 0.82. This means that a 1\% increase in the percentage of people with a bachelor's degree in a county decreases the odds of voting for Trump by 18\%. In Table \ref{tab:odds_bachelor}, we present the mean and median values of the odds ratio for an increase of 1\% in the bachelor rate found using the GNN and Skip-GNN models.\\

\begin{table}[h!]
    \centering
    \caption{Bachelor Rate Odds Ratio.}
    \begin{tabular}{ccccc}
    \hline
         &  GNN & GNN (SGLD) & Skip-GNN & Skip-GNN (SGLD)\\
    \hline
       Mean & 0.70 & 0.53 & 0.69 & 0.68 \\
       Median & 0.70 & 0.54 & 0.71 & 0.66 \\
    \hline
    \end{tabular}
    \label{tab:odds_bachelor}
\end{table}

\begin{figure}[ht]
    \centering
    \begin{minipage}[t]{.45\textwidth}
        \centering
        \includegraphics[width=\linewidth]{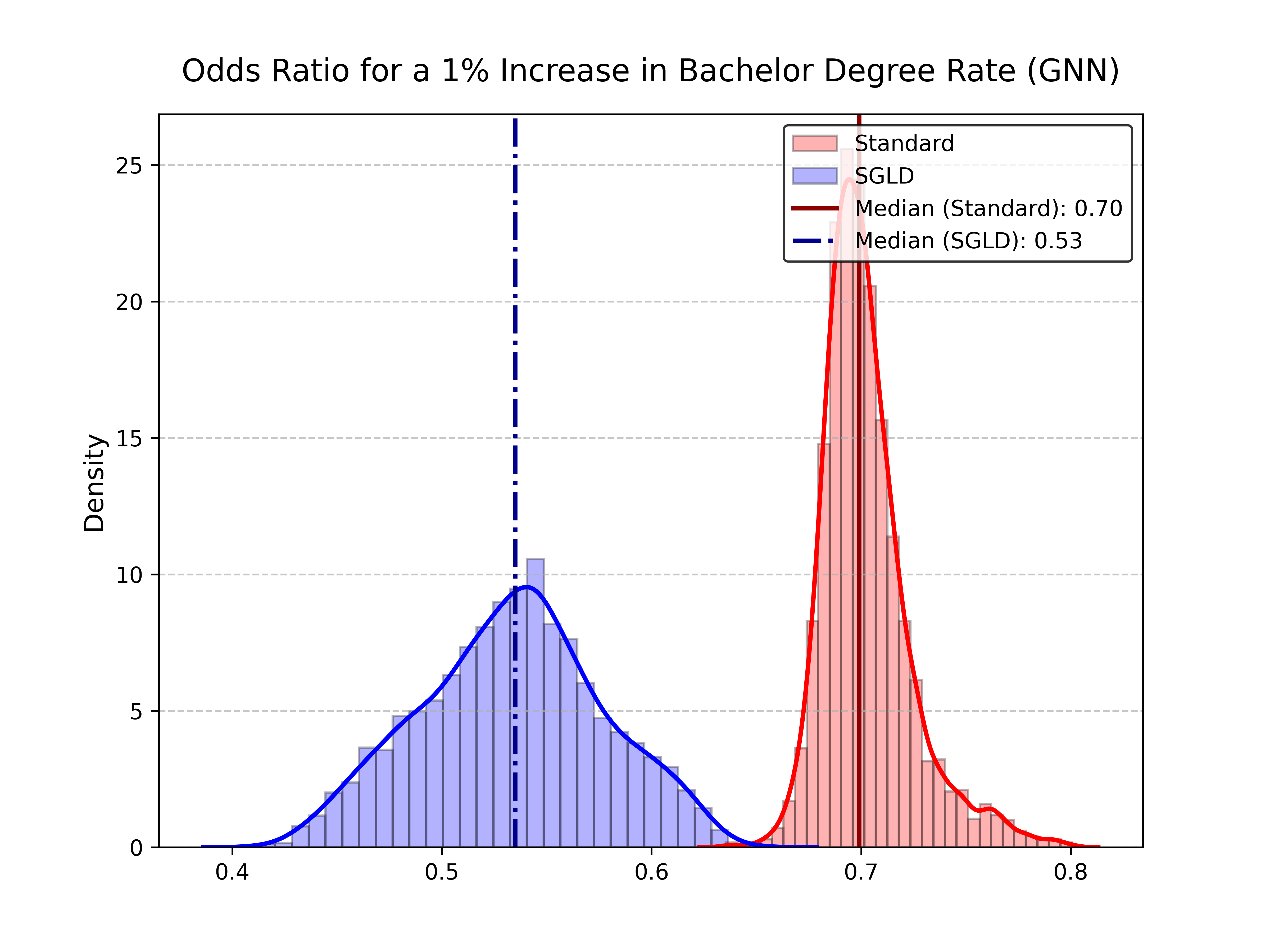}
        \caption{Odds ratio for an increase of 1\% in the bachelor rate under the GNN model.}
        \label{fig:odds_bachelor_GNN}
    \end{minipage}%
    \hfill
    \begin{minipage}[t]{.45\textwidth}
        \centering
        \includegraphics[width=\linewidth]{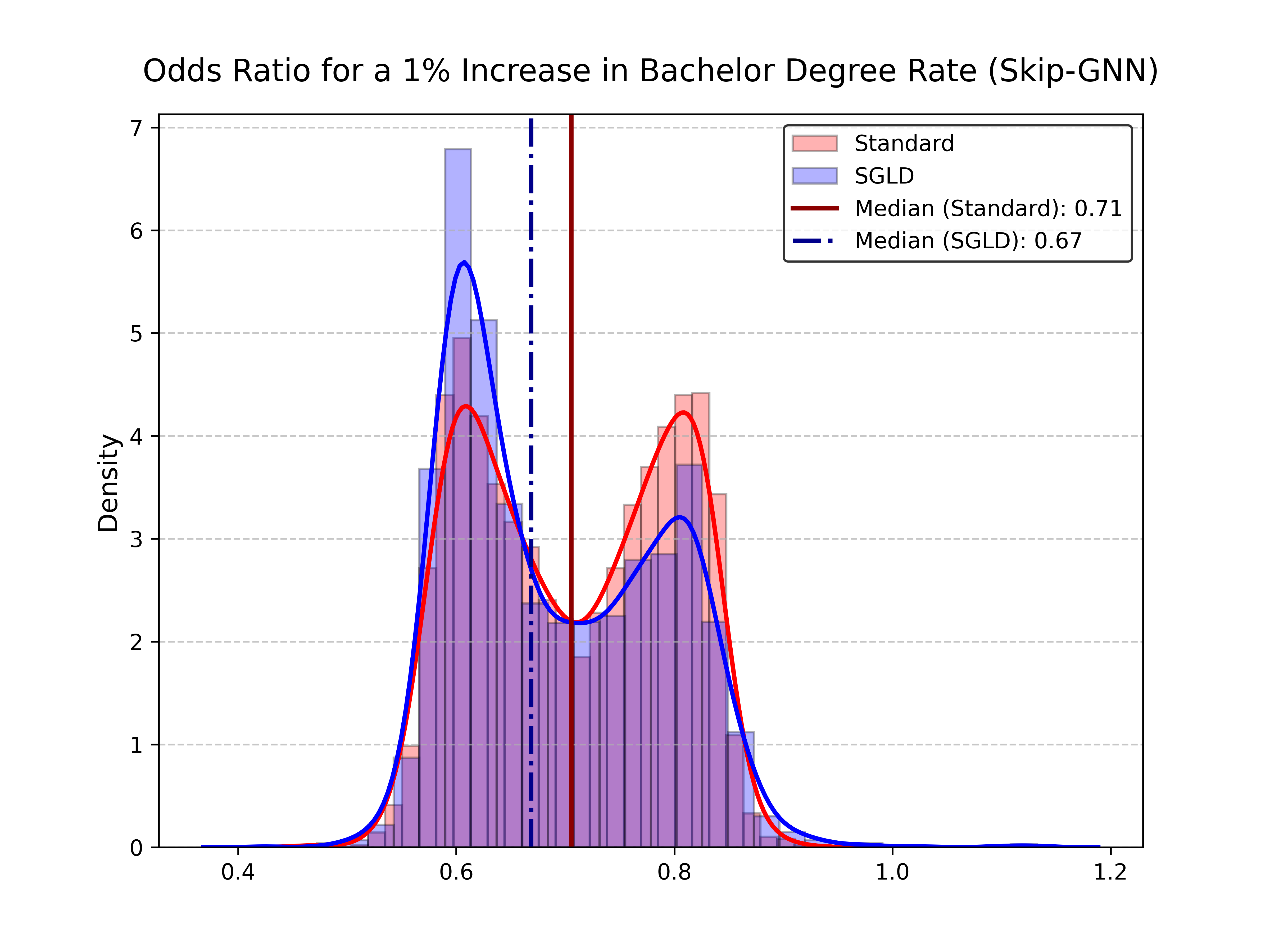}
        \caption{Odds ratio for for an increase of 1\% in the bachelor rate under the Skip-GNN model.}
        \label{fig:odds_bachelor_skip_GNN}
    \end{minipage}
\end{figure}

The odds ratio histogram for the GNN model is presented in Figure \ref{fig:odds_bachelor_GNN}, while the histogram for the Skip-GNN model is presented in Figure \ref{fig:odds_bachelor_skip_GNN}. \\

 For both models, the odds ratios are generally below 1.00, which indicates that an increase in the bachelor's degree rate reduces the odds of voting for Trump in most counties.

\subsubsection{Unemployment Rate Odds Ratio}

The odds ratio for a 1\% increase in the unemployment rate under the logit model is 0.64. Therefore, a 1\% increase in the unemployment rate decreases the odds of voting for Trump by 36\%. In Table \ref{tab:odds_unemployment}, we present the mean and median values of the odds ratio for an increase of 1\% in the unemployment rate found using the GNN and Skip-GNN models

\begin{table}[h!]
    \centering
    \caption{Unemployment Rate Odds Ratio.}
    \begin{tabular}{ccccc}
    \hline
         &  GNN & GNN (SGLD) & Skip-GNN & Skip-GNN (SGLD)\\
    \hline
       Mean & 0.67 & 0.66 & 0.53 & 0.57 \\
       Median  & 0.66 & 0.66 & 0.53 & 0.54 \\
    \hline
    \end{tabular}
    \label{tab:odds_unemployment}
\end{table}

\begin{figure}[ht]
    \centering
    \begin{minipage}[t]{.45\textwidth}
        \centering
        \includegraphics[width=\linewidth]{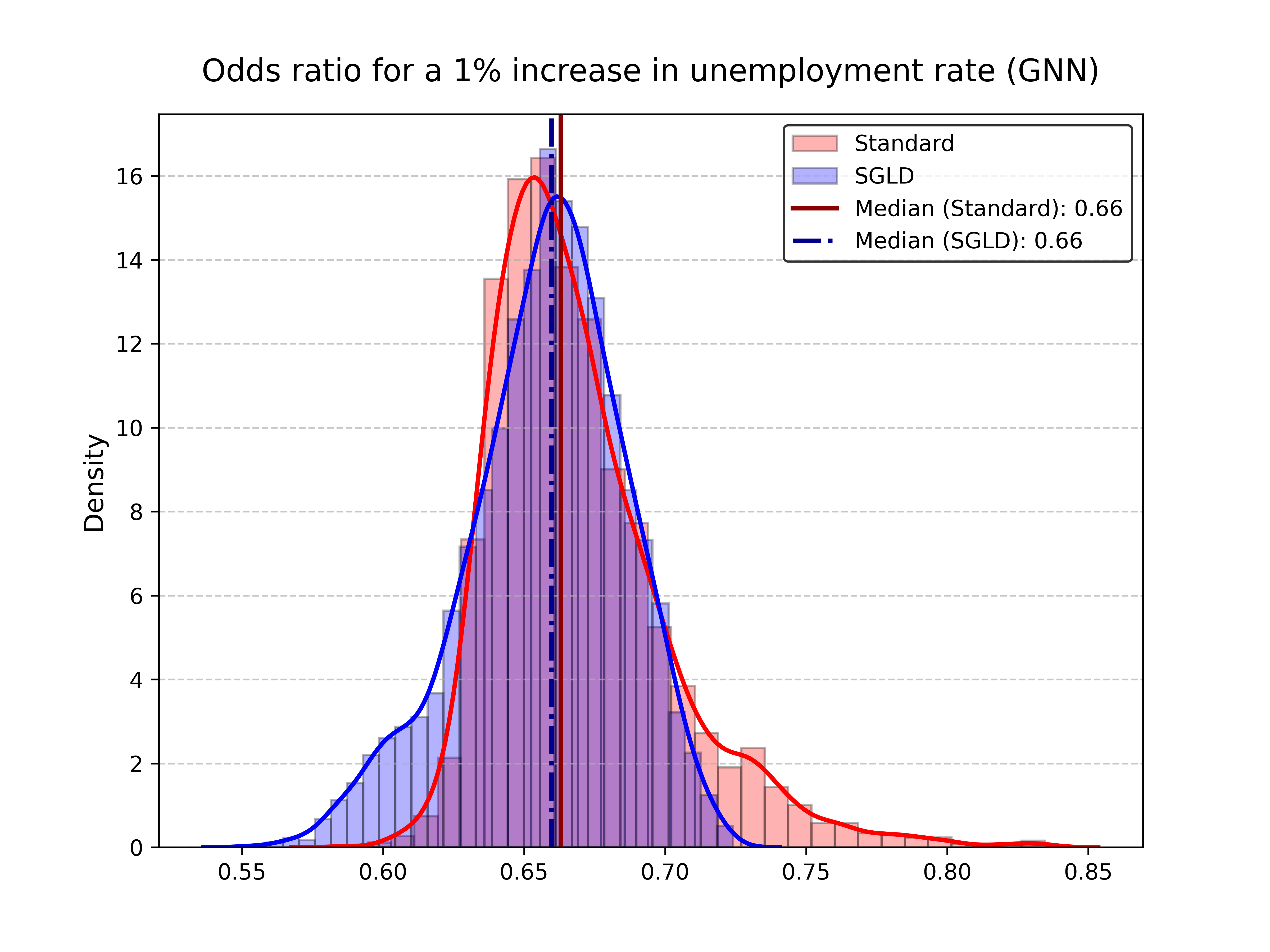}
        \caption{Odds ratio for an increase of 1\% in the unemployment rate under the GNN model.}
        \label{fig:odds_unemployment_GNN}
    \end{minipage}%
    \hfill
    \begin{minipage}[t]{.45\textwidth}
        \centering
        \includegraphics[width=\linewidth]{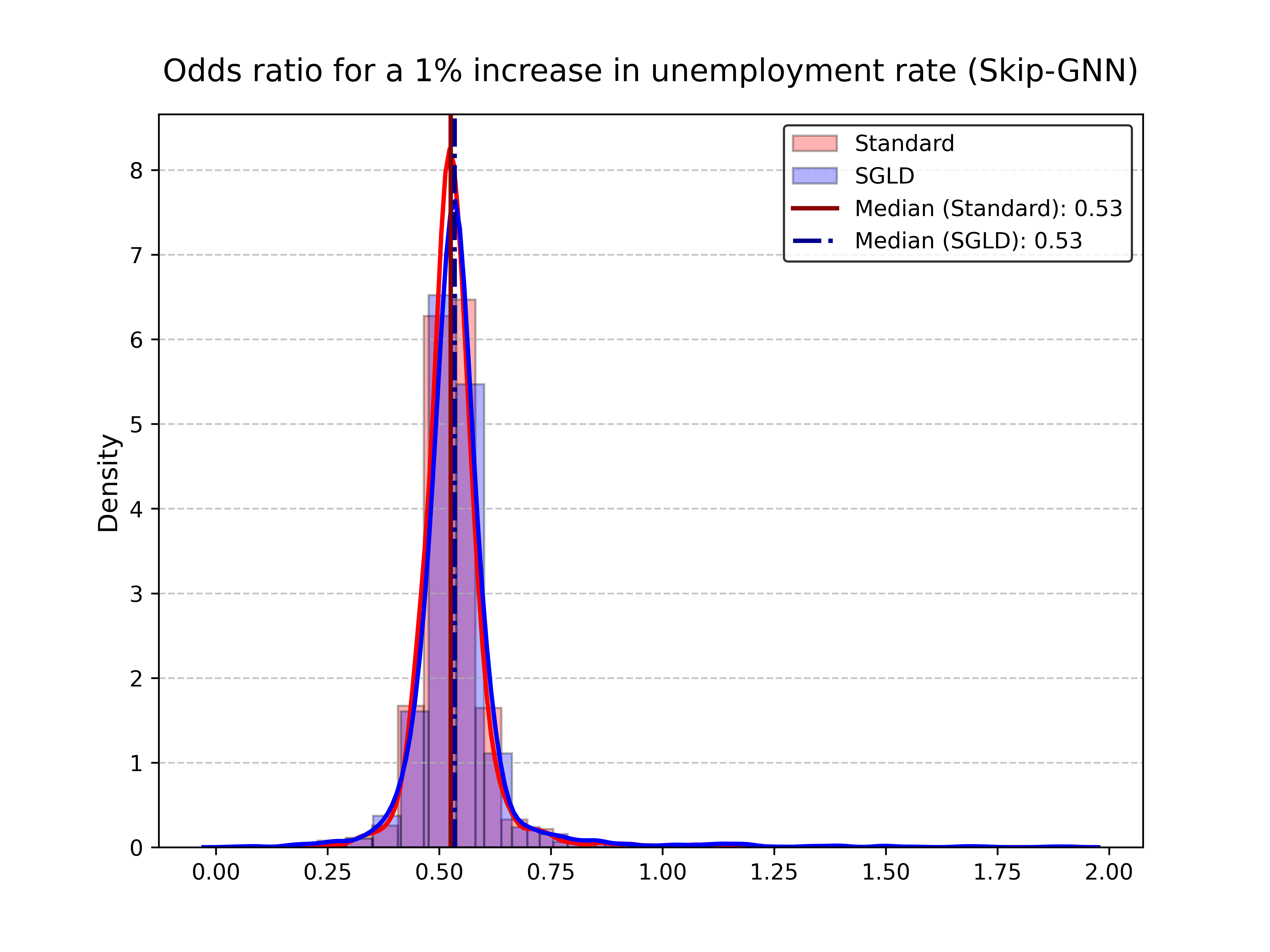}
        \caption{Odds ratio for for an increase of 1\% in the unemployment rate under the Skip-GNN model.}
        \label{fig:odds_unemployment_skip_GNN}
    \end{minipage}
\end{figure}

The histogram for the GNN model is presented in Figure \ref{fig:odds_unemployment_GNN}, while the one for the Skip-GNN model is presented in Figure \ref{fig:odds_unemployment_skip_GNN}. \\

As with the case, under both models, the results indicate that with a 1\% increase in the unemployment rate, the odds of voting for Trump decrease in most counties.

\subsection{Model Accuracy across Case Studies}

To estimate the prediction performance out of sample of all models, we employed 5-fold cross-validation. In all cases, we compute the weighted (or balanced) accuracy, which is the average of specificity and sensitivity, also called the true positive and true negative rates for binary classification problems. In the multinomial mode choice problem, the weighted accuracy is calculated as the average per-class precision. This metric provides a more reliable measure of predictive performance as it accounts for class imbalances.\\

We observed a very modest improvement in prediction accuracy for our Skip-GNN model compared to the logit model in the binary mode choice scenario, and around a 6 percentage point increase in the multinomial mode choice setting (for both the IIA and unconstrained versions of our model). For the binary election dataset, we observed an improvement of more than 6 percentage points for our Skip-GNN model compared to the standard logit\footnote{To ensure a fair comparison against the deep learning models that are estimated using a weighted binary cross-entropy loss function, we estimated the logit models using per-class weights in the log-likelihood. The weighted accuracy computed for the US 2016 binary election with the logit model, without class weights, was estimated to be 72.68\%. For the mode choice problems, the weighted accuracies with and without class weights are similar to one another.}. The general-purpose GNN is outperformed by our Skip-GNN model across all datasets, and for the binary mode choice problem, it does not even achieve a prediction performance on par with the logit model.\\

These gains align with the findings of \cite{wang2021comparing}, which reported an average performance improvement of around 5 percentage points for deep neural networks over traditional Discrete Choice Models (DCMs). For our case studies, it is important to note that even for the mode choice problems where the logit model outperforms the general-purpose GNN architecture, our Skip-GNN model is capable of attaining the highest out-of-sample performance. Detailed out-of-sample weighted accuracy metrics for all estimated models are presented in Table \ref{tab:comparison}.\\

\begin{table}[h!]
\small
	\centering
	\caption{Accuracy on the test set model comparison}
	\begin{tabular}{ccccc}
		\hline
		 & Logit & Skip-GNN (IIA) & Skip-GNN & GNN\\
		\hline
		\centering Binary mode choice &  83.92\% & NA & \textbf{84.53\%} & {\color{red} 76.75\%} \\
        \centering Multinomial mode choice &  72.66\% & \textbf{77.70\%} & \textbf{78.45\%} & {\color{red} 64.36\%} \\
        \centering US 2016 binary election & 81.21\% & NA & \textbf{87.29\%} & 81.87\% \\
		\hline
	\end{tabular}
	\label{tab:comparison}	
\end{table}

\subsection{Individual-Level Inference}

One of the advantages of deep learning models for discrete choice is the associate ability to perform inference at the individual level. For instance, in the binary case study presented in this paper, we estimate the Value of Travel Time (VOTT) for all individuals in the dataset, considering their socio-demographics and the attributes of the alternatives presented to them. Traditionally, in discrete choice modeling, this level of granularity is typically achieved using models that incorporate random preference heterogeneity, such as mixed-logit models, or by designing latent utility functions to capture deterministic preference heterogeneity based on socio-demographic interactions with alternative attributes. However, deep learning models achieve individual-level estimates through automatic feature learning, leveraging hidden interactions between input variables.\\

\begin{figure}[h]
    \centering
    \begin{subfigure}[t]{\textwidth}
        \centering
        \includegraphics[width=\linewidth]{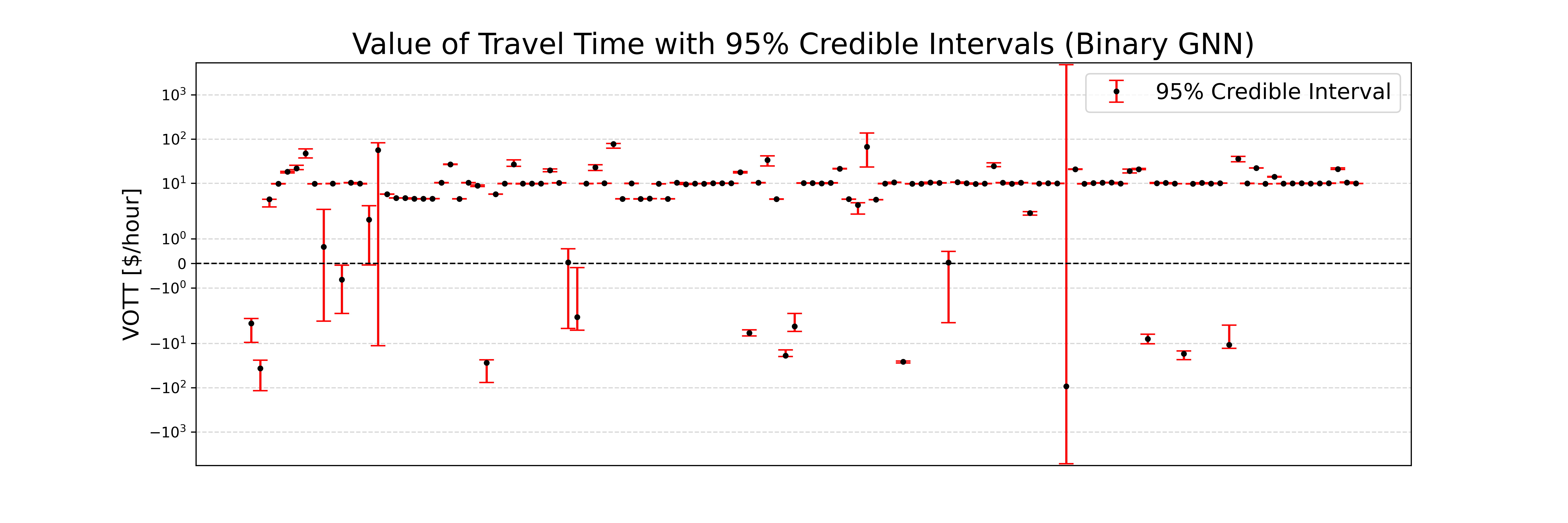}
        \caption{Credible intervals for individual-level VOTT for binary GNN. Subsample of 123 observations.}
        \label{fig:ci_vott_gnn_binary_mode}
    \end{subfigure}
    \vspace{1em} 
    \begin{subfigure}[t]{\textwidth}
        \centering
        \includegraphics[width=\linewidth]{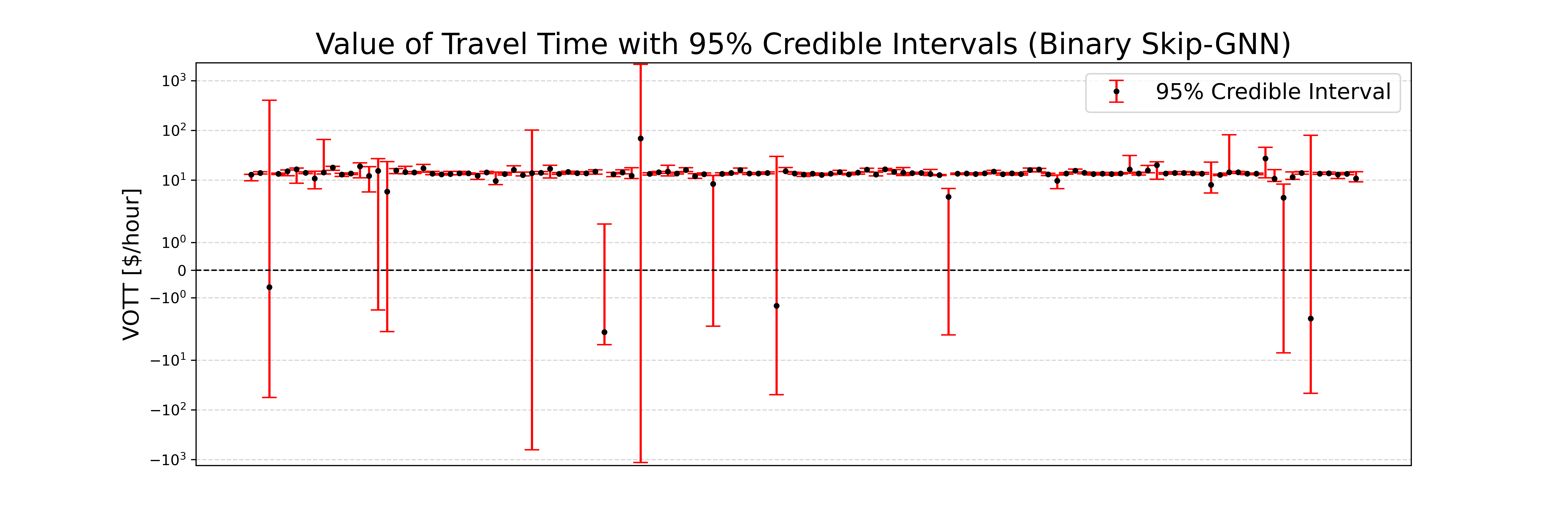}
        \caption{Credible intervals for individual-level VOTT for binary Skip-GNN. Subsample of 123 observations.}
        \label{fig:ci_vott_skip_gnn_binary_mode}
    \end{subfigure}
    \caption{Credible intervals for individual-level VOTT for the binary GNN and Skip-GNN models. Subsample of 123 observations.}
    \label{fig:ci_vott_combined}
\end{figure}

Using SGLD, we are thus able to provide individual-level estimates for the VOTT as well as 95\% credible intervals. In Figure \ref{fig:ci_vott_gnn_binary_mode}, we present the credible intervals at the individual level for the VOTT for the GNN model for a sub-sample of 123 individuals, and in Figure \ref{fig:ci_vott_skip_gnn_binary_mode}, we show the corresponding intervals for our Skip-GNN model. It is important to note that for the GNN model, there are credible intervals with upper and lower limits that are negative. This indicates, with high posterior probability, that the VOTT for these individuals is negative, a result that defies behavioral intuition. In contrast, for our Skip-GNN model, while some individual-level VOTT median values are negative, the credible intervals for those individuals include both positive and negative values. This implies that there is insufficient posterior evidence under our model to suggest that the VOTT for those individuals is negative. \\ 

\begin{figure}[h]
    \centering
    \begin{subfigure}[t]{\textwidth}
        \centering
        \includegraphics[width=\textwidth]{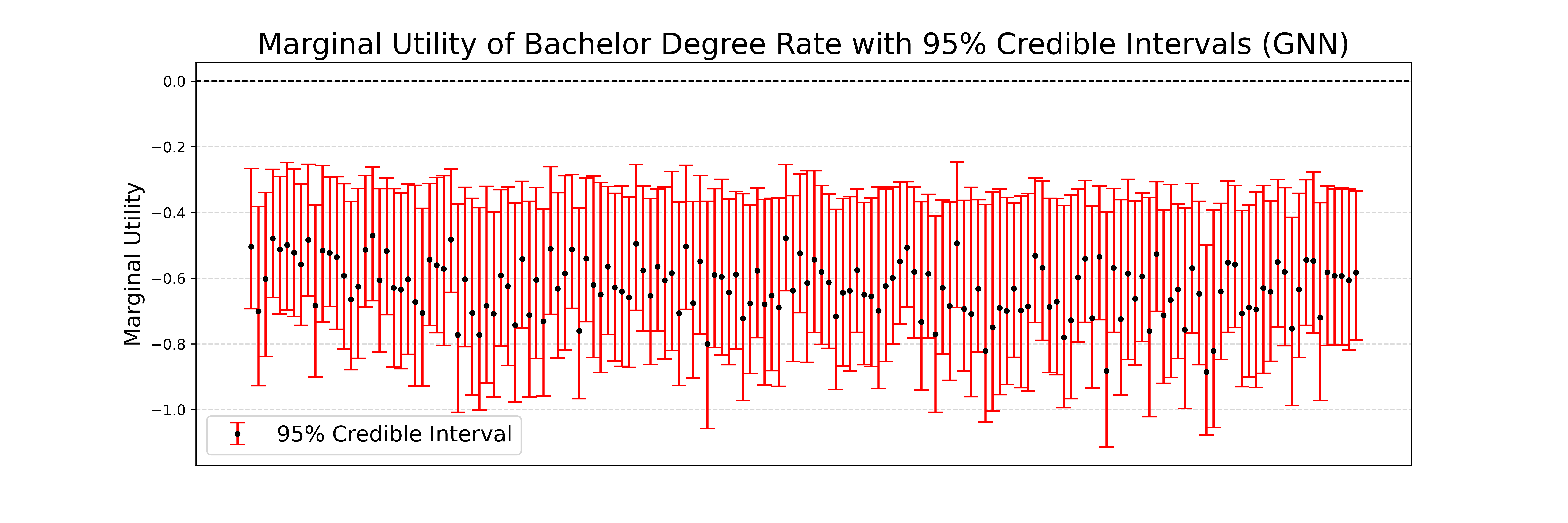}
        \caption{Credible intervals for county-level bachelor rate odds-ratio for GNN. Subsample of 156 observations.}
        \label{fig:ci_bachelor_gnn_election}
    \end{subfigure}
    \vspace{1em} 
    \begin{subfigure}[t]{\textwidth}
        \centering
        \includegraphics[width=\textwidth]{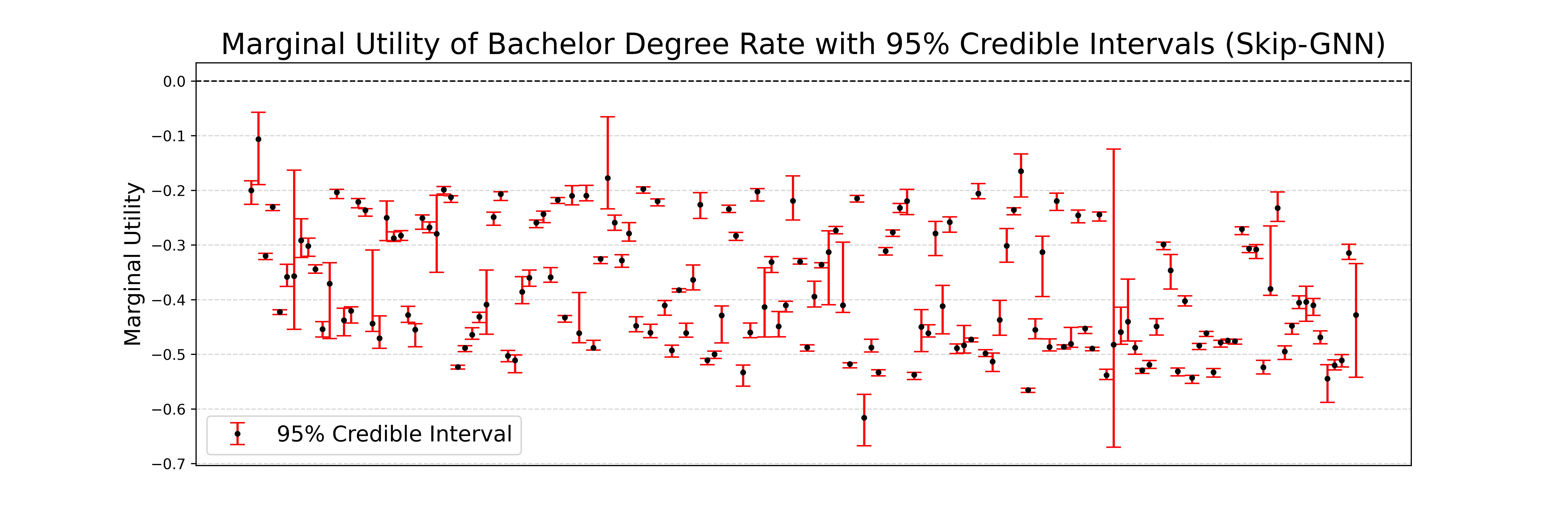}
        \caption{Credible intervals for county-level bachelor rate odds-ratio for Skip-GNN. Subsample of 156 observations.}
        \label{fig:ci_bachelor_skip_gnn_election}
    \end{subfigure}
    \caption{County-level credible intervals for bachelor rate odds-ratios using GNN and Skip-GNN models.}
    \label{fig:ci_unemployment_combined}
\end{figure}

As another example, in Figures \ref{fig:ci_bachelor_gnn_election}  and \ref{fig:ci_bachelor_skip_gnn_election} we provide the county-level credible intervals for the bachelor rate odds-ratios for both the GNN and Skip-GNN models, respectively. In this case, the credible intervals for the GNN model are evidently wider than the ones for our Skip-GNN model and the credible intervals overlap. 

As mentioned earlier, this paper does not examine the representation of epistemic uncertainty for constructing credible intervals and performing hypothesis testing. Instead, the aim of this section is to highlight the importance of interval estimation and demonstrate how it enables researchers to draw meaningful conclusions and compare insights derived from competing models. Future work will provide a more comprehensive comparison of methods for representing epistemic uncertainty with Bayesian and not Bayesian methods, and evaluate their empirical coverage through simulation analysis.

\section{Conclusions \label{conclusions}}

In this paper, we have introduced a novel and interpretable GNN architecture for discrete choices, that we have named Skip-GNN, that captures social influence and exogenous interaction effects. Our model design enables interactions between attributes and socio-demographics through automatic feature learning. Our design decisions set the general scale and location of utilities and clearly distinguish the individual's private and socially informed components of utilities. Importantly, we observe that our approach enhances the behavioral intuition of the econometric parameter estimates without relying on hard or soft gradient constraints. \\

We have presented our model making sure  to provide justifications for every design decision based on both discrete choice microeconometric theory and empirical findings from the machine learning literature. We offer more than just a useful deep learning architecture capable of modeling social influence and representing the Independence of Irrelevant Alternatives (IIA) if needed; we also equip discrete choice modelers with design principles. For example, the use of Batch Normalization (BatchNorm) helps set the scale and location of utilities. We have demonstrated that these principles can be used to develop models that not only deliver superior behavioral insights compared to off-the-shelf deep learning models but also attain high predictive performance. \\

We provide a binary version of our model and two alternatives for multinomial problems: one that reflects IIA and another without that constraint (which is a straightforward generalization of the binary model). For the IIA scenario, we build on ideas previously presented by Wang (2020) \cite{wang2020deep} and address socio-demographic variables by creating independent embeddings for each alternative, following a similar approach to that in Arkoudi (2023) \cite{arkoudi2023combining}. We have called this IIA variant Skip-GNN-IIA. Both our Skip-GNN and Skip-GNN-IIA models exhibit high predictive performance and align with behavioral intuition, demonstrating that our model is suitable for either modeling flexible substitution patterns or for being restricted by IIA. \\

We tested our models on mode choice data from NYC in both binary and multinomial settings. Our models attain the highest out-of-sample prediction performance and exhibit strong alignment with behavioral intuition. For instance, in estimating the marginal utilities of travel time and trip cost, which are attributes that when increased make alternatives less attractive, our architecture consistently produces negative values across the sample of individuals. This contrasts with the outcomes from a general-purpose deep learning model which, despite presenting negative median values for the sample, reveals a substantial proportion of individuals with positive marginal utilities for travel time and trip cost. This observation, coupled with the detailed interpretation of each design decision, underscores the importance of tailoring architectures to provide more reliable and interpretable behavioral insights. \\

We also used 2016 binary election data from the U.S., aggregated by county, along with a network constructed using Facebook friendship data \cite{Tomlinson_Benson_2024}. We observed a significant improvement in the weighted accuracy of our Skip-GNN model compared to both standard GNNs and traditional logit models. Leveraging the interpretability of our models, we computed estimates for the odds ratios associated with an increase in median income, bachelor's degree rate, and unemployment rate. Our findings suggest that an increase in median income raises the odds of voting for Trump, whereas an increase in the bachelor's degree rate and unemployment rate decreases them in most counties. \\

While providing insights aligned with behavioral intuition, we also observe an increase in out-of-sample weighted accuracy for our models compared to standard DCMs and off-the-shelf GCNs across all three datasets, with a 6-percentage-point improvement in both the binary election data and the multinomial mode choice dataset. These results align with previous empirical studies showing performance gains when applying deep learning to discrete choice settings \cite{wang2021comparing}. \\

We used Stochastic Gradient Langevin Dynamics (SGLD) to account for epistemic uncertainty. SGLD enabled us to provide interval estimates for the value of travel time savings (VOTT) and odds ratios at the individual level. To our knowledge, this study is the first to provide interval estimates for a deep learning model applied to discrete choices. In future work, we will focus more comprehensively on interval estimation and hypothesis testing using approximate Bayesian inference. \\

Future work will thus focus on two main areas, namely: (i) using approximate Bayesian approaches in deep learning to better represent the epistemic uncertainty associated with discrete choice models, and (ii) exploring the connection between the private component of utility, as presented in this paper, and Gaussian Processes. Ultimately, we believe that incorporating deep learning into discrete choice modeling can be effectively achieved by carefully designing architectures (such as the one presented here), approaching the estimation problem from a Bayesian perspective, or—most likely—a combination of both.

\section*{Acknowledgments}
 This research was supported by the National Science Foundation Award No. SES-2342215. We are also thankful for the financial support provided by the Fulbright Scholarship Program, which is sponsored by the U.S. Department of State, the Colombian Fulbright Commission, and the Colombian Science Ministry. This project is solely the responsibility of the authors and does not necessarily represent the official views of the Fulbright Program, the U.S. government, or the Colombian government.

\newpage
\bibliographystyle{elsarticle-num-names} 
\bibliography{biblio}

\end{document}